\documentclass[letterpaper]{article} %
\usepackage{aaai25}  %
\usepackage{times}  %
\usepackage{helvet}  %
\usepackage{courier}  %
\usepackage[hyphens]{url}  %
\usepackage{graphicx} %
\usepackage{natbib}  %
\usepackage{caption} %
\usepackage{algorithm}
\usepackage{algorithmic}

\usepackage[size=tiny]{todonotes}
\usepackage{amsfonts}
\usepackage{amsmath}
\usepackage{cleveref} %
\usepackage[ruled,vlined, algo2e]{algorithm2e}
\usepackage{subcaption}
\usepackage{multirow}

\title{Differentially Private Prototypes for Imbalanced Transfer Learning}
\author{
    Dariush Wahdany\textsuperscript{\rm 1,3}\thanks{Part of the work was conducted as visiting researcher at CISPA},
    Matthew Jagielski\textsuperscript{\rm 2},
    Adam Dziedzic\textsuperscript{\rm 3},
    Franziska Boenisch\textsuperscript{\rm 3}
}
\affiliations{
    \textsuperscript{\rm 1}Fraunhofer AISEC\\
    \textsuperscript{\rm 2}Google DeepMind\\
    \textsuperscript{\rm 3}CISPA Helmholtz Center for Information Security\\
    dariush.wahdany@aisec.fraunhofer.de, jagielski@google.com,  adam.dziedzic@cispa.de, boenisch@cispa.de
}

\usepackage{xspace}

\def\*#1{\mathbf{#1}}
\newcommand\norm[1]{\lVert#1\rVert}

\DeclareMathOperator*{\argmin}{arg\,min}
\newtheorem{definition}{Definition}
\newtheorem{theorem}{Theorem}

\newtheorem{lemma}{Lemma}

\newtheorem{proposition}[theorem]{Proposition}

\newcommand{\ie}{\textit{i.e.,}\@\xspace}
\newcommand{\eg}{\textit{e.g.,}\@\xspace}

\newcommand{\ourstext}{DPPL}
\newcommand{\ours}{\texttt{\ourstext}\@\xspace}
\newcommand{\DPthreetext}{DPPL-Public}
\newcommand{\DPfourtext}{DPPL-PublicK}
\newcommand{\DPtwotext}{DPPL-Mean}
\newcommand{\DPthree}{\texttt{\DPthreetext}\@\xspace}
\newcommand{\DPfour}{\texttt{\DPfourtext}\@\xspace}
\newcommand{\DPtwo}{\texttt{\DPtwotext}\@\xspace}

\begin{document}
\newlength{\figwidth}
\setlength{\figwidth}{0.445\columnwidth}
\frenchspacing
\maketitle
\begin{abstract}
Machine learning (ML) models have been shown to leak private information from their training datasets. 
Differential Privacy (DP), typically implemented through the differential private stochastic gradient descent algorithm (DP-SGD), has become the standard solution to bound leakage from the models.
Despite recent improvements, DP-SGD-based approaches for private learning still usually struggle in the high privacy ($\varepsilon\le1)$
and low data regimes, and when the private training datasets are imbalanced.
To overcome these limitations, we propose Differentially Private Prototype Learning (DPPL) as a new paradigm for private transfer learning.
DPPL leverages publicly pre-trained encoders to extract features from private data and generates DP prototypes that represent each private class in the embedding space and can be publicly released for inference.
Since our DP prototypes can be obtained from only a few private training data points and without iterative noise addition, they offer high-utility predictions and strong privacy guarantees even under the notion of \textit{pure DP}.
We additionally show that privacy-utility trade-offs can be further improved when leveraging the public data beyond pre-training of the encoder: in particular, we can privately sample our DP prototypes from the publicly available data points used to train the encoder.
Our experimental evaluation with four state-of-the-art encoders, four vision datasets, and under different data and imbalancedness regimes demonstrate DPPL's high performance under strong privacy guarantees in challenging private learning setups.
\end{abstract}

\section{Introduction}
Machine learning (ML) models are known to leak private information about their training datasets~\citep{carlini2022membership,fredrikson2015model,shokri2017membership}. 
As a solution to provably upper-bound privacy leakage, differential privacy (DP)~\citep{dworkCalibratingNoiseSensitivity2006} has emerged as the de-facto standard for private training.
It is usually implemented in ML through the differential private stochastic gradient descent (DP-SGD) algorithm which bounds the contribution of each data point during training and iteratively injects controlled amounts of noise~\citep{abadiDeepLearningDifferential2016}. 
Thereby, DP-SGD has been shown to increase training time and decrease the final model's utility.

While, over the last years, there has been significant progress in improving both computational 
efficiency~\cite{bu2021fast,he2022exploring,li2021large,lee2021scaling,subramani2021enabling} and privacy-utility trade-offs~\cite{bu2022scalable,de2022unlocking}, there are a few relevant setups where DP training still yields unfavorable results.
These include the \textit{high privacy regime} (expressed in DP with small values of the privacy parameter $\varepsilon$, such as $\varepsilon\le1$),
the low data regime, \ie when only a \textit{few private data points} are available for training, and when the training dataset is \textit{imbalanced}, \ie when some classes have significantly more data points than others~\citep{buda2018systematic,liu2019large,reed2001pareto}.

There are various reasons why DP training is challenging in these setups~\citep{feldmanDoesLearningRequire2020, esipova2023disparate}. One of these is that DP protects small sets of examples due to its ``group privacy'' property, providing a provable bound on how much a DP algorithm can learn from small data \cite{feldmanDoesLearningRequire2020}.
Beyond this concern, the iterative noise addition weakens the signal from the training data, especially when only a few training data points are available. %
Moreover, standard approaches for learning in imbalanced setups, such as changing the sampling \cite{domingosMetaCostGeneralMethod1999, kubat1997addressing, japkowiczClassImbalanceProblem, lewisHeterogeneousUncertaintySampling1994, ling1998data, zadaPureNoiseRescue2022}, generating synthetic data for the minority classes \cite{chawlaSMOTESyntheticMinority2002}, or weighing the training loss \cite{caoLearningImbalancedDatasets2019} are not directly compatible with DP or incur additional privacy costs.
In a similar vein, each training iteration with DP training incurs additional privacy costs~\citep{abadiDeepLearningDifferential2016}, making it hard to keep $\varepsilon$ low, \ie to stay in the high privacy regime.

\begin{figure}[t]
    \centering
    \includegraphics[width=\columnwidth]{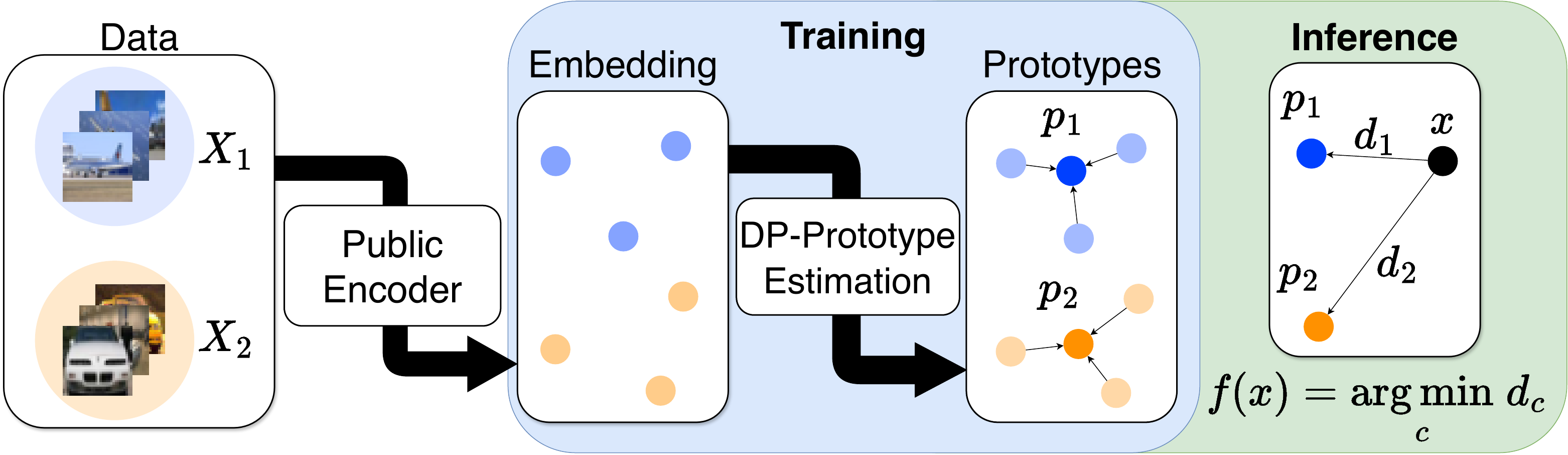}
    \caption{\textbf{Overview of \ourstext.} We split the private data $\*X$ per class $c$ into $\*X_c$'s, infer them through a publicly pre-trained encoder, and estimate per-class prototypes $\*p_c$ in the embedding space with DP.     
    Classification of samples is performed by returning the label of the closest prototype $\*p_c$ in the embedding space according to some distance function $d$.}
    \label{fig:our_method}
    \vspace{-0.4cm}
\end{figure}

To address all of these challenges, we propose \textit{Differential Private Prototype Learning} (\ours), a novel approach for private learning that combines prototypical networks~\citep{snellPrototypicalNetworksFewshot2017}, a standard algorithm for non-private few shot learning, with recent advances in training high-performance private models with DP that leverage powerful encoder models pre-trained on public data~\cite{caronEmergingPropertiesSelfSupervised2021a,he2022masked,radford2021learning} combined with private transfer learning~\cite{li2021large,yu2021differentially,gu2022choosing,tramer2022considerations,ganesh2023public,hu2021lora,houlsby2019parameter,li2023exploring,mehtaDifferentiallyPrivateImage2023}.
The main idea of our \ours is to use the encoder as a feature extractor for the private data and to generate DP prototypes in the embedding space for each private class.
To classify new data points, we then simply have to infer these points through the encoder and to return the label of the closest prototype.

Relying on DP prototypes for private learning offers significant advantages over iterative private training or fine-tuning.
First, our prototypes do not require iterative noise addition.
This enables to obtain less noisy predictions at lower privacy costs and improves privacy-utility trade-offs in the high privacy regime.
Second, the prototypes are inherently balanced, \ie it is possible to obtain good prototypes also at the low data regime or for underrepresented classes from imbalanced private training datasets.
Third, DP prototypes are fast to obtain, enable fast inference, and, due to the DP post-processing guarantees---which express that no query to them will incur additional privacy costs---can be publicly released for performing predictions.

We propose multiple algorithms for obtaining DP prototypes, and find that prior approaches for training models with DP do not yet leverage the full capacity of the public data~\cite{li2021large, yu2021differentially, mehtaDifferentiallyPrivateImage2023}:
these prior approaches use the public data only for pre-training the encoder. 
Yet, we make the observation that we can leverage the public data additionally during the transfer learning step.
By privately selecting per-class private prototypes from the public data, we can significantly decrease the privacy costs of our prototypes (even under the strong notion of \textit{pure DP}, \ie $\varepsilon$-DP)  and further improve privacy-utility trade-offs.

By performing thorough experimentation with four state-of-the-art encoders and four standard vision datasets, we show that \ours provides strong utility in the high privacy regimes. %
Additionally, we highlight that \ours is able to provide good privacy-utility trade-offs when only a few private training data points are available and that it yields state-of-the-art performance on imbalanced classification tasks.
Thereby, \ours represents a new powerful learning paradigm for private training with DP.

In summary, we make the following contributions:
\begin{itemize}
    \item We propose \ours, a novel alternative to private fine-tuning that combines recent advances in DP transfer learning with private few shot learning and can even yield pure DP guarantees.
    \item We perform extensive empirical evaluation which highlights that \ours yields strong privacy-utility trade-offs, in particular in the high privacy regime and for imbalanced data. 
    \item To further improve DP transfer learning, we show that we can leverage the public data beyond the pre-training step of the feature encoder by privately selecting public prototypes from it.
\end{itemize}

\section{Background} %
\textbf{Transfer Learning.} %
We consider transfer learning where a publicly available pre-trained encoder $\hat{E}$ is used to extract features $\hat{\*X}$ from a (private) dataset $D=(\*X,\*y)$.
Those features are then used to perform downstream classification by learning a function $f$ maximizing $\text{Pr}_{\*x, y\in D}[f(\hat{E}(\*x))=y]$.

\textbf{Differential Privacy.}
Differential privacy (DP)~\citep{dworkCalibratingNoiseSensitivity2006} is a mathematical framework that provides privacy guarantees in ML by formalizing the intuition that a learning algorithm $\mathcal{A}: I \rightarrow S$, executed on two neighboring datasets $D$, $D'$ that differ in only one data point,\ie, $D=D'\cup\{x\}$ (\textit{add/remove DP}), will yield roughly the same output, \ie $\Pr[\mathcal{A}(D)\in S] \leq e^\epsilon \cdot \Pr[\mathcal{A}(D')\in S] + \delta$ (\textit{approximate DP}).  
In this inequality, $\varepsilon$ is the privacy budget that specifies by how much the output is allowed to differ and $\delta$ is the probability of differing more.
If $\delta=0$, we refer to it as \textit{pure DP}, a strictly stronger notion of privacy. We will also refer to zero-concentrated DP (zCDP)~\citep{bunConcentratedDifferentialPrivacy2016}, which requires that $D_\alpha(\mathcal{A}(D)||\mathcal{A}(D'))\leq \xi + \rho\alpha\;\forall_{\alpha\in(1,\infty)}$, where $D_\alpha$ is the Rényi divergence of order $\alpha$. zCDP is a relaxation of pure DP, but stricter than approximate DP. $(0,\rho)$-zCDP can also be expressed simply as $\rho$-zCDP.
We provide more details on DP in \Cref{app:dp}.
The standard approach for learning ML models with DP guarantees is differentially private stochastic gradient descent (DPSGD)~\citep{song2013stochastic, abadiDeepLearningDifferential2016}.
DPSGD clips model gradients to a given norm 
to limit the impact of individual data points on the model updates and adds a controlled amount of Gaussian noise
to implement formal privacy guarantees during training.

\textbf{Exponential Mechanism.}
The exponential mechanism~\citep{mcsherry2007mechanism} offers a way to implement pure DP guarantees. 
Given a set of possible outputs $\*X'$, it samples an output $\*x'$ according to some utility function $u$ with probability $\text{Pr}[\text{EM}_u(\*X)=\hat{\*x}] \propto \exp{\left(\frac{\epsilon}{\Delta u}u(\*X,\hat{\*x})\right)}$. This algorithm satisfies $2\epsilon$-DP.
\Cref{app:exponential} shows more details on the exponential mechanism and utility function.

\textbf{Prototypical Networks.}
Prototypical networks 
\citep{snellPrototypicalNetworksFewshot2017} 
are used for  few-shot classification, \ie they provide a way on adapting a classifier to new unseen classes with access only to a small number of data points from each new class.
Their main components are a set of prototypes $\*p_c \in \mathbb{R}^M$ and an embedding function $f_\phi:\mathbb{R}^D\rightarrow\mathcal{R}^M$. Each prototype for a class $c$ is the mean of the embedded points belonging to that class, \ie $\*p_c=\frac{1}{|\*X_c|}\underset{\*x\in \*X_c}{\sum}f_\phi(\*x)$.
Given a distance function $d: \mathbb{R}^M \times \mathbb{R}^M\rightarrow[0,\infty)$, the model classifies a point $\*x$ based on its nearest prototype in the embedded space as $\hat{y}(\*x)=\underset{c}{\argmin}\;d(f_\phi(\*x),\*p_c)$.

\textbf{DP Mean Estimation.}
Obtaining differentially private means $\mu = 1/n\sum_n \*x$ for $\*x\in \mathbb{R}^d$ is challenging in high dimensions. A straightforward approach \cite{kamathPrimerPrivateStatistics2020} consists of clipping all samples to some $\ell_2$ norm, adding noise scaled according to the clip norm and then reporting the noisy mean of the clipped samples.
\textit{FriendlyCore}~\citep{tsfadiaFriendlyCorePracticalDifferentially2022} is a framework for pre-processing the input data of private algorithms, such that the algorithms being executed on this pre-processed data need to be private only for relaxed conditions. It improves especially for the cases where the samples have a high $\ell_2$ norm and high dimensionality $d$. The \textit{CoinPress} algorithm~\citep{biswasCoinPressPracticalPrivate2020} estimates the mean iteratively, clipping the samples not w.r.t. the origin but to the estimated mean of the previous step. This approach is especially useful when the mean is far away from the origin and generally considered state-of-the-art for dimensionalities in the low thousands. \Cref{fig:coin_vs_naive} shows that the straightforward approach outperforms all other methods given strong priors on the $\ell_2$ norms of the samples. We provide more details in \Cref{app:meanestimation}.

\section{Related Work}

\paragraph{Private Transfer Learning.}
Standard approaches for DP transfer learning rely on the DPSGD algorithm to train a classifier on top of the representations output by a pre-trained encoder, and potentially also to privately update existing or added model parameters on the sensitive data~\citep{yu2021differentially, de2022unlocking, li2021large, mehtaLargeScaleTransfer2022}. Notably, there also exists an approach for transfer learning from few samples, DP-FiLM introduced by \citet{tobabenEfficacyDifferentiallyPrivate2023}.
Such approaches have been shown effective, for loose DP guarantees (\ie large $\varepsilon$), yet suffer from severe utility drops in strong privacy regimes (\ie with small $\varepsilon$).
This is because %
of the iterative nature of the DPSGD algorithm with multiple rounds of noise addition that negatively impact performance.
To overcome these limitations, \citet{mehtaDifferentiallyPrivateImage2023} proposed Differentially Private Least Squares (DP-LS), DP-Newton and DPSGD with Feature Covariance (DP-FC). DP-LS takes advantage of the closed form solution for least squares to avoid running many iterations of gradient descent. DP-Newton employs a second-order optimization to solve the smaller problem of transfer learning more efficiently. DP-FC integrates second order information by utilizing the covariance of the features without paying the composition cost of DP-Newton. All methods have three hyperparameters. 
In contrast to theirs, our method only has a single optional hyperparameter, does not rely on higher order optimization and utilizes parallel composition to solve each class independently in a single iteration, resulting in lower privacy costs especially for imbalanced datasets.

\textbf{Leveraging Public Data for Private Training.}
Public data has, so far, been leveraged for privacy-preserving knowledge transfer to protect sensitive data~\citep{pate_2017,pate_2018}, to reduce the sample complexity within DP distribution learning~\citep{bie2022private,ben2024private}, and for pre-training public encoders to then perform private transfer learning~\cite{li2021large,yu2021differentially,gu2022choosing,tramer2022considerations,ganesh2023public,hu2021lora,houlsby2019parameter,li2023exploring,mehtaDifferentiallyPrivateImage2023}. 
In a similar vein as previous work that determines the importance of public samples to private data \citep{jiDifferentialPrivacyBased2013a}, our approach goes beyond the latter and additionally leverages the public pre-training data of the encoder during the private transfer learning step by selecting public prototypes to represent our private classes.

\textbf{Private Training on Unbalanced Datasets.}
DP has been shown to disproportionately harm utility for underrepresented sub-groups, \ie groups with fewer data points~\citep{bagdasaryanDifferentialPrivacyHas2019,suriyakumar2021chasing}. 
This is because the weak signal from these groups is more affected by the added noise. Additionally, the clipping operation in DPSGD changes the direction of the overall gradient, which adds a compounding bias over the runtime of the training, that disproportionally affects minority classes~\citep{esipova2023disparate}. To mitigate this issue, \citet{esipova2023disparate} propose DPSGD-Global-Adapt, which clips only some gradients and instead scales most gradients, thus preserving the overall direction. The algorithm adaptively learns the clipping threshold, keeping the amount of clipped gradients low. Another approach is to add fairness through in- or post-processing~\cite{jagielskiDifferentiallyPrivateFair2019}, which trades off accuracy against fairness and requires additional privacy budget.
In a non-private setting, solutions for improving utility of small subgroups include changing the sampling \cite{domingosMetaCostGeneralMethod1999, kubat1997addressing, japkowiczClassImbalanceProblem, lewisHeterogeneousUncertaintySampling1994, ling1998data, zadaPureNoiseRescue2022}, generating synthetic data for the minority classes \cite{chawlaSMOTESyntheticMinority2002, wangLowShotLearningImaginary2018a}, or weighing the training loss \cite{caoLearningImbalancedDatasets2019}.  
However, these approaches are not directly compatible with DP or incur additional privacy costs. %

\section{Differentially Private Prototyping }

\textbf{Setup and Assumptions.}
\label{assumptions}
We aim at learning a private classifier based on a sensitive labeled dataset $D=(\*X,\*y)$ with $C$ different classes.
We assume the availability of a standard public pre-trained vision encoder $\hat{M}$, such as DINO\footnote{ \url{https://github.com/facebookresearch/dinov2}} or MAE\footnote{\url{https://github.com/facebookresearch/mae}} encoders, that return high-dimensional feature vectors for their input data points.
Additionally, we assume the availability of a general purpose public dataset $\hat{D}=(\hat{\*X},\dots)$, such as ImageNet~\cite{dengImageNetLargescaleHierarchical2009}. 
Note that $\hat{D}$ can also be from a different distribution than $D$ and $\hat{M}$'s pre-training data, as we show experimentally in \Cref{fig:ablation-cc3m}, and does not require labels. In case of available labels for $\hat{D}$, we just discard them. %

\textbf{Overview.} Our goal is to obtain private prototypes $\*p_1,\dots \*p_C$ that represent every class $C$ from the private dataset $D$ in the embedding space. 
To classify a new unseen data point $\*x'$, we simply have to retrieve the most representative prototype and return its label. 
Concretely, we have to infer $\*x'$ through the encoder $\hat{M}$,  retrieve the prototype with the minimum distance in embedding space to $\*x'$ and return its label as the prediction
$ y' = \min\limits_{c \in C} d(\hat{M}(\*x'),\*p_c)$. 
We detail the general approach in \Cref{fig:our_method}.

Note that if the private prototypes are obtained with DP guarantees, using them for predictions will not incur additional privacy costs due to the DP post-processing guarantees.
Hence, our DP prototypes can be publicly released, similar to privately trained ML models. 
We experimented with multiple ways for implementing DP prototypes and identified the two most promising approaches:
\DPtwo generates a private prototype by calculating a DP mean on all data points of a given class in the embedding space.
Our \DPthree takes advantage of the public dataset $\hat{D}$ and privately selects a data point from $\hat{D}$ to act as a prototype for each private class.

\subsection{\DPtwotext: Private Means}
\label{sub:method_means}

\textbf{Intuition.}
Non-private prototypical networks~\citep{snellPrototypicalNetworksFewshot2017} consist of two steps, namely the training of a projection layer at the output of the encoder and the estimation of the class prototypes. 
In the private setup, both these steps would depend on the private data and therefore each incur additional privacy costs. 
To keep privacy cost low, we forgo projection layer training, as we find it is unnecessary when given a strong pre-trained encoder (see \Cref{app:projection}).
Hence, for our private \DPtwo, we only implement the estimation of the prototypes without projection.

\textbf{Non-Private Means.}
Given a training class $c$ and corresponding samples $\*X_c \in \mathbb R^{n_c\times d}$, the non-private prototype of each class is the mean of the embeddings $\frac 1{n_c}\sum_{i=0}^{n_c}\hat{M}(\*x_i)$

\textbf{Our \DPtwotext: Private Means.}
To privately estimate the means, we rely on the Gaussian Mechanism. We first clip each $\*x_i\in \mathbb R^{d}$ to a $\ell_2$ norm $r$. The estimate is then 
\begin{align}
    \*p_c=\mathcal{N}\left(\*0,\frac{2r^2}{n_c^2\rho}\right) + \frac 1{n_c}\sum_{i=0}^{n_c}\hat{M}(clip_{\ell_2}(\*x_i),r)
\end{align} where $\rho$ is the zCDP privacy budget. To improve the utility at strict privacy budgets, we include a single optional hyperparameter $k_{\text{pool}} \geq 1$, describing the kernel size of an average pooling layer before the mean estimation to reduce dimensionality, reducing the dimension from $d$ to $d/k_{\text{pool}}$.

\textbf{Privacy Analysis.}
The privacy analysis of \DPtwo follows the analysis of the Gaussian Mechanism. By clipping each sample to $\ell_2$ norm of $r$ we obtain $\Delta \*p_c = 2r/n$, since the $\ell_2$ distance between the previous mean and any new sample can be $2r$ at maximum and its influence diminishes with the number of samples $n$. 
We use parallel composition: each disjoint class computes a $\rho$-zCDP mean prototype, making the privacy cost $\rho$ for the entire private dataset.

\subsection{\DPthreetext: Privately Selecting Public Prototypes}
\label{sub:method_public_prototypes}
\textbf{Intuition.}
Our main idea for \DPthree is to leverage public data beyond the pre-training stage for learning a private classifier based on the sensitive data.
Therefore, we privately select public prototypes for each training class, \ie a data point from the public dataset that represent the given class well.

\textbf{Non-Private Selection.}
A good public prototype $\hat{\*x}_c$ for a given training class $c$ represents that class well in the embedding space of encoder $\hat{M}$.
To select such a good prototype per class, we first calculate the embeddings $\*E=\hat{M}(\*X)$ and $\hat{\*E}=\hat{M}(\hat{\*X})$ for the private and public data points, respectively. 
Then, based on the private labels $\*y$, we split the embeddings of $\*X$ in $C$ subsets $\*E_1, \dots, \*E_C$.
Without any privacy considerations, a public prototype $\hat{\*x}_c$ for class $c$ could then be chosen as the data point that minimizes the average distance according to metric $d$, to all training data points $\*x_i$ in class $c$ as
\begin{equation}
    \hat{\*x}_c = \min\limits_{\hat{\*x} \in \hat{\*X}} \frac{\sum_{i=0}^{|\*X_c|}d(\hat{M}(\*x_i), \hat{M}(\hat{\*x}))}{|\*X_c|} \text{.}
\end{equation}

\textbf{Our \DPthreetext: Private Selection.}
\label{sub:top1-method}
The previously described approach, however, does not take any privacy of the training data $D$ into account. 
To perform public prototype selection with $\varepsilon$-DP guarantees, we rely on the exponential mechanism~\cite{mcsherry2007mechanism}.
We use the cosine similarity and add $+1$ as our distance metric $d$, therefore obtaining a bounded and non-negative metric in $[0,2]$.
The utility function that indicates the goodness of each each public sample for a given class $c$ is
\begin{equation}
\label{eq:private_prototype_selection}
    u(\hat{\*x},c) = \sum_{i=0}^{|\*X_c|}1+\frac{\hat{M}(\*x_i)\cdot\hat{M}(\hat{\*x})}{\norm{\hat{M}(\*x_i)}  \norm{\hat{M}(\hat{\*x})}} \text{.}
\end{equation}
To improve utility at strict privacy budgets, we include two optional hyperparameters $d_{\text{max}}\in(0,2]$ and $d_{\text{min}}\in[0,d_{\text{max}})$, which clips the distances to $[d_{\text{min}},d_{\text{max}}]$, reducing the sensitivity to $\Delta u=d_{\text{max}}-d_{\text{min}}$. 

We detail the full algorithm for privately selecting public prototypes in Algorithm \ref{alg:find_prototypes}.
\begin{algorithm2e}[h]
    \caption{Privately Select Public Prototypes}
    \label{alg:find_prototypes}
    \KwIn{Private dataset $D=(\*X,\*y)$ with $C$ classes, privacy budget $\epsilon$, public pre-trained encoder $\hat{M}$, public dataset $\hat{D}=(\hat{\*X},\dots)$,
    hyperparameters $d_{\text{max}}$,$d_{\text{min}}$}
    \KwOut{Prototypes $\*P = \{\*p_c \in \hat{D} | c \in \*y\}$ }
    \SetKwFunction{FMain}{SelectPublicPrototypes()}
    \SetKwProg{Fn}{Function}{:}{}
    \Fn{\FMain}{
    $\*E \gets M(\*X)$ %
    $\hat{\*E} \gets M(\hat{\*X})$ %
     \ForEach{class $c \in C$}{%
    $\*E_c \gets \{\*e_i \in \*E | y_i = c\}$ %
    $u_c(\hat{\*x}_i) = \sum_{\*e\in \*E_c}\text{clip}(1+\frac{\*e\cdot\hat{\*e_i}}{‖\*e‖‖\hat{\*e_i}‖},d_{\text{max}},d_{\text{min}})$; %
     $\*p_c \propto \exp{(\frac{\epsilon u_c}{d_{\text{max}}-d_{\text{min}}})}$ %
        } 
    }
    \KwRet{$\{\*p_c| c \in \*y\}$}\;
\end{algorithm2e}

\textbf{Privacy Analysis.} Our proposed \DPthree fulfills $\varepsilon$-DP. 
We provide a sketch of the full proof from \Cref{app:dp3proof} here.
We first note that $\Delta u =d_{\text{max}}- d_{\text{min}}$. As mentioned above, we add +1 to each cosine similarity, which is therefore non-negative. Therefore, $u$ is positively monotonic w.r.t. $\*X$. The exponential mechanism with $\Pr[\text{EM}(\*X)=\hat{\*x}]\propto \exp{\frac{\varepsilon u(\*X,\hat{\*x})}{\Delta u}}$ is $\varepsilon$-DP if $u$ is monotonic w.r.t. the private data $\*X$ \cite{mcsherry2007mechanism}. Therefore, executing \DPthree on a single class yields $\varepsilon$-DP.
Additionally, since we calculate prototypes per-class and the classes are non-overlapping, parallel composition applies, \ie, the total privacy costs are also $\varepsilon$-DP.

\section{Empirical Evaluation}
\label{sec:experiments}

\textbf{Experiment Setup.}
\label{sub:experiment-setup}
We experiment with CIFAR10~\cite{krizhevskyLearningMultipleLayers2009}, CIFAR100~\cite{krizhevskyLearningMultipleLayers2009}, STL10~\cite{coatesAnalysisSingleLayerNetworks2011} and FOOD101~\cite{bossardFood101MiningDiscriminative2014} as private datasets.
From these datasets, we construct exponentially long-tailed imbalanced datasets with various imbalance ratios (IRs), the ratio between the number of samples in the largest and smallest class, following \citet{cuiClassBalancedLossBased2019} and \citet{caoLearningImbalancedDatasets2019}.
Concretely, the number of samples in each class decreases exponentially with a factor of $n(c)=\exp(-c \lambda)$, where $\lambda=\log(\text{IR})/C$. For the balanced case($\text{IR}=1$), $\lambda=\log(1)/C=0$ and therefore $n(c)=1\,\forall_c$.
We detail the imbalancing process and depict the effect on the resulting absolute class sizes per dataset further in \Cref{app:unbalancedness}. 
We compare our DP prototypes on the features obtained from three vision transformers based on the original architecture from \citet{dosovitskiyImageWorth16x162020} Vit-B-16~\citep{singhRevisitingWeaklySupervised2022}, namely ViT-L-16~\citep{oquabDINOv2LearningRobust2023}, ViT-H-14~\citep{singhRevisitingWeaklySupervised2022} and a ResNet-50~\citep{caronEmergingPropertiesSelfSupervised2021a, heDeepResidualLearning2016}. All models, except for ViT-L-16, which is trained on LVD-142M introduced by \citet{oquabDINOv2LearningRobust2023}, are trained on ImageNet-1K \cite{dengImageNetLargescaleHierarchical2009}.
For \DPthree, we use the $64\times64$ downscaled version of ImageNet-1K upscaled to between $128\times128$ and $512\times512$ depending on the encoder. 
Notably, we evaluate our methods on 
the standard \textit{balanced} test set. This corresponds to reporting a \textit{balanced accuracy} for the imbalanced setups.
A full description of our experimental setup is provided in \Cref{app:experimental_setup}.

\textbf{Baselines.}
For the baseline comparisons we compare to standard linear probing with DPSGD, as it's a common way of DP transfer-learning. Furthermore, we compare to DP-LS from \citet{mehtaDifferentiallyPrivateImage2023} which is the current state-of-the-art for DP transfer learning across all privacy regimes and to DPSGD-Global-Adapt from 
\citet{esipova2023disparate} as it is specifically designed for for training on imbalanced datasets. We outline in \Cref{app:potentialbaselines} the experimentally-supported reasons against including DP-FC and DP-FiLM. 

\textbf{Comparing Results over Different Notions of DP.}
Since we are comparing our new proposed methods that implement \textit{pure DP} or \textit{pure} $\rho$-zCDP guarantees against other baselines that also implement zCDP, we convert all privacy guarantees to $\rho$-zCDP. We also present a pure-DP $\varepsilon$ equivalent by inverting the $\rho=\epsilon^2/2$ conversion from pure DP to zCDP. However, this does not imply that these algorithms fulfil pure DP.
We detail all comparison implementations and conversion theorems used in \Cref{app:dp_conversions}.

\begin{figure}[t]
\centering 
    \includegraphics[width=0.35\textwidth]{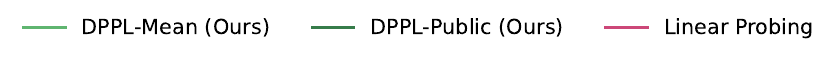}

    \begin{subfigure}[t]{\figwidth}
         \centering
         \includegraphics[width=\textwidth]{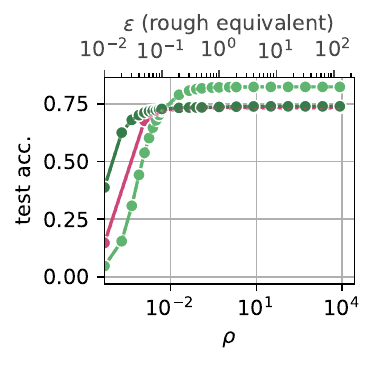}
         \caption{IR=1.}
     \end{subfigure}
    \begin{subfigure}[t]{\figwidth}
         \centering
         \includegraphics[width=\textwidth]{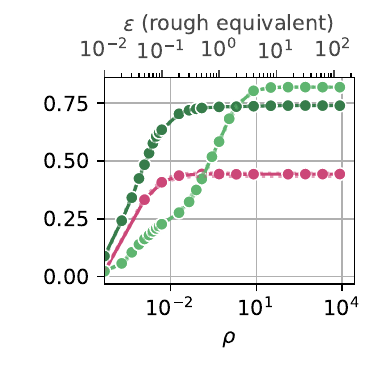}
         \caption{IR=10.}
     \end{subfigure}
     \begin{subfigure}[t]{\figwidth}
         \centering
         \includegraphics[width=\textwidth]{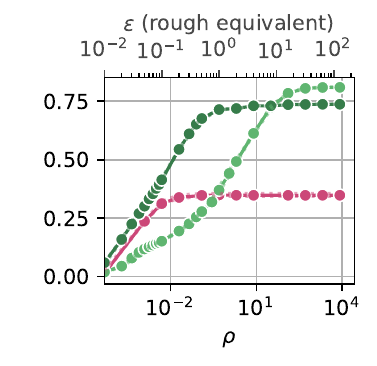}
         \caption{IR=50.}
     \end{subfigure}
     \begin{subfigure}[t]{\figwidth}
         \centering
         \includegraphics[width=\textwidth]{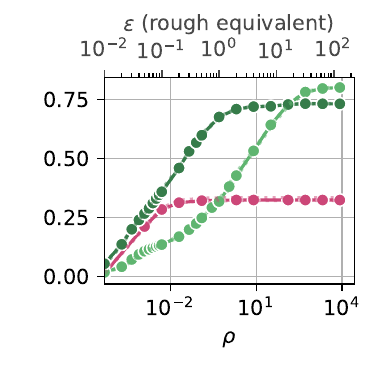}
         \caption{IR=100.}
     \end{subfigure}
\caption{\textbf{DP Prototypes on CIFAR100.}
We present the balanced test accuracy of our methods vs. standard linear probing with DP-SGD on CIFAR100 and ViT-H-14
at different levels of imbalance rations (IR), using ImageNet as public data for \DPthree.
We plot the mean test accuracy over multiple runs and represent the upper and lower quantiles for all methods by the dotted lines.}
\label{fig:cifar100_different_imbalance_vs_linear}
\end{figure}

\subsection{DP Prototypes: High Utility in High Privacy and Extreme Imbalance} %
\label{sub:results}We evaluate our DP prototypes at different privacy regimes in the range corresponding to standard approximate DP~\citep{dworkCalibratingNoiseSensitivity2006} of $0.01<\varepsilon<100$ %
and under different IRs.
In \Cref{fig:cifar100_different_imbalance_vs_linear}, we benchmark our methods vs. standard DP linear probing on CIFAR100 with ViT-H-14. 
Our results highlight that over all levels of IRs larger than 1, our \DPthree significantly outperforms linear probing in all privacy regimes. Additionally \DPthree yields strong performance already at very low $\varepsilon$, such as $\varepsilon=0.1$ for IR=1. 
As data becomes more imbalanced, all methods require larger privacy budgets to yield similarly high performance.
Further, our results indicate that our \DPtwo method underperforms \DPthree and DP linear probing for low $\varepsilon$. We find that this results from the noise added during the mean calculation leading diverging estimations. We provide further detail on the cause in \Cref{app:meanestimation-viz}.
Yet, we observe that at higher epsilon, \DPtwo outperforms \DPthree. This suggests that the most beneficial way for leveraging DP prototypes might be an adaptive method where \DPthree is chosen for high performance in the high privacy regimes and \DPtwo can further boost performance for larger $\varepsilon$.

\begin{figure}[t]
\centering 
    \includegraphics[width=0.35\textwidth]{figures/04_results/single_plots/ours_plus_linear/cifar100/vit_h_14/legend.pdf}
    
    \begin{subfigure}[t]{\figwidth}
         \centering
         \includegraphics[width=\textwidth]{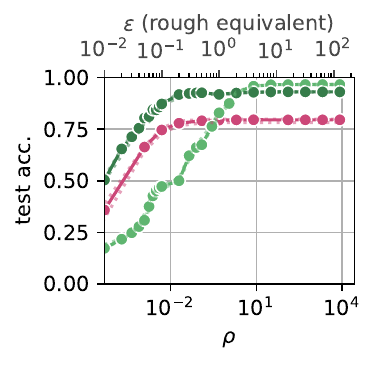}
         \caption{CIFAR10}
     \end{subfigure}
     \begin{subfigure}[t]{\figwidth}
         \centering
         \includegraphics[width=\textwidth]{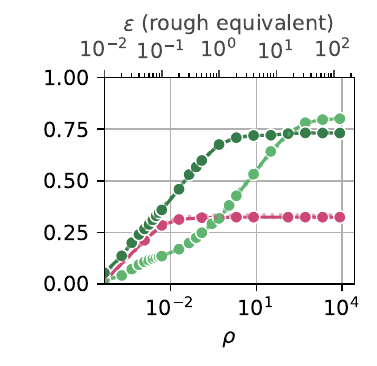}
         \caption{CIFAR100.}
     \end{subfigure}
     \begin{subfigure}[t]{\figwidth}
         \centering
         \includegraphics[width=\textwidth]{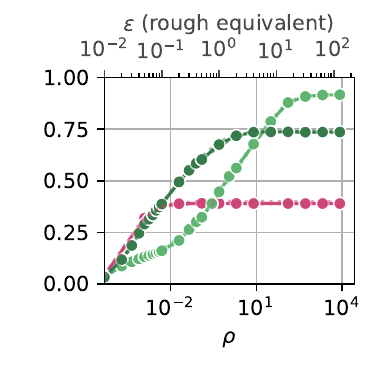}
         \caption{FOOD101}
     \end{subfigure}
     \begin{subfigure}[t]{\figwidth}
         \centering
         \includegraphics[width=\textwidth]{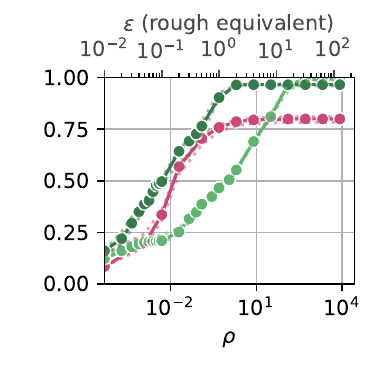}
         \caption{STL10}
     \end{subfigure}
\caption{\textbf{DP Prototypes on various imbalanced datasets.}
We present the balanced test accuracy for CIFAR10, CIFAR100, FOOD101 and STL10 at an imbalance ratio of $100$ on ViT-H-14, using ImageNet as public data for \DPthree.
We compare to standard Linear Probing with DP-SGD. 
We plot the mean test accuracy over multiple runs and represent the upper/lower quantiles by the dotted lines.
\Cref{sub:appendix-imbalanced} shows more results.}
\label{fig:different_dataset_vs_linear}
\vspace{-0.5cm}
\end{figure}

We further assess whether the observed trend holds over different datasets.
Therefore, we depict the results of our methods vs. standard DP linear probing for different datasets and the ViT-H-14 encoder under the most challenging setup with IR$=100$ in \Cref{fig:different_dataset_vs_linear}.
The observed trends are indeed consistent between all datasets.
We provide full results over all datasets and IRs in \Cref{sub:appendix-imbalanced}.

\begin{figure}[t]
    \centering 
    \begin{minipage}{\columnwidth}
        \includegraphics[width=\textwidth]{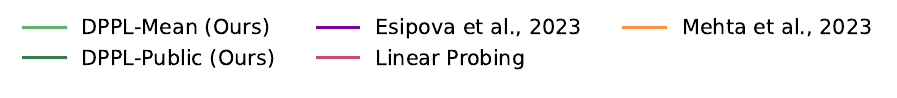}
    \end{minipage}
    \begin{minipage}{\columnwidth}
        \hfill
        \begin{subfigure}[t]{\figwidth}
            \centering
            \includegraphics[width=\textwidth]{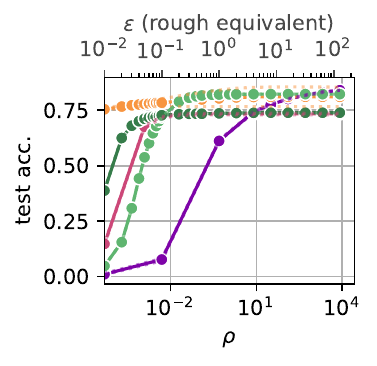}
            \caption{IR=1.}
        \end{subfigure}
        \hfill
        \begin{subfigure}[t]{\figwidth}
            \centering
            \includegraphics[width=\textwidth]{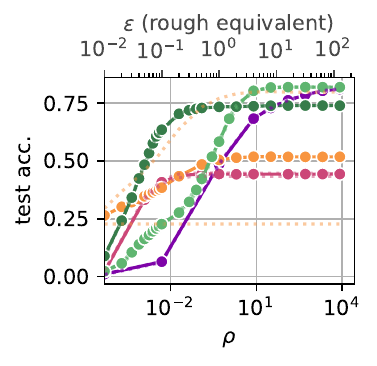}
            \caption{IR=10.}
        \end{subfigure}
        \hfill

        \hfill
        \begin{subfigure}[t]{\figwidth}
            \centering
            \includegraphics[width=\textwidth]{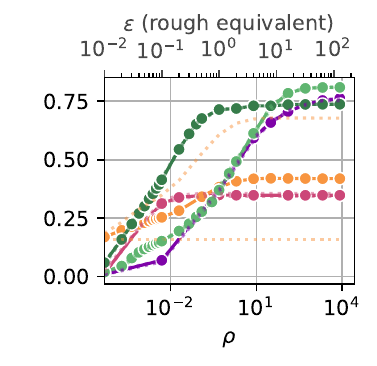}
            \caption{IR=50.}
        \end{subfigure}
        \hfill
        \begin{subfigure}[t]{\figwidth}
            \centering
            \includegraphics[width=\textwidth]{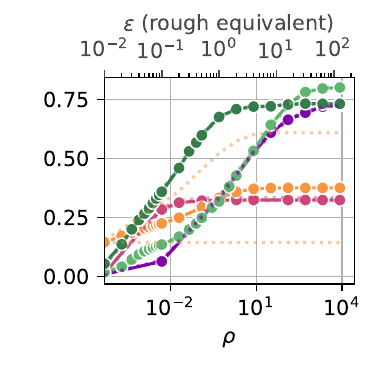}
            \caption{IR=100.}
        \end{subfigure}
        \hfill
    \end{minipage}
    
    \caption{\textbf{Comparing against baselines on CIFAR100.}
    We present the results of our methods vs. state-of-the-art methods (DP-LS and DPSGD-Global-Adapt) on the CIFAR100 dataset using ViT-H-14 under different IRs.
    \DPthree uses ImageNet as public data.
    Dotted lines represent the upper/lower quantiles.
    Similar results for CIFAR10, Food101, and STL10 are presented in Appendix E.1.
    }
    \label{fig:cifar100_baseline}
\end{figure}

\subsection{DP Prototypes Improve over State-of-the-Art Baselines in Imbalanced Setups}
We further benchmark our methods against DP-LS, the current state-of-the-art method for private transfer learning by~\citet{mehtaDifferentiallyPrivateImage2023}, and DPSGD-Global-Adapt by~\citet{esipova2023disparate}, a DP method deliberately designed to achieve high utility under imbalancedness of the private data.
\begin{figure}
\centering

\begin{minipage}{\figwidth}
\includegraphics[width=\textwidth]{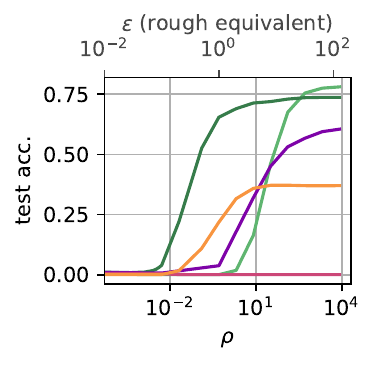}
\end{minipage}
\begin{minipage}{0.38\columnwidth}
\includegraphics[width=\textwidth]{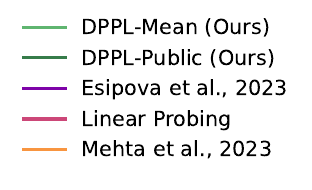}
\end{minipage}

\caption{\textbf{Accuracies of the minority classes.} We depict the test accuracy on CIFAR100 with ViT-H-14 embeddings for the minority classes (smallest $25\%$ of classes) at $\text{IR}=50$.}
\label{fig:small_minority}
\end{figure}
Our results in \Cref{fig:cifar100_baseline} highlight that while in the balanced setup, \citet{mehtaDifferentiallyPrivateImage2023} outperforms the other methods, our DP prototypes outperform all other methods the more imbalanced the setup becomes.
In particular \DPthree outperforms in high privacy regimes (\ie for small $\varepsilon$), while \DPtwo is better in lower privacy regimes, mostly outperforming even DPSGD-Global-Adapt.

The advantage of our methods against the baselines become even more obvious as we do not consider the accuracy over the entire balanced test set (equivalent to balanced accuracy), but look specifically at accuracy on minority classes, see \Cref{fig:small_minority}.
Therefore, we take the smallest $25\%$ of training classes in terms of number of their training data points and measure their accuracy on a balanced test set consisting of only those classes. 
Our results for CIFAR100 on ViT-H-14 under IR$=50$ in \Cref{fig:small_minority} highlight that our DP prototypes significantly outperform all baselines.
Full results for the minority classes are depicted in \Cref{app:minority-acc}.

\subsection{Understanding the Success of DP Prototypes}
To better understand the success of our DP prototypes, we perform various ablations. The full set of ablations and their results is presented in \Cref{app:ablations}. %

\begin{figure}[t]
\centering 
    \includegraphics[width=0.25\textwidth]{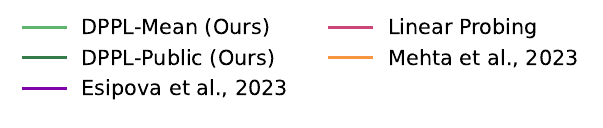}
    
    \begin{subfigure}[t]{\figwidth}
         \centering
         \includegraphics[width=\textwidth]{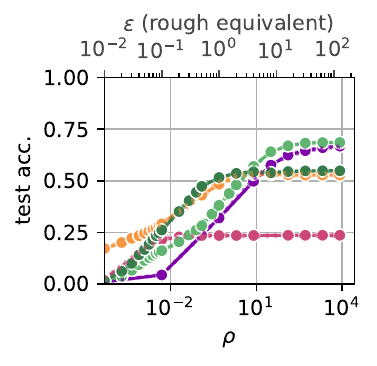}
         \caption{ViT-B-16.}
     \end{subfigure}
     \begin{subfigure}[t]{\figwidth}
         \centering
         \includegraphics[width=\textwidth]{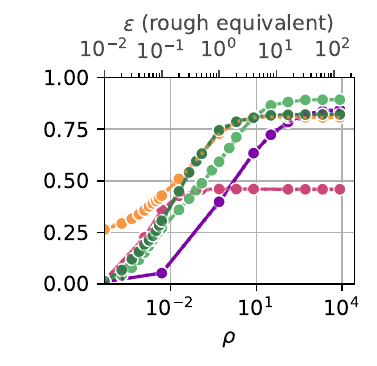}
         \caption{ViT-L-16.}
     \end{subfigure}
     \begin{subfigure}[t]{\figwidth}
         \centering
         \includegraphics[width=\textwidth]{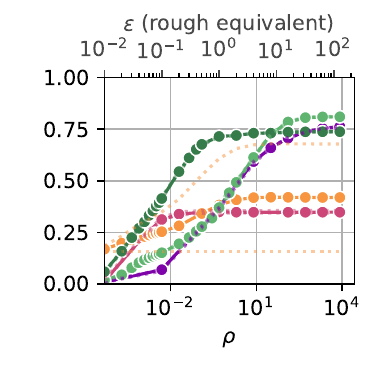}
         \caption{ViT-H-14.}
     \end{subfigure}
     \begin{subfigure}[t]{\figwidth}
         \centering
         \includegraphics[width=\textwidth]{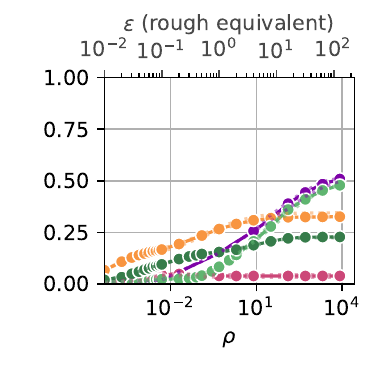}
         \caption{ResNet-50.}
     \end{subfigure}
\caption{\textbf{Choice of Encoder.}
We report the test accuracy for our methods and the baselines for CIFAR100, using ImageNet as public data for \DPthree.
We observe that the success of all methods depends on the quality of the underlying encoder.
}
\vspace{-0.3cm}
\label{fig:ablation-encoder}
\end{figure}
\paragraph{Effect of the Publicly Pre-trained Encoder.}
We first assess the impact of the encoder used as a feature extractor. Therefore, we apply our method and the baselines with different encoder architectures.
Our results in \Cref{fig:ablation-encoder} highlight that the encoder performance impacts all methods alike. 
In particular, we observe that all methods obtain better results with stronger encoders. For example, the ViT-H-14 yields to significantly higher private prediction accuracy that the much smaller ViT-B-16. 
Additionally, none of the methods yields satisfactory results using the ResNet50.

\begin{figure}[t]

    \centering 
        \includegraphics[width=0.6\columnwidth]{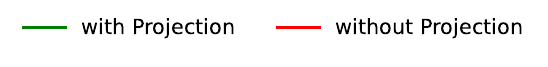}
    
        \begin{subfigure}[t]{\figwidth}
             \centering
             \includegraphics[width=\textwidth]{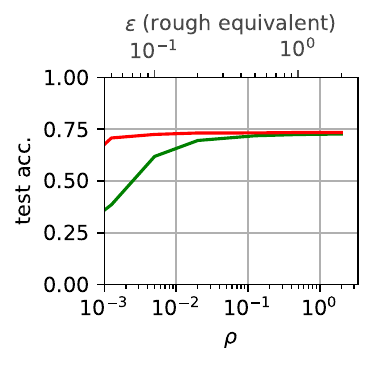}
             \caption{\DPthree.}
         \end{subfigure}
         \begin{subfigure}[t]{\figwidth}
             \centering
             \includegraphics[width=\textwidth]{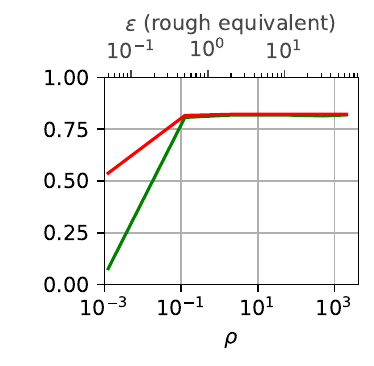}
             \caption{\DPtwo.}
         \end{subfigure}
    \caption{\textbf{Impact of the Projection Layer.}
    }
    \label{fig:ablation-projection}
    
    \end{figure}

\paragraph{Impact of the Projection Layer.} We evaluate whether a projection layer, usually part of a prototypical network, can increase the utility. %
We present our results in \Cref{fig:ablation-projection} for CIFAR100 on ViT-H-14, using ImageNet as public for \DPthree. 
They highlight that \DPthree does not benefit from the projection. Even non-privately ($\epsilon=\infty$), the accuracy of \DPthree with projection is $72.6\%$ and without projection is $74.0\%$, showing that it is not just the decreased privacy budget for the sampling that reduces the utility, but a fundamental misalignment between how the projection is trained and the public prototyping.
We observe the same effect for \DPtwo and conclude that, although this limits adaptability (see \Cref{app:limitations}), with a strong enough pre-trained encoder, it is sufficient for \ours to estimate prototypes without projection.

    \begin{figure}[t]
    \begin{minipage}{\columnwidth}
    \centering 
        \begin{subfigure}[t]{\figwidth}
             \centering
             \includegraphics[width=0.9\textwidth]{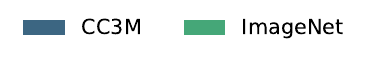}
             \includegraphics[width=\textwidth]{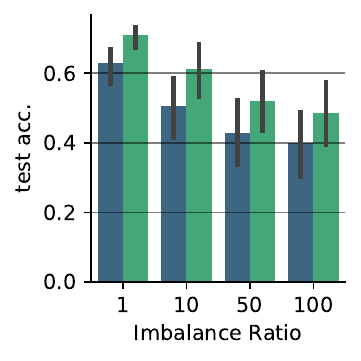}
             \caption{Public Dataset.}
             \label{fig:ablation-cc3m}
         \end{subfigure}
         \begin{subfigure}[t]{\figwidth}
             \centering
             \includegraphics[width=0.65\textwidth]{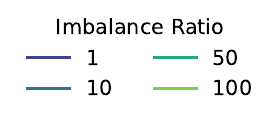}
             \includegraphics[width=\textwidth]{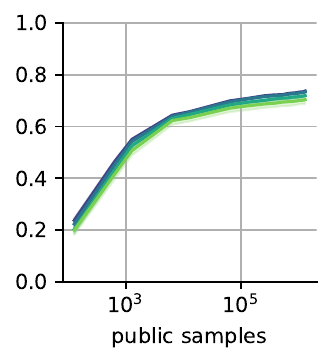}
             \caption{Public Data Size.}
             \label{fig:ablation-public_size}
         \end{subfigure}
    \caption{\textbf{Impact of the Public Data.}
    }
    \end{minipage}
    \end{figure}

\paragraph{Improving through Multiple Per-Class Prototypes.}
We experiment with extending our \DPthree beyond a single per-class prototype---the common standard for prototypical networks.
Therefore, we introduce the variation \DPfour which selects the top-K public prototypes per class.
We extend the algorithm from \citet{gillenwaterJointExponentialMechanism2022} to sample multiple prototypes jointly using the exponential mechanism. 
Then, we classify based on the mean distance to each class's prototype.
Our results in \Cref{fig:ablation-topk} show that \DPfour's privacy-utility trade-offs are between \DPtwo and \DPthree, indicating that the private means can be ---to a certain degree--- approximated by multiple public prototypes. \DPfour could therefore replace \DPtwo in cases of high dimensionality, where a mean estimation is not feasible.
We provide more details, privacy proof, and a full set of results in \Cref{app:topk-method}. 

\paragraph{Impact of the Public Data for Prototype Selection.}
To assess whether the public data for prototype estimation needs to be from the same distribution as the one used to pre-train the encoder, we conduct experiments with a different public dataset. We evaluate \DPthree using 2,298,112 samples from CC3M introduced by \citet{sharmaConceptualCaptionsCleaned2018} as public data instead of ImageNet which is used to pre-train the encoder. We show the relation between the accuracy using ImageNet as public data and CC3M in \Cref{fig:ablation-cc3m} for CIFAR100 on ViT-H-14. %
We find that \DPthree works well with both public datasets, highlighting the flexibility of our approach.
Still, we observe that ImageNet yields better results which suggests that it is particularly beneficial to leverage the public data already available for pre-training also in the private transfer learning step.

\paragraph{Size of the Public Dataset for \DPthreetext.}
We also evaluate the success of \DPthree for different sizes of the public dataset that the public prototypes can be chosen from.
Therefore, we randomly draw subsets of different sizes from ImageNet and apply \DPthree.
Our results for CIFAR100 (100 classes) and ViT-H-14 in \Cref{fig:ablation-public_size} %
indicate that with growing public dataset size, our method's success increases. \Cref{fig:app-publicdatasize} shows that tasks less similar to the pretraining data are more sensitive to dataset size.
Note that even for public dataset of more than one million images, the selection of our public prototypes for 100 classes (\ie the "training") takes 34.3 seconds on a single GPU as we depict in \Cref{tab:runtime}. For 10 classes (\eg CIFAR10), it takes 5 seconds. 
Hence, choosing a larger public dataset does not represent a practical limitation.

\section{Conclusions and Future Work}

We propose \ours as a novel alternative to private fine-tuning with DP.
\ours builds DP prototypes on top of features extracted by a publicly pre-trained encoder, that can be later used as a classifier.
The prototypes can be obtained without iterative noise addition and yield high utility even in high-privacy regimes, with few private training data points, and in unbalanced training setups.
We show that we can further boost performance of our DP prototypes by leveraging the public data beyond training of the encoder and using them to draw the public prototypes from (\DPthree).
Future work at improving utility of high-dimensional DP mean estimation will benefit our \DPtwo, which can, in the future, serve as an additional benchmark for private mean estimation algorithms.

\section*{Acknowledgements}
This research project was supported by the Google Safety Engineering Center. The project was funded by the Initiative and Networking Fund of the Helmholtz Association in the framework of the Helmholtz AI project call under the name "PAFMIM", funding number ZT-I-PF-5-227. Responsibility for the content of this publication lies with the authors. We thank the Helmholtz Information and Data Science Academy (HIDA) for supporting this research through the Helmholtz Visiting Researcher Grant.

\small
\bibliography{fullpaper}
\normalsize

\normalsize
\clearpage
\appendix
\section{Extended Background}
\label{app:background}

\subsection{Differential Privacy}
\label{app:dp}

\begin{definition}[$(\xi, \rho)$-zCDP from \citet{bunConcentratedDifferentialPrivacy2016}]
\label{def:zcdp}
A randomised mechanism $M: \mathcal{X}^n \rightarrow \mathcal{Y}$ is $(\xi, \rho)$-zero-concentrated differentially private (henceforth $(\xi, \rho)$-zCDP) if, for all $x, x^{\prime} \in \mathcal{X}^n$ differing on a single entry and all $\alpha \in(1, \infty)$,
$$
\mathrm{D}_\alpha\left(M(x) \| M\left(x^{\prime}\right)\right) \leq \xi+\rho \alpha,
$$
where $\mathrm{D}_\alpha\left(M(x) \| M\left(x^{\prime}\right)\right)$ is the $\alpha$-Rényi divergence between the distribution of $M(x)$ and the distribution of $M\left(x^{\prime}\right)$.
\end{definition} 
$(0,\rho)$-zCDP can also be expressed simply as $\rho$-zCDP.
\begin{definition}[$(\alpha, \epsilon)$-RDP from \citet{mironovRenyiDifferentialPrivacy2017}]
\label{def:rdp}
    A randomized mechanism $f: \mathcal{D} \mapsto$ $\mathcal{R}$ is said to have $\epsilon$-Rényi differential privacy of order $\alpha$, or $(\alpha, \epsilon)$-RDP for short, if for any adjacent $D, D^{\prime} \in \mathcal{D}$ it holds that
$$
D_\alpha\left(f(D) \| f\left(D^{\prime}\right)\right) \leq \epsilon .
$$
\end{definition}
\begin{definition}[$\mu$-GDP from \citet{dongGaussianDifferentialPrivacy2019}]
\label{def:gdp}
    A randomized mechanism $M: \mathcal{D} \mapsto$ $\mathcal{R}$ is said to have $\mu$-Gaussian differential privacy, $\mu$-GDP for short, if it operates on a statistic $\Theta$ as
$$
M(D)=\Theta(D)+\xi
$$
where $\xi \sim \mathcal N(0, \text{sens}(\theta)^2/\mu^2)$
\end{definition}

\begin{theorem}[Parallel composition from \citet{mcsherryPrivacyIntegratedQueries2009}]
\label{thm:parallel-comp}
Let $M_i$ each provide $\epsilon$-differential privacy. Let $D_i$ be arbitrary disjoint subsets of the input domain $D$. The sequence of $M_i\left(X \cap D_i\right)$ provides $\epsilon$-differential privacy.
\end{theorem}

\subsection{The Gaussian Mechanism}
\begin{proposition}[\citet{cesarBoundingConcentratingTruncating2021}]
\label{prop:gaussian-zcdp}
    Let $q: \mathcal{X}^n \rightarrow \mathbb{R}$ be a sensitivity-$\Delta$ query. Consider the mechanism $M: \mathcal{X}^n \rightarrow \mathbb{R}$ that on input $\*x$, releases a sample from $\mathcal{N}\left(q(\*x), \sigma^2\right)$. Then $M$ satisfies $\left(\Delta^2 / 2 \sigma^2\right)$-zCDP.
\end{proposition}
\subsection{The Exponential Mechanism}
\label{app:exponential}
The exponential mechanism~\citep{mcsherry2007mechanism} aims to give the best output $\hat{\*x}\in \hat{\*X}$ w.r.t. a utility function $u(\*X,\hat{\*x}) : \*X \times \hat{\*X} \rightarrow \mathbb{R}$ where $\*X$ is a private dataset and $\hat{\*X}$ a public dataset. It can be described as a randomized mapping $\text{EM}_u:\*X\rightarrow \hat{\*X}$ where
\begin{align}
    P[\text{EM}_u(\*X)=\hat{\*x}] \propto \exp{\left(\frac{\epsilon}{\Delta u}u(\*X,\hat{\*x})\right)}
\end{align}
\begin{lemma}[\citet{mcsherry2007mechanism}]
    The exponential mechanism is $2\epsilon$-DP.
\end{lemma}
\begin{definition}
A utility function $U(D,\*x)$ is positively (negatively) monotonic if for any point $\*x$ and any datasets $D$ and $D'$, $U(D,\*x) \leq U(D\cup D', \*x)$$ \quad \left(U(D,\*x) \geq U(D\cup D', \*x)\right)$. 
\end{definition}
\begin{lemma}[\citet{mcsherry2007mechanism}]
\label{lem:monotonic-exp}
    Given a monotonic utility function, the exponential mechanism is $\epsilon$-DP.
\end{lemma}
The exponential mechanism fulfils not only differential privacy, but also the stricter bounded range \cite{durfeePracticalDifferentiallyPrivate2019a}. Using a monotonic utility function leads to an improved privacy bound because the sensitivity and range of monotonic functions are equal \cite{dongOptimalDifferentialPrivacy2020a}.
\begin{lemma}[\citet{cesarBoundingConcentratingTruncating2021}]
\label{lem:exp-zcdp}
    The $\epsilon$-DP exponential mechanism is $\epsilon^2/8$-zCDP.
\end{lemma}
\subsection{Private Mean Estimation}
\label{app:meanestimation}
For the mean estimation, we investigated both \textit{CoinPress} and a naïve estimator based on the Gaussian Mechanism. While both provide zCDP, \textit{CoinPress} provides guarantees for a substitute neighborhood and the Gaussian Mechanism for a add/remove neighborhood, meaning they are not compared under the same guarantees. 
\subsubsection{CoinPress}
The \textit{CoinPress} algorithm \cite{dongOptimalDifferentialPrivacy2020a} aims to privately estimate the mean $\*\mu = 1/n\sum_n \*x$ for some private $\*x\in \mathbb R ^d$.
Each step is initiated with a center $\*c_i$ and radius $r_i$ with $||\mu-c_i||_2 \leq r_i$. Commonly used for $(r_0,c_0)$ are $(\sqrt{d},\*0)$. All points further away from $\*c_i$ than $r_i+\gamma$, where $\gamma$ is chosen s.t. $\Pr[||\mathcal{N}(\*0,d)||_2 < \gamma] \geq 0.99$ are $\ell_2$-clipped to $r_i+\gamma$. Finally, Gaussian noise is added to all points. Then $\*c_{i+1}$ is the mean of the noised and clipped points and $r_{i+1}$ defined through the new Gaussian tailbounds of the points.
\subsubsection{Naïve}
We formulate the mean estimation problem as a private query $q(\*X) : \mathbb{R}^{n\times d} \rightarrow \mathbb{R}^d$ that we want to release from the private database $\*X$ where
\begin{align}
    q(\*X)=\frac{1}{\norm{\*X}}\sum_{\*x\in\*X}\*x
\end{align}
Without further bounds, this query cannot satisfy Differential Privacy. We therefore clip all $\*x\in\*X$ to some $\ell_2$ norm $r$ to obtain $\overline{\*X} = \{clip_{\ell_2}(\*x,r) | \*x \in \*X\}$ . From this we obtain 
\begin{align}
    \Delta q(\overline{\*X}) = \frac{2r}{n}.
\end{align}
and using \Cref{prop:gaussian-zcdp} a $\rho$-zCDP mean estimation query $q_\rho$ as
\begin{align}
    q_\rho(\*X)=q(\overline{\*X}) + \mathcal{N}(\*0,{2r^2}/({n^2\rho})).
\end{align}
To finally obtain a mean $\*p_c$ for each class $c\in C$ while fulfilling $\rho$-zCDP with respect to the entirety of $\*X$, we utilize parallel composition (\Cref{thm:parallel-comp}) over the disjoint class subsets $\*X_c$.

\section{Extended Experimental Setup}
\label{app:experimental_setup}

\subsection{Computational Resources and Libraries}
\label{app:misc}
Our implementation is in Jax \cite{jax2018github} v.0.4.31 with CUDA12. Private linear probing was conducted with PyTorch \cite{paszkePyTorchImperativeStyle2019a} v.2.3.1 and made private using the Opacus \cite{opacus} v.1.5.2 privacy engine. We relied on Optuna \cite{akibaOptunaNextgenerationHyperparameter2019a} v.3.6.1 using the Tree-structured Parzen Estimator \cite{bergstraAlgorithmsHyperParameterOptimization2011} for all algorithmic hyperparameter optimizations. Converting and visualizing privacy guarantees was done in part with AutoDP \cite{wangSubsampledRenyiDifferential2019, zhuPoissionSubsampledRenyi2019} v.0.2.3.1. Scaling the experiments has been aided by Ray \cite{moritzRayDistributedFramework2018a} v.2.23.0 and Dask \cite{rocklinDaskParallelComputation2015} v.2024.8.0. Configurations were handled by Hydra \cite{Yadan2019Hydra} v.1.3.2. The experiments were conducted using NVIDIA A100 GPUs and an AMD EPYC 7742 64-Core Processor with 1TB of RAM on Ubuntu 22.04 in Python 3.11. In total, obtaining all results required approximately $300$ GPU hours, resulting in roughly $105$ kWh of electric energy usage.

\subsection{Imbalanced Datasets}
\label{app:unbalancedness}
\begin{figure}[t]
\centering 
    \includegraphics[width=0.25\textwidth]{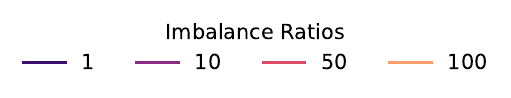}

     \begin{subfigure}[t]{0.15\textwidth}
         \centering
         \includegraphics[width=\textwidth]{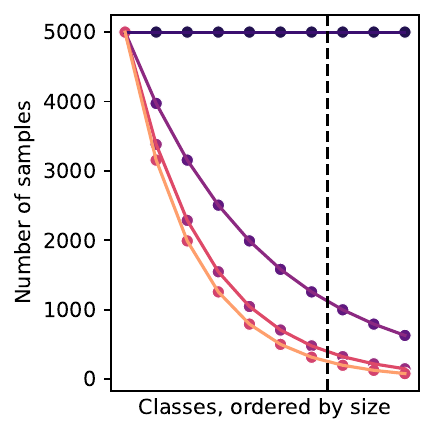}
         \caption{All IRs.}
     \end{subfigure}
     \hfill
    \begin{subfigure}[t]{0.15\textwidth}
         \centering
         \includegraphics[width=\textwidth]{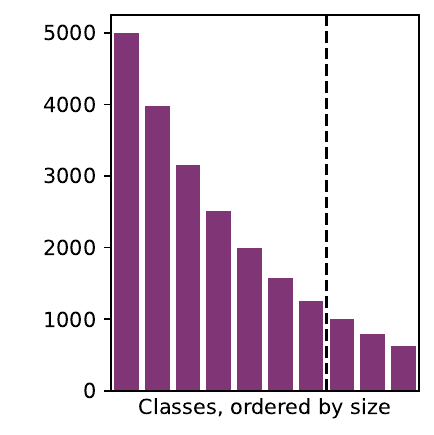}
         \caption{IR=10.}
     \end{subfigure}
     \hfill
    \begin{subfigure}[t]{0.15\textwidth}
         \centering
         \includegraphics[width=\textwidth]{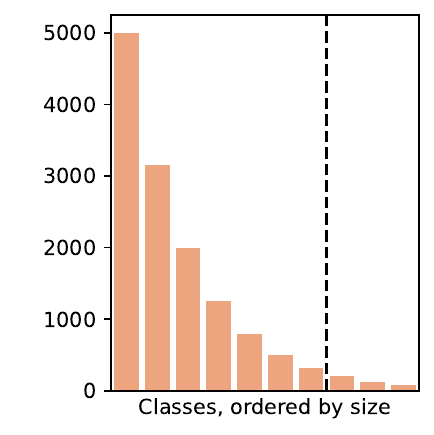}
         \caption{IR=100.}
     \end{subfigure}
\caption{\textbf{Visualizing the effect of the imbalance on CIFAR10.} 
We order the classes by the number of corresponding samples and plot them. The classes right of the dotted line are considered to be the minority classes.}
\label{fig:imbalanced-datasets}
\end{figure}

We present the distribution of the number of data points per class in \Cref{tab:imbalanced-samples} and \Cref{fig:imbalanced-datasets}
\begin{table}[t]
\centering
\label{tab:imbalanced-samples}
\resizebox{\columnwidth}{!}{%
\begin{tabular}{l|clllllll}
\multirow{2}{*}{Dataset} & \multicolumn{8}{c}{Imbalance Ratio}                                                                                    \\
                         & \multicolumn{2}{c|}{1}    & \multicolumn{2}{c|}{10}         & \multicolumn{2}{c|}{50}        & \multicolumn{2}{c}{100} \\ \hline
 & \multicolumn{1}{l}{Median} & \multicolumn{1}{l|}{Min} & Median & \multicolumn{1}{l|}{Min} & Median & \multicolumn{1}{l|}{Min} & Median & Min \\ \hline
CIFAR10                  & \multicolumn{2}{c|}{5000} & 1594 & \multicolumn{1}{l|}{500} & 724 & \multicolumn{1}{l|}{100} & 517         & 50        \\
CIFAR100                 & \multicolumn{2}{c|}{500}  & 158  & \multicolumn{1}{l|}{50}  & 71  & \multicolumn{1}{l|}{10}  & 50          & 5         \\
FOOD101                  & \multicolumn{2}{c|}{750}  & 237  & \multicolumn{1}{l|}{75}  & 106 & \multicolumn{1}{l|}{15}  & 75          & 8         \\
STL10                    & \multicolumn{2}{c|}{500}  & 160  & \multicolumn{1}{l|}{50}  & 73  & \multicolumn{1}{l|}{10}  & 52          & 5         \\
FLOWERS101               & \multicolumn{2}{c|}{1}    & 3    & \multicolumn{1}{l|}{1}   & 7   & \multicolumn{1}{l|}{1}   & 10          & 1        
\end{tabular}%
}
\caption{\textbf{Number of data points over different rations of imbalance.} Median and average number of samples per class for each dataset and imbalance ratio. We construct the imbalanced datasets as described in \citet{cuiClassBalancedLossBased2019} and subsequently \citet{caoLearningImbalancedDatasets2019} (Exponential Long-Tailed).}
\end{table}

\section{Ablations}
\label{app:ablations}
\subsection{Public Dataset Size}
\begin{figure}[t]
    
    \centering 
    \includegraphics[width=0.6\columnwidth]{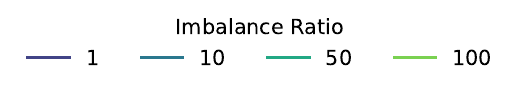}

    \begin{subfigure}[t]{0.35\columnwidth}
         \centering
         \includegraphics[width=\textwidth]{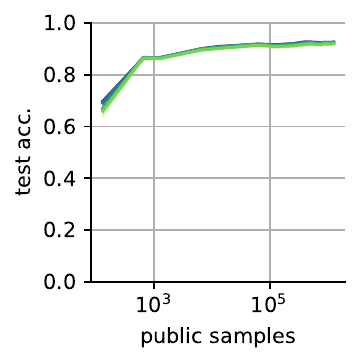}
         \caption{CIFAR10.}
     \end{subfigure}
    \begin{subfigure}[t]{0.35\columnwidth}
         \centering
         \includegraphics[width=\textwidth]{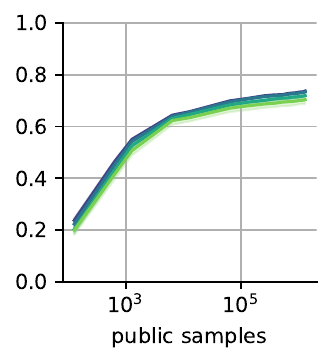}
         \caption{CIFAR100.}
     \end{subfigure}
     \begin{subfigure}[t]{0.35\columnwidth}
         \centering
         \includegraphics[width=\textwidth]{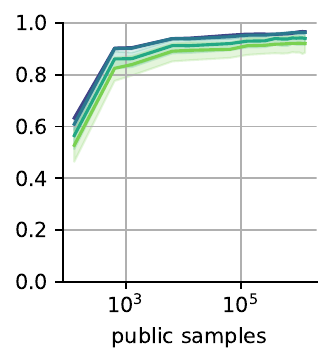}
         \caption{STL10.}
     \end{subfigure}
     \begin{subfigure}[t]{0.35\columnwidth}
         \centering
         \includegraphics[width=\textwidth]{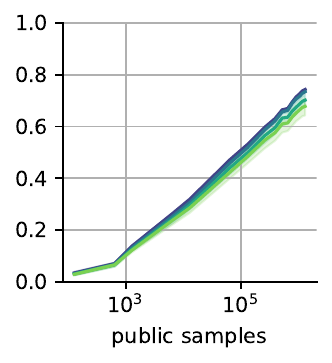}
         \caption{FOOD101.}
     \end{subfigure}
     
     \caption{\textbf{Varying public dataset sizes for \DPthreetext}.
     We randomly subsample the public dataset, limiting \DPthree's prototype selection to fewer samples, and evaluate the resulting changes in accuracy. Tasks with less similarity to the pretraining dataset, \eg FOOD101, are more sensitive to dataset size.}
\label{fig:app-publicdatasize}
\end{figure}

The public data used for \DPthree, ImageNet, consists of 1,281,167 samples. We evalute the impact of using a smaller public dataset by randomly subsampling. \Cref{fig:app-publicdatasize} shows the resulting accuracy. The accuracy and amount of public samples are positively correlated in all cases, but for some datasets, \ie CIFAR10, STL10, the accuracy seemingly asymptotically approaches a maximum accuracy. For FOOD101 it seems ImageNet is not large enough. We note that the imbalance ratio doesn't change the amount of public data required. 

\subsection{Comparing Different Mean Estimations}
\label{app:meanestimation-viz}
\begin{figure}[t]
    \centering
    \begin{minipage}{0.52\columnwidth}
        \includegraphics[width=\textwidth]{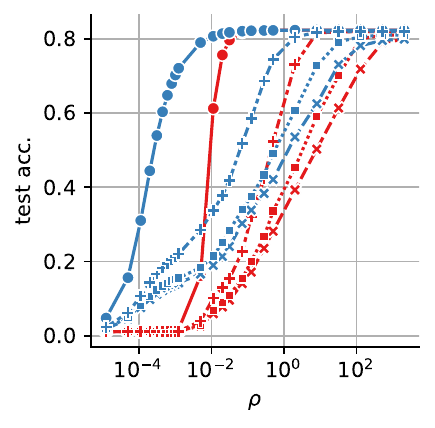}
    \end{minipage}
    \begin{minipage}{0.32\columnwidth}
        \includegraphics[width=\textwidth]{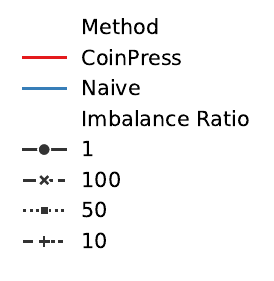}
    \end{minipage}

    \caption{\textbf{Comparing \textit{CoinPress} and naïve mean estimation results} for CIFAR100 and ViT-H-14 embeddings.}
    \label{fig:coin_vs_naive}
\end{figure}

We find that the naïve estimator outperforms \textit{CoinPress} given the strong priors on the $\ell_2$ norms of the embeddings and the fact that they are generally close to the origin. \Cref{fig:coin_vs_naive} compares the accuracy from both mean estimation methods.

\begin{figure}[t]
\centering 
     \begin{subfigure}[t]{0.4\columnwidth}
         \centering
         \includegraphics[width=\textwidth]{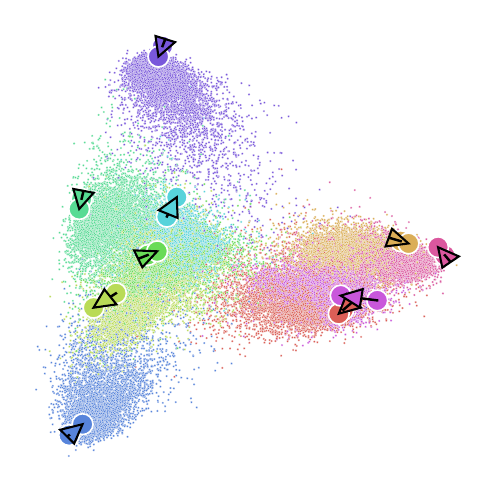}
         \caption{$\rho=10^{-3}$: True vs. DP means.}
     \end{subfigure}
     \hfill
    \begin{subfigure}[t]{0.4\columnwidth}
         \centering
         \includegraphics[width=\textwidth]{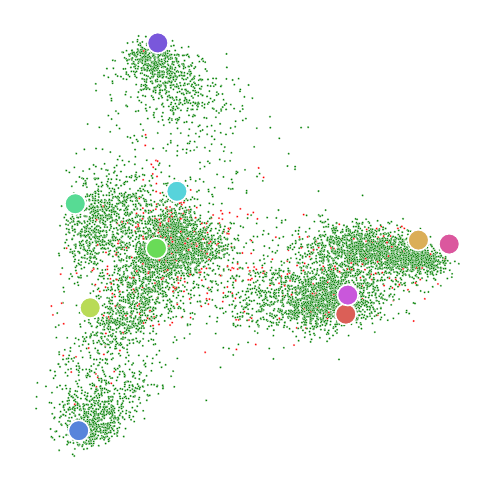}
         \caption{$\rho=10^{-3}$: Classification results.}
     \end{subfigure}

    \begin{subfigure}[t]{0.4\columnwidth}
         \centering
         \includegraphics[width=\textwidth]{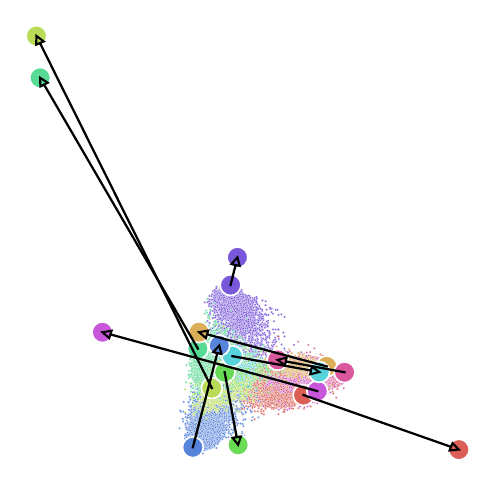}
         \caption{$\rho=10^{-4}$: True vs. DP means.}
     \end{subfigure}
     \hfill
    \begin{subfigure}[t]{0.4\columnwidth}
         \centering
         \includegraphics[width=\textwidth]{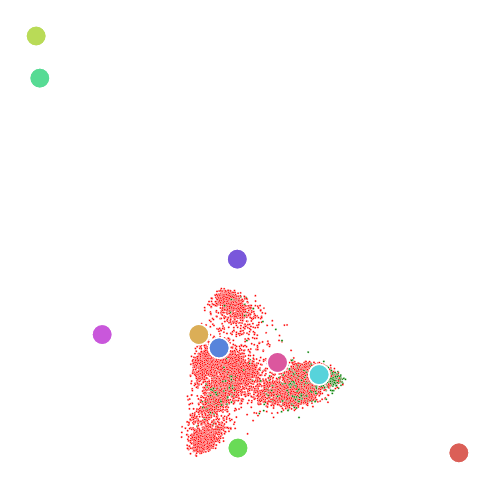}
         \caption{$\rho=10^{-4}$: Classification results.}
     \end{subfigure}
     \hfill
\caption{\textbf{Visualizing the mean estimation on CIFAR10.} 
We estimate the means using \textit{CoinPress}~\cite{biswasCoinPressPracticalPrivate2020} at $\rho\in\{10^{-3},10^{-4}\}$. On the left, we show the non-private means and connect them with arrows to the privately estimated ones on top of the train set. Colors indicate the classes. On the right, we show the privately estimated means and the test set. Green points represent correctly classified samples, red points misclassified ones.}
\label{fig:meanestimation-viz}
\end{figure}

\begin{figure}[t]
\centering 
\includegraphics[width=0.6\columnwidth]{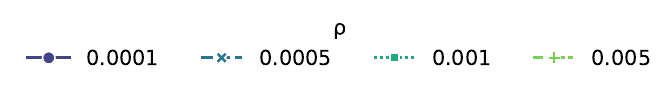}

\begin{subfigure}[t]{0.48\columnwidth}
     \centering
     \includegraphics[width=\textwidth]{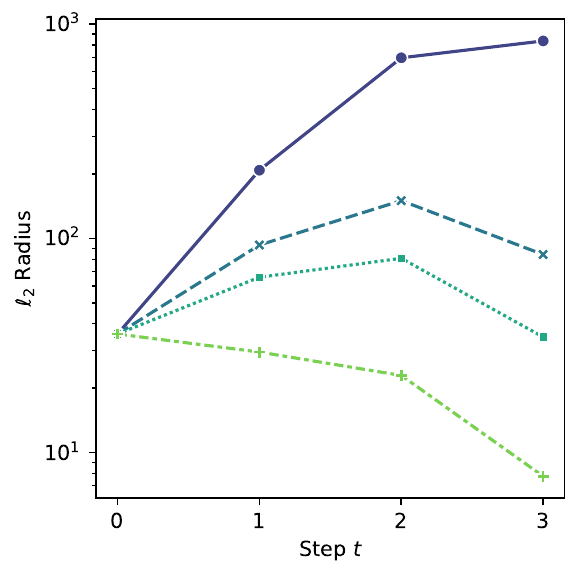}
     \caption{Radius.}
 \end{subfigure}
\begin{subfigure}[t]{0.48\columnwidth}
     \centering
     \includegraphics[width=\textwidth]{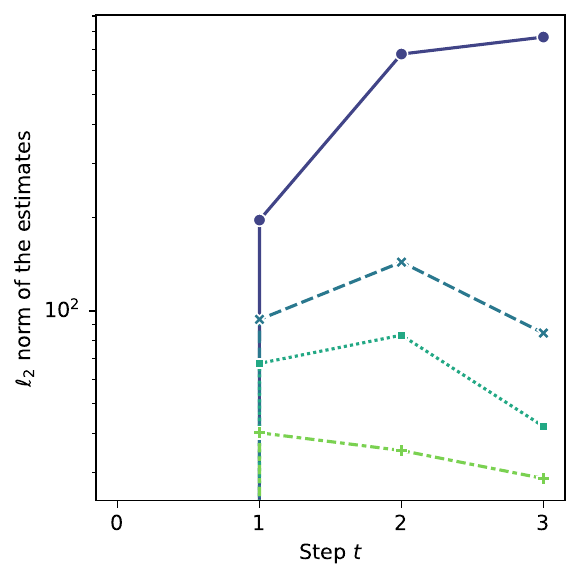}
     \caption{Norm of the estimation.}
 \end{subfigure}

\caption{\textbf{Analyzing the steps of \textit{CoinPress}.} 
We estimate the means using \textit{CoinPress}~\cite{biswasCoinPressPracticalPrivate2020} for different $\rho$. We see that for low values of $\rho$, the radius and the estimates diverge.}
\label{fig:radius-viz}
\end{figure}

\Cref{fig:meanestimation-viz} shows how the \textit{CoinPress} private mean estimation behaves for reasonable and too low privacy levels. For too low privacy values, the mean estimation breaks down. We identify the underlying cause as the divergence of the bounding radius and visualize it in \Cref{fig:radius-viz}. Each successive radius is supposed to be decreasing in size, successively bounding the estimated mean to a smaller space. This is achieved by taking the mean of clipped and noised samples. The clipping decreases the average norm and thus reduces the radius. For very low privacy budgets, the noising of the samples outweighs this effect, and the norms instead grow with each step, leading to a diverging radius. As we take the mean of increasingly diverging samples, the estimates of the mean diverge.

While the naïve estimator also diverges at low privacy budgets, we find the minimum privacy budget required to be lower compared to \textit{CoinPress}.

\subsection{Top-K Public Prototyping}
\label{app:topk-method}

\begin{figure}
\centering
\begin{minipage}{0.2\textwidth}
    \centering
    \includegraphics[width=\textwidth]{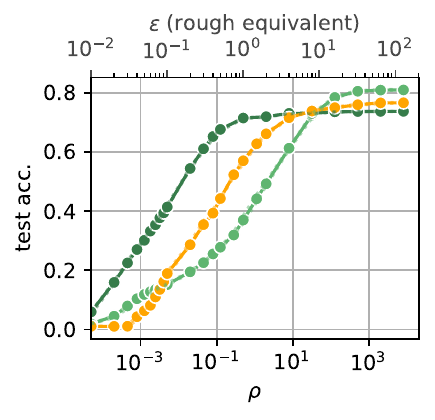}
\end{minipage}
\begin{minipage}{0.15\textwidth}
    \centering
    \includegraphics[width=\textwidth]{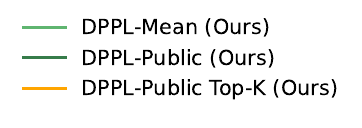}
\end{minipage}
\caption{\textbf{Top-K}.}
\label{fig:ablation-topk}
\end{figure}

Prototypical Networks have been extended to two prototypes per class, leading to increased generalization and robustness \cite{songDualPrototypicalNetwork2022}. We generalize this concept to $K$ prototypes per class. We propose our Differentially Private Unordered Top-$K$ Selection as an adaption of the algorithm from \citet{gillenwaterJointExponentialMechanism2022} to sample these multiple prototypes jointly using the exponential mechanism.

\subsubsection{Differentially Private Unordered Top-$K$ Selection}
Let $(u_1, \dots, u_n)$ be the utilities of the public samples in decreasing order and $K$ the number of prototypes to select. Since the order of the prototypes is not important in this context, we define the utility $U(\*X,S)$ of a set of prototypes $S=\{u_{S_1}, \cdots, u_{S_K}\}$ w.r.t. the private datasets $\*X$ as
\begin{align}
      U(\*X,S) = \begin{cases}-u_K+\min _{k \in[K]}u_{s_k} & \text { if } s_1, \ldots, s_K \\ & \text { are distinct. } \\ -\infty & \text { otherwise }\end{cases}  
\end{align}
\begin{lemma}
\label{lem:sensitivity-U}
    $\Delta U = \Delta u$.
\end{lemma}
\textit{Proof.} The choice of $-\infty$ for repeating sequences does not depend on the private data and therefore doesn't affect the sensitivity. Furthermore, the utility of a set is only dependent on the lowest utility in the set $u_{\min{}}$ and the $K$th true best utility $u_K$. The utility of a set can thus be formulated as $U(\*X,S)=u_K-u_{\min}$. $u$ is monotonic and has sensitivity $\Delta u$, in other words, insertion of a private sample can only increase each utility $u$ by a maximum of $\Delta u$. It follows that insertion or removal of a private sample can only change $U$ by $\pm\Delta u$, \ie $\Delta U = \Delta u$

Therefore, each set has the utility of its worst entry, unless two entries repeat, in which case the utility is $-\infty$ and such set therefore never selected. If we select the true $K$-best prototypes, the utility is $0$ and otherwise it's negative. Each utility is not unique. Instead, a utility can occur as many times as the number of possible combinations of samples with a higher utility. Given a utility $u_y$, we can therefore obtain the number of possible sets with that utility 
\begin{align}
    m(u_y) = {y \choose K}
\end{align}
The entire algorithm then consists of
\begin{enumerate}
    \item privately sampling a utility $u_y$ with $P[EM(x)=y] \propto {y \choose K}\exp{\frac{\epsilon u_y}{2\Delta U}}$,
    \item fixing the corresponding $\hat{\*x}_y$ as part of the set, and
    \item uniformly sampling the remaining $K-1$ prototypes without replacement, s.t. $\{\hat{\*x}_i | u_i \geq u_y\}$.
\end{enumerate}
Note that while \Cref{lem:sensitivity-U} implies the sensitivity of $u$ and $U$ are the same, our effective privacy costs still double, since $U$ is no longer monotonic. We perform the sampling using Proposition 5 from \citet{medinaDuffDatasetDistanceBasedUtility2021}.

\subsection{Classification with Multiple Prototypes}
Given $K$ multiple prototypes $\*P_c\in\mathcal{R}^{K\times d}$ for each class $c \in [C]$, we classify $\*x$ based on the minimum average distance
\begin{equation}
    f(\*x) = \underset{c}{\arg \min} \frac{1}{K}\sum_{i=1}^K d(\*x,\*p_{c,k})\text{.}
\end{equation}

\begin{figure}[t]
\centering 
    \includegraphics[width=0.7\columnwidth]{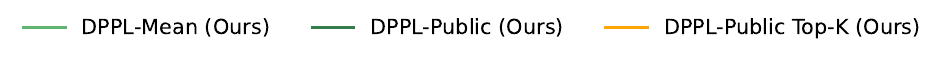}

    \makebox[20pt]{\raisebox{20pt}{\rotatebox[origin=c]{90}{ViT-B-16}}}%
    \begin{subfigure}[t]{0.22\columnwidth}
         \centering
         \includegraphics[width=\textwidth]{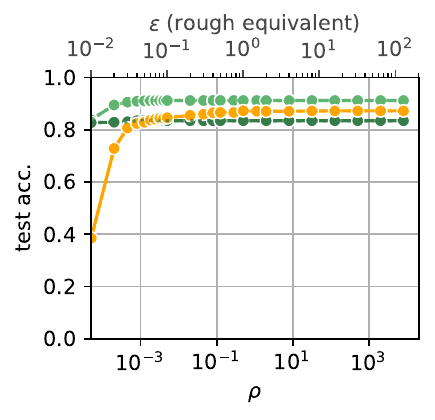}
     \end{subfigure}
     \hfill
    \begin{subfigure}[t]{0.22\columnwidth}
         \centering
         \includegraphics[width=\textwidth]{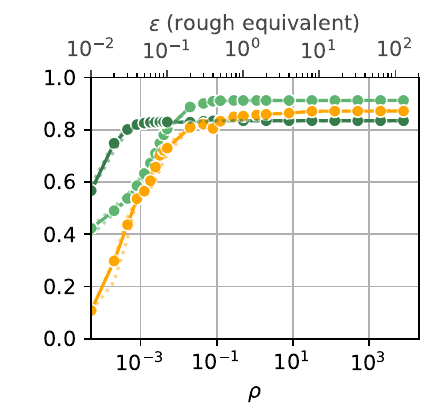}
     \end{subfigure}
     \hfill
     \begin{subfigure}[t]{0.22\columnwidth}
         \centering
         \includegraphics[width=\textwidth]{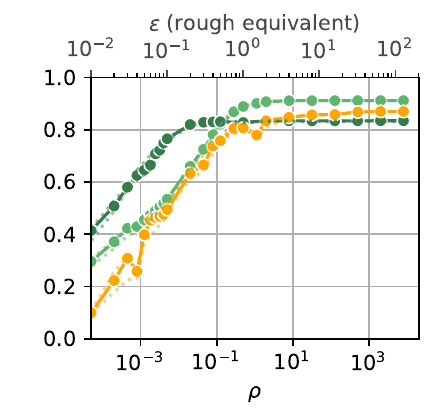}
     \end{subfigure}
     \hfill
     \begin{subfigure}[t]{0.22\columnwidth}
         \centering
         \includegraphics[width=\textwidth]{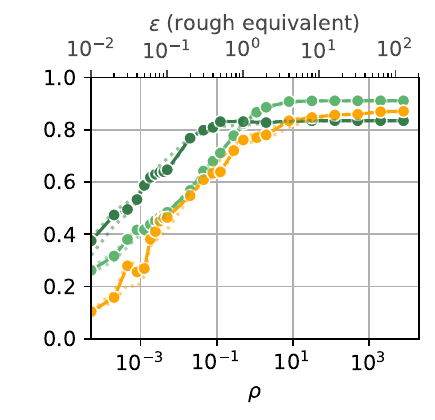}
     \end{subfigure}

    \makebox[20pt]{\raisebox{20pt}{\rotatebox[origin=c]{90}{ViT-L-16}}}%
    \begin{subfigure}[t]{0.22\columnwidth}
         \centering
         \includegraphics[width=\textwidth]{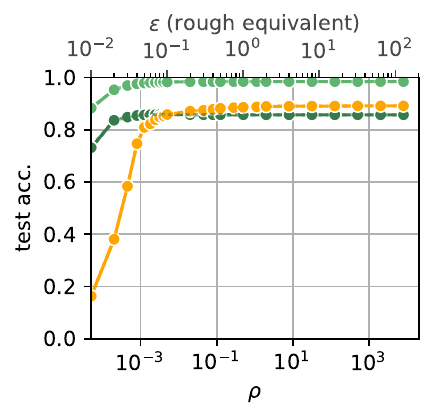}
     \end{subfigure}
     \hfill
     \begin{subfigure}[t]{0.22\columnwidth}
         \centering
         \includegraphics[width=\textwidth]{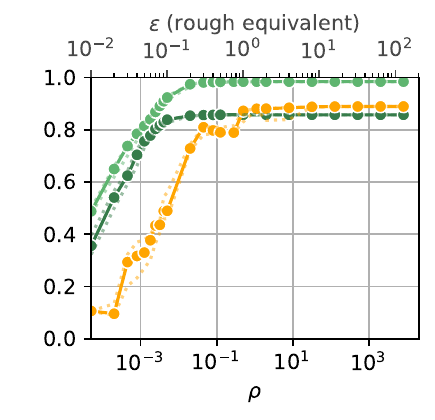}
     \end{subfigure}
     \hfill
     \begin{subfigure}[t]{0.22\columnwidth}
         \centering
         \includegraphics[width=\textwidth]{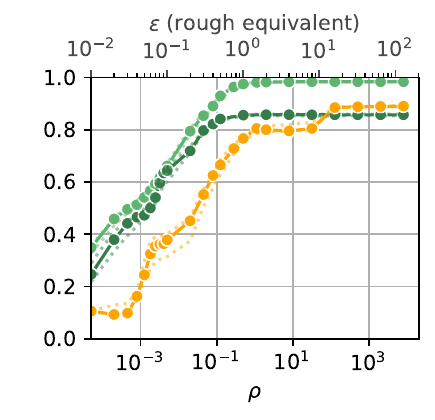}
     \end{subfigure}
     \hfill
     \begin{subfigure}[t]{0.22\columnwidth}
         \centering
         \includegraphics[width=\textwidth]{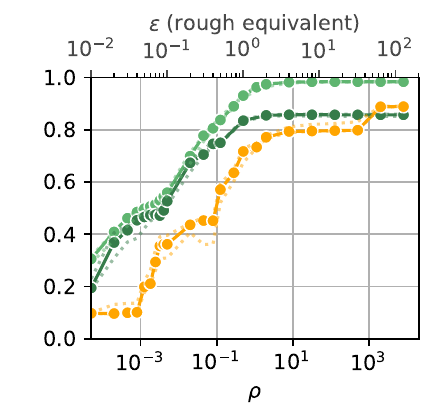}
     \end{subfigure}

    \makebox[20pt]{\raisebox{20pt}{\rotatebox[origin=c]{90}{ViT-H-14}}}%
     \begin{subfigure}[t]{0.22\columnwidth}
     \centering
     \includegraphics[width=\textwidth]{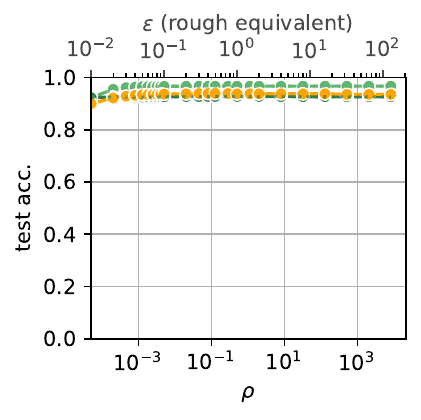}
     \end{subfigure}
     \hfill
    \begin{subfigure}[t]{0.22\columnwidth}
         \centering
         \includegraphics[width=\textwidth]{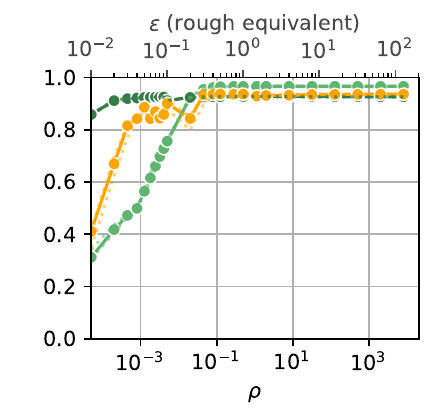}
     \end{subfigure}
     \hfill
     \begin{subfigure}[t]{0.22\columnwidth}
         \centering
         \includegraphics[width=\textwidth]{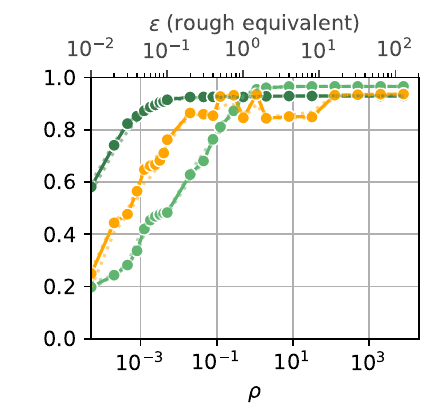}
     \end{subfigure}
     \hfill
     \begin{subfigure}[t]{0.22\columnwidth}
         \centering
         \includegraphics[width=\textwidth]{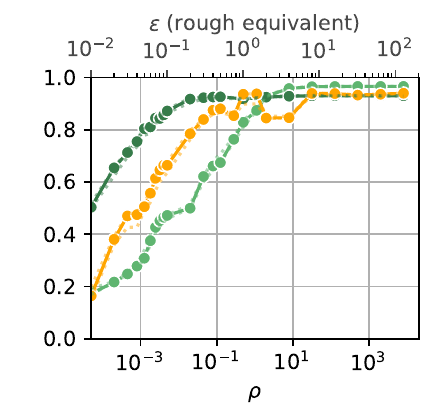}
     \end{subfigure}

    \makebox[20pt]{\raisebox{20pt}{\rotatebox[origin=c]{90}{ResNet-50}}}%
    \begin{subfigure}[t]{0.22\columnwidth}
         \centering
         \includegraphics[width=\textwidth]{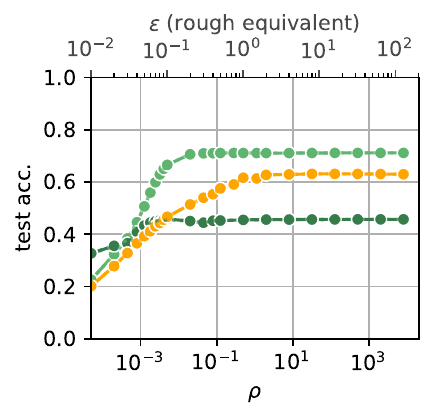}
         \caption{IR=1.}
     \end{subfigure}
     \hfill
     \begin{subfigure}[t]{0.22\columnwidth}
         \centering
         \includegraphics[width=\textwidth]{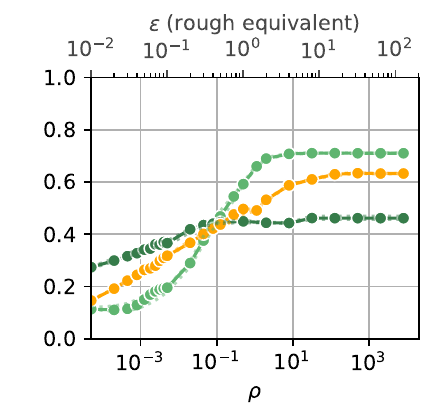}
         \caption{IR=10.}
     \end{subfigure}
     \hfill
     \begin{subfigure}[t]{0.22\columnwidth}
         \centering
         \includegraphics[width=\textwidth]{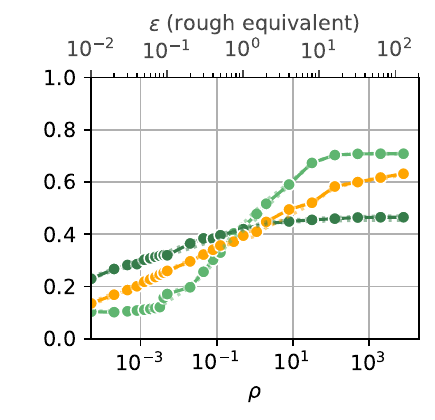}
         \caption{IR=50.}
     \end{subfigure}
     \hfill
     \begin{subfigure}[t]{0.22\columnwidth}
         \centering
         \includegraphics[width=\textwidth]{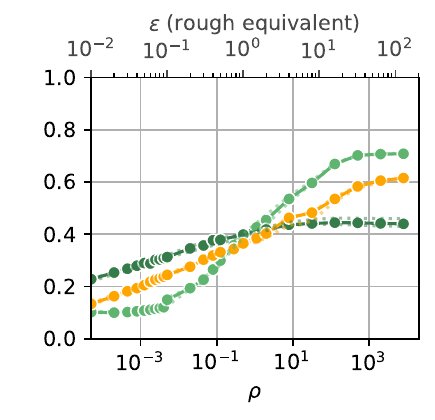}
         \caption{IR=100.}
     \end{subfigure}
\caption{\textbf{\DPthreetext \, with Top-$K$.}
We present the results including \DPthree Top-$K$ of CIFAR10 on ViT-B-16, ViT-H-14, ViT-L-16 and ResNet-50, using ImageNet as public data for \DPthree, at different levels of imbalance rations (IR).
}
\label{fig:app-topk-cifar10}
\end{figure}

\begin{figure}[t]
\centering 
    \includegraphics[width=0.7\columnwidth]{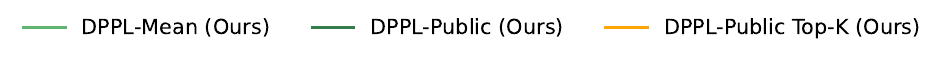}

    \makebox[20pt]{\raisebox{20pt}{\rotatebox[origin=c]{90}{ViT-B-16}}}%
    \begin{subfigure}[t]{0.22\columnwidth}
         \centering
         \includegraphics[width=\textwidth]{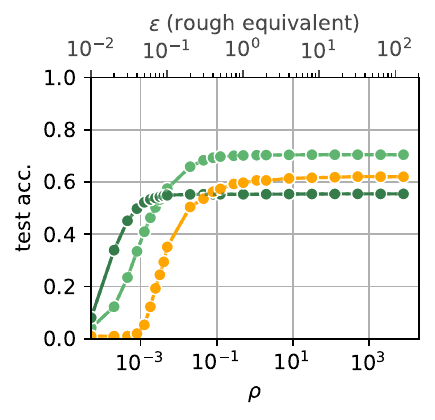}
     \end{subfigure}
     \hfill
    \begin{subfigure}[t]{0.22\columnwidth}
         \centering
         \includegraphics[width=\textwidth]{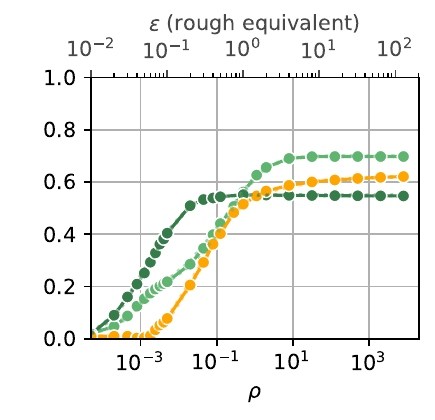}
     \end{subfigure}
     \hfill
     \begin{subfigure}[t]{0.22\columnwidth}
         \centering
         \includegraphics[width=\textwidth]{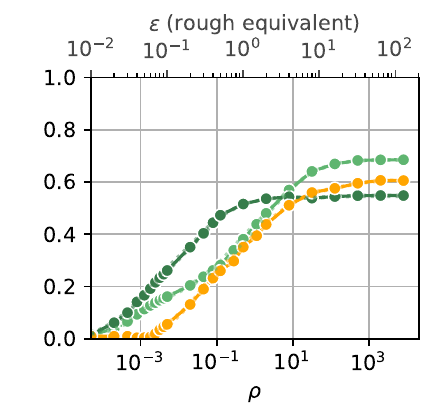}
     \end{subfigure}
     \hfill
     \begin{subfigure}[t]{0.22\columnwidth}
         \centering
         \includegraphics[width=\textwidth]{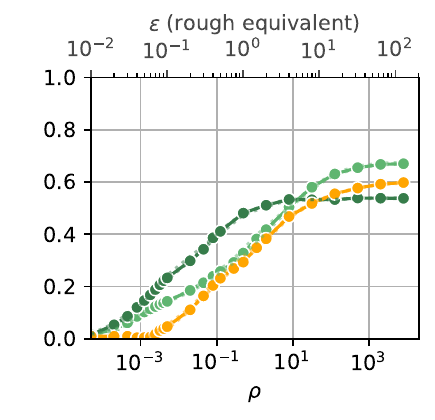}
     \end{subfigure}

    \makebox[20pt]{\raisebox{20pt}{\rotatebox[origin=c]{90}{ViT-L-16}}}%
    \begin{subfigure}[t]{0.22\columnwidth}
         \centering
         \includegraphics[width=\textwidth]{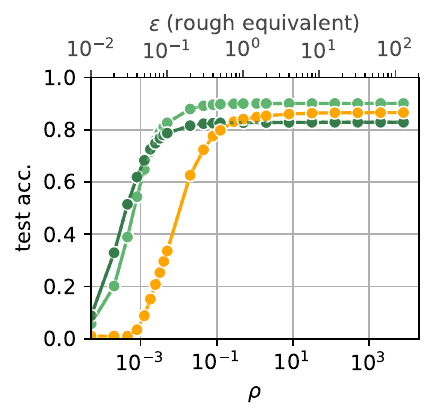}
     \end{subfigure}
     \hfill
     \begin{subfigure}[t]{0.22\columnwidth}
         \centering
         \includegraphics[width=\textwidth]{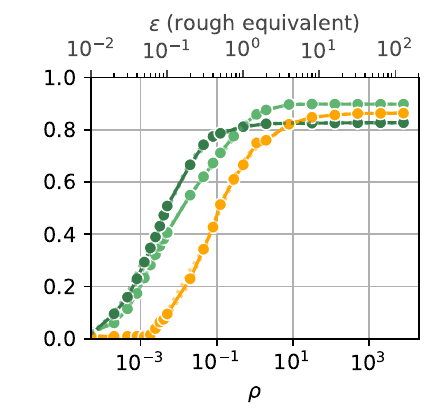}
     \end{subfigure}
     \hfill
     \begin{subfigure}[t]{0.22\columnwidth}
         \centering
         \includegraphics[width=\textwidth]{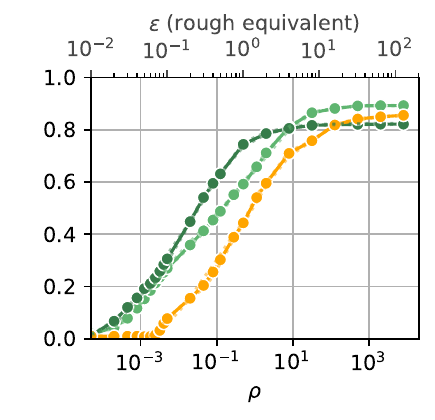}
     \end{subfigure}
     \hfill
     \begin{subfigure}[t]{0.22\columnwidth}
         \centering
         \includegraphics[width=\textwidth]{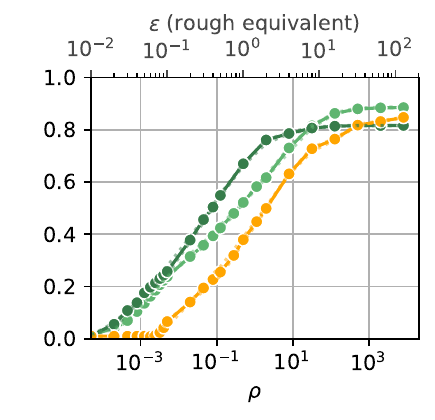}
     \end{subfigure}

    \makebox[20pt]{\raisebox{20pt}{\rotatebox[origin=c]{90}{ViT-H-14}}}%
     \begin{subfigure}[t]{0.22\columnwidth}
     \centering
     \includegraphics[width=\textwidth]{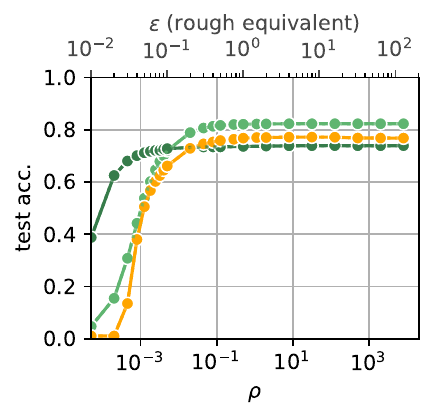}
     \end{subfigure}
     \hfill
    \begin{subfigure}[t]{0.22\columnwidth}
         \centering
         \includegraphics[width=\textwidth]{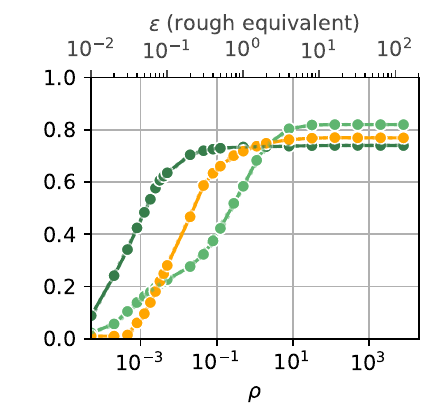}
     \end{subfigure}
     \hfill
     \begin{subfigure}[t]{0.22\columnwidth}
         \centering
         \includegraphics[width=\textwidth]{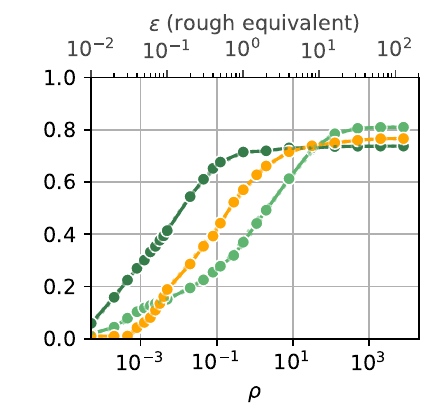}
     \end{subfigure}
     \hfill
     \begin{subfigure}[t]{0.22\columnwidth}
         \centering
         \includegraphics[width=\textwidth]{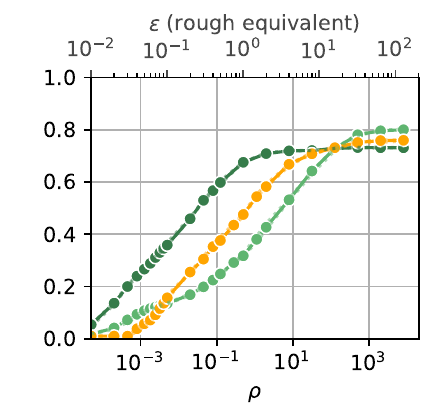}
     \end{subfigure}

    \makebox[20pt]{\raisebox{20pt}{\rotatebox[origin=c]{90}{ResNet-50}}}%
    \begin{subfigure}[t]{0.22\columnwidth}
         \centering
         \includegraphics[width=\textwidth]{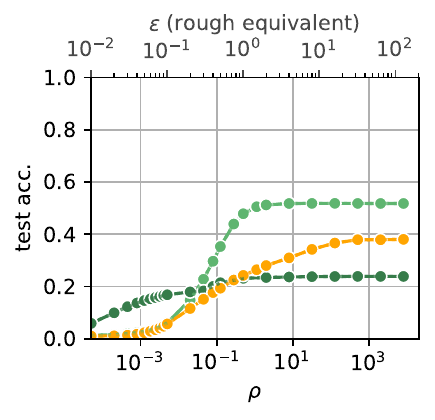}
         \caption{IR=1.}
     \end{subfigure}
     \hfill
     \begin{subfigure}[t]{0.22\columnwidth}
         \centering
         \includegraphics[width=\textwidth]{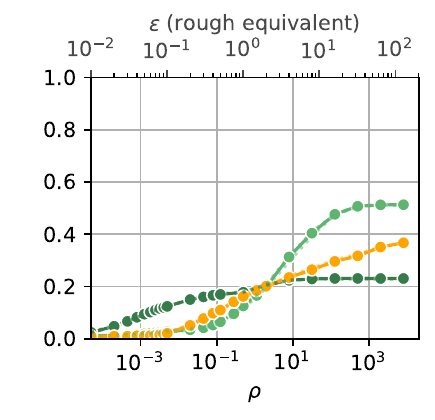}
         \caption{IR=10.}
     \end{subfigure}
     \hfill
     \begin{subfigure}[t]{0.22\columnwidth}
         \centering
         \includegraphics[width=\textwidth]{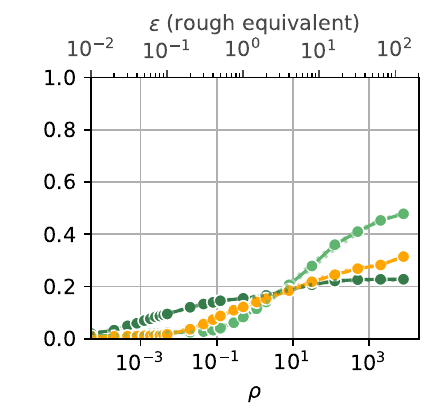}
         \caption{IR=50.}
     \end{subfigure}
     \hfill
     \begin{subfigure}[t]{0.22\columnwidth}
         \centering
         \includegraphics[width=\textwidth]{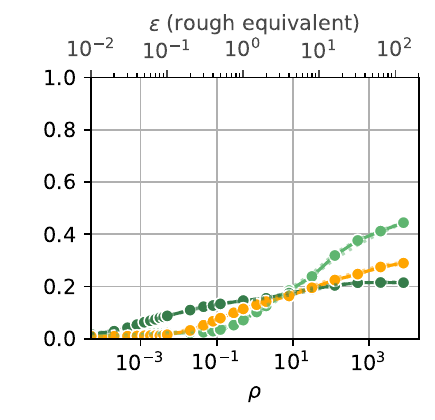}
         \caption{IR=100.}
     \end{subfigure}
\caption{\textbf{\DPthreetext \, with Top-$K$.}
We present the results including \DPthree Top-$K$ of CIFAR100 on ViT-B-16, ViT-H-14, ViT-L-16 and ResNet-50, using ImageNet as public data for \DPthree, at different levels of imbalance rations (IR).
}
\label{fig:app-topk-cifar100}
\end{figure}

\begin{figure}[t]
\centering 
    \includegraphics[width=0.7\columnwidth]{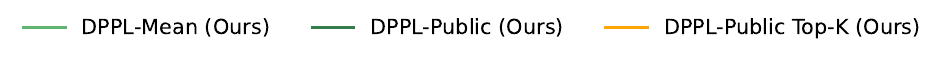}

    \makebox[20pt]{\raisebox{20pt}{\rotatebox[origin=c]{90}{ViT-B-16}}}%
    \begin{subfigure}[t]{0.22\columnwidth}
         \centering
         \includegraphics[width=\textwidth]{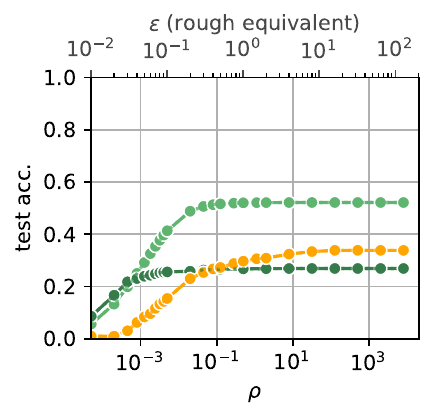}
     \end{subfigure}
     \hfill
    \begin{subfigure}[t]{0.22\columnwidth}
         \centering
         \includegraphics[width=\textwidth]{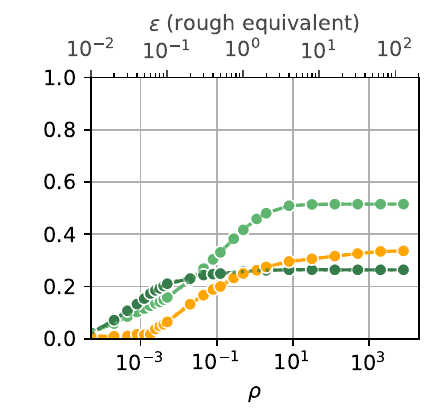}
     \end{subfigure}
     \hfill
     \begin{subfigure}[t]{0.22\columnwidth}
         \centering
         \includegraphics[width=\textwidth]{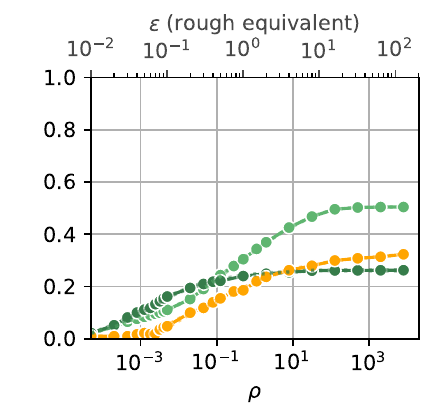}
     \end{subfigure}
     \hfill
     \begin{subfigure}[t]{0.22\columnwidth}
         \centering
         \includegraphics[width=\textwidth]{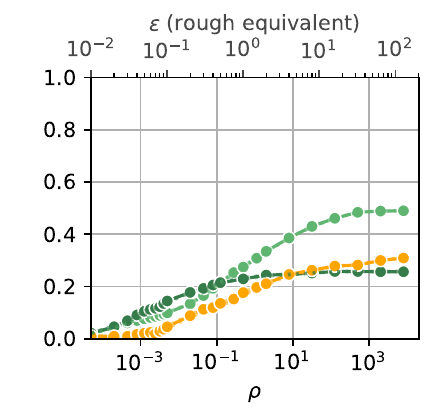}
     \end{subfigure}

    \makebox[20pt]{\raisebox{20pt}{\rotatebox[origin=c]{90}{ViT-L-16}}}%
    \begin{subfigure}[t]{0.22\columnwidth}
         \centering
         \includegraphics[width=\textwidth]{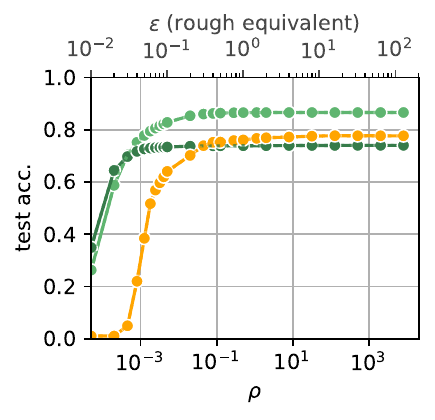}
     \end{subfigure}
     \hfill
     \begin{subfigure}[t]{0.22\columnwidth}
         \centering
         \includegraphics[width=\textwidth]{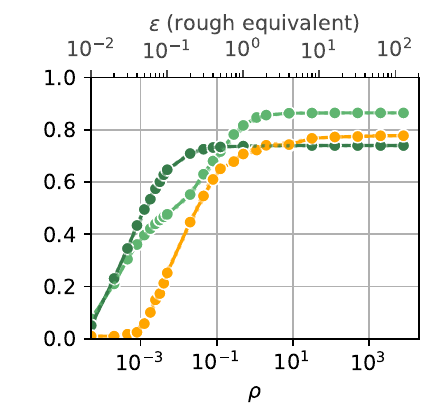}
     \end{subfigure}
     \hfill
     \begin{subfigure}[t]{0.22\columnwidth}
         \centering
         \includegraphics[width=\textwidth]{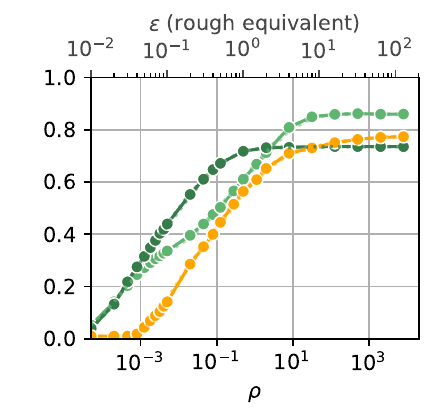}
     \end{subfigure}
     \hfill
     \begin{subfigure}[t]{0.22\columnwidth}
         \centering
         \includegraphics[width=\textwidth]{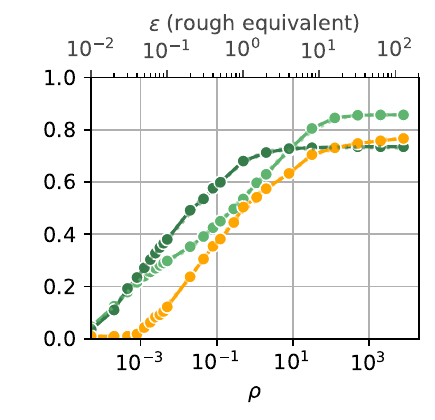}
     \end{subfigure}

    \makebox[20pt]{\raisebox{20pt}{\rotatebox[origin=c]{90}{ViT-H-14}}}%
     \begin{subfigure}[t]{0.22\columnwidth}
     \centering
     \includegraphics[width=\textwidth]{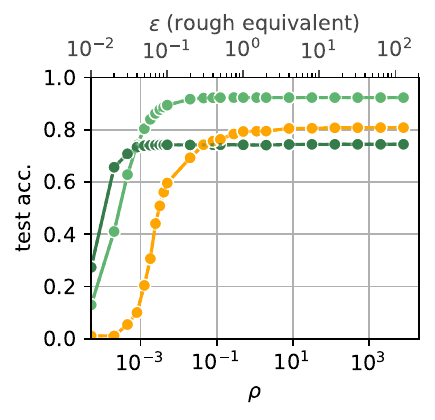}
     \end{subfigure}
     \hfill
    \begin{subfigure}[t]{0.22\columnwidth}
         \centering
         \includegraphics[width=\textwidth]{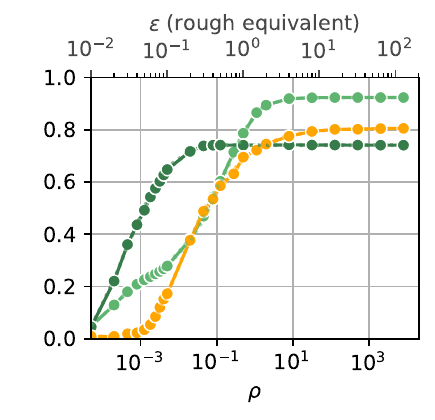}
     \end{subfigure}
     \hfill
     \begin{subfigure}[t]{0.22\columnwidth}
         \centering
         \includegraphics[width=\textwidth]{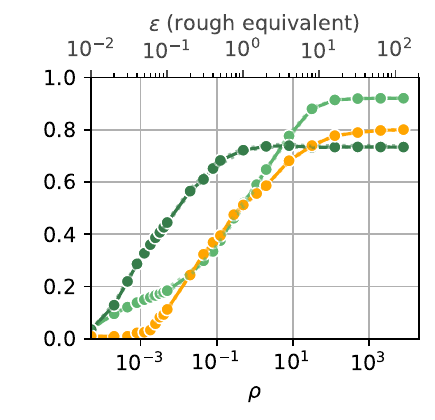}
     \end{subfigure}
     \hfill
     \begin{subfigure}[t]{0.22\columnwidth}
         \centering
         \includegraphics[width=\textwidth]{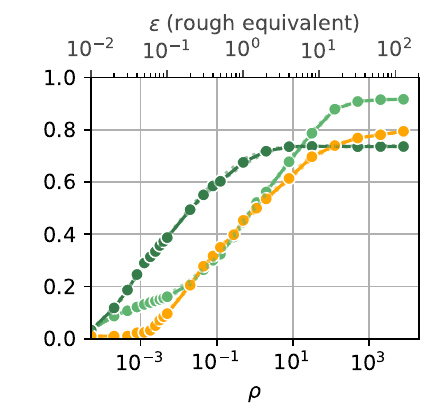}
     \end{subfigure}

    \makebox[20pt]{\raisebox{20pt}{\rotatebox[origin=c]{90}{ResNet-50}}}%
    \begin{subfigure}[t]{0.22\columnwidth}
         \centering
         \includegraphics[width=\textwidth]{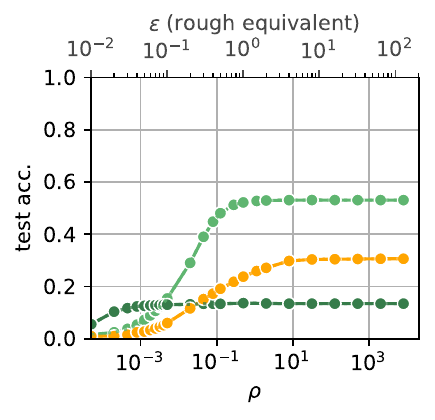}
         \caption{IR=1.}
     \end{subfigure}
     \hfill
     \begin{subfigure}[t]{0.22\columnwidth}
         \centering
         \includegraphics[width=\textwidth]{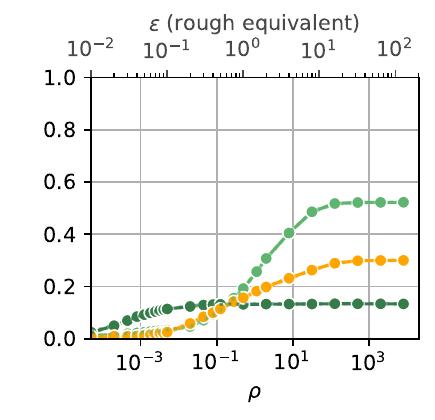}
         \caption{IR=10.}
     \end{subfigure}
     \hfill
     \begin{subfigure}[t]{0.22\columnwidth}
         \centering
         \includegraphics[width=\textwidth]{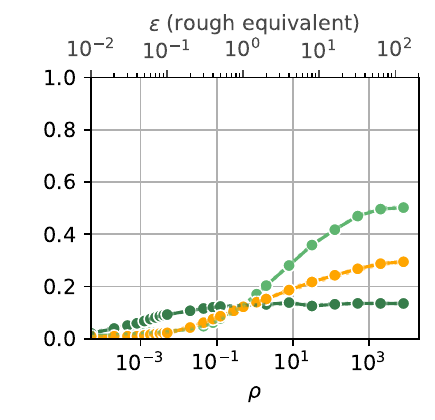}
         \caption{IR=50.}
     \end{subfigure}
     \hfill
     \begin{subfigure}[t]{0.22\columnwidth}
         \centering
         \includegraphics[width=\textwidth]{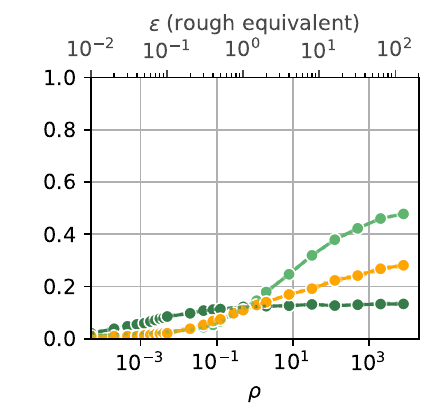}
         \caption{IR=100.}
     \end{subfigure}
\caption{\textbf{\DPthreetext \, with Top-$K$.}
We present the results including \DPthree Top-$K$ of FOOD101 on ViT-B-16, ViT-H-14, ViT-L-16 and ResNet-50, using ImageNet as public data for \DPthree, at different levels of imbalance rations (IR).
}
\label{fig:app-topk-food101}
\end{figure}

\begin{figure}[t]
    \centering 
        \includegraphics[width=0.7\columnwidth]{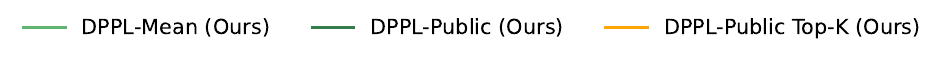}
    
        \makebox[20pt]{\raisebox{20pt}{\rotatebox[origin=c]{90}{ViT-B-16}}}%
        \begin{subfigure}[t]{0.22\columnwidth}
             \centering
             \includegraphics[width=\textwidth]{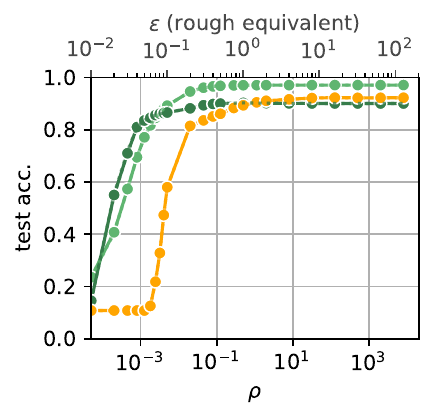}
         \end{subfigure}
         \hfill
        \begin{subfigure}[t]{0.22\columnwidth}
             \centering
             \includegraphics[width=\textwidth]{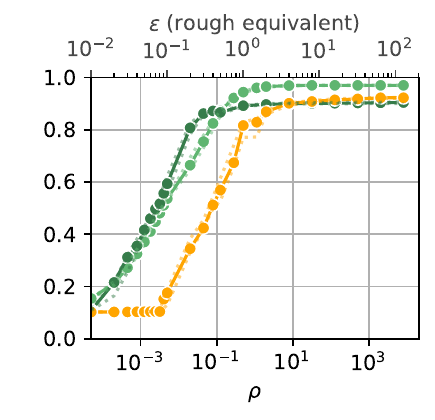}
         \end{subfigure}
         \hfill
         \begin{subfigure}[t]{0.22\columnwidth}
             \centering
             \includegraphics[width=\textwidth]{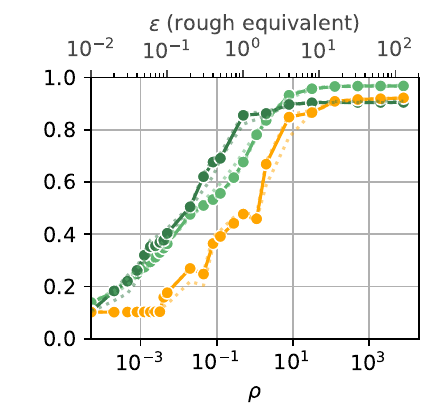}
         \end{subfigure}
         \hfill
         \begin{subfigure}[t]{0.22\columnwidth}
             \centering
             \includegraphics[width=\textwidth]{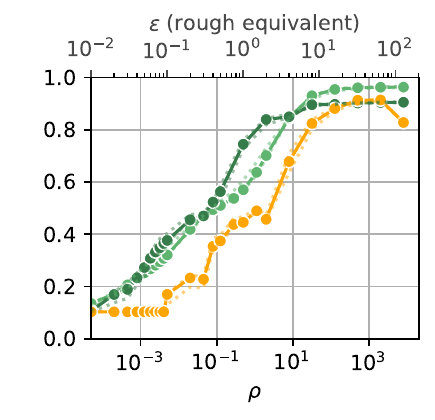}
         \end{subfigure}
    
        \makebox[20pt]{\raisebox{20pt}{\rotatebox[origin=c]{90}{ViT-L-16}}}%
        \begin{subfigure}[t]{0.22\columnwidth}
             \centering
             \includegraphics[width=\textwidth]{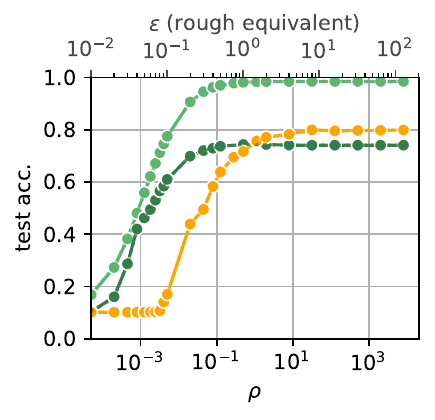}
         \end{subfigure}
         \hfill
         \begin{subfigure}[t]{0.22\columnwidth}
             \centering
             \includegraphics[width=\textwidth]{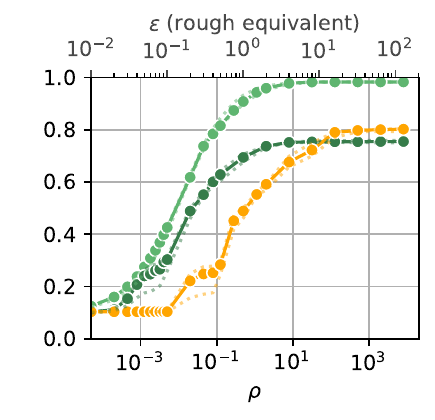}
         \end{subfigure}
         \hfill
         \begin{subfigure}[t]{0.22\columnwidth}
             \centering
             \includegraphics[width=\textwidth]{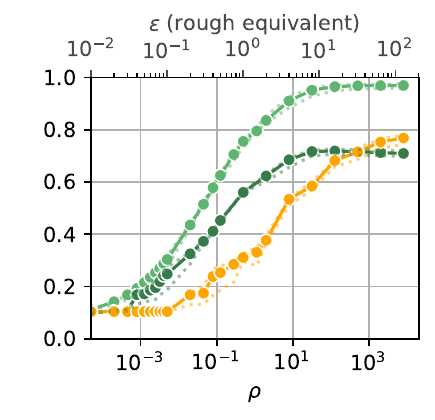}
         \end{subfigure}
         \hfill
         \begin{subfigure}[t]{0.22\columnwidth}
             \centering
             \includegraphics[width=\textwidth]{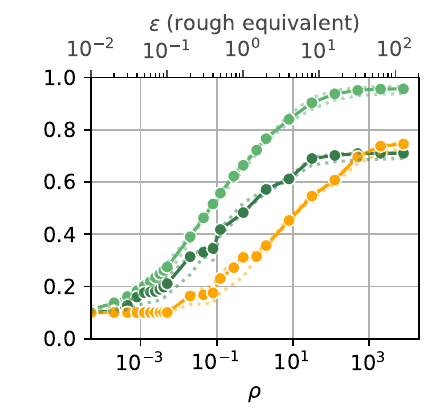}
         \end{subfigure}
    
        \makebox[20pt]{\raisebox{20pt}{\rotatebox[origin=c]{90}{ViT-H-14}}}%
         \begin{subfigure}[t]{0.22\columnwidth}
         \centering
         \includegraphics[width=\textwidth]{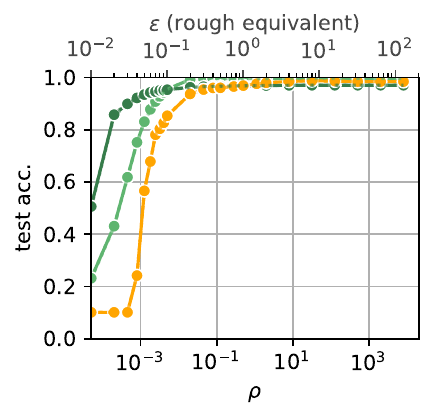}
         \end{subfigure}
         \hfill
        \begin{subfigure}[t]{0.22\columnwidth}
             \centering
             \includegraphics[width=\textwidth]{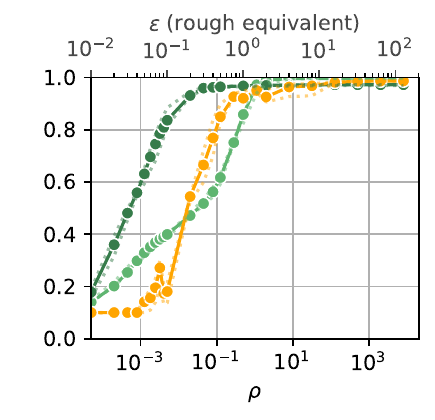}
         \end{subfigure}
         \hfill
         \begin{subfigure}[t]{0.22\columnwidth}
             \centering
             \includegraphics[width=\textwidth]{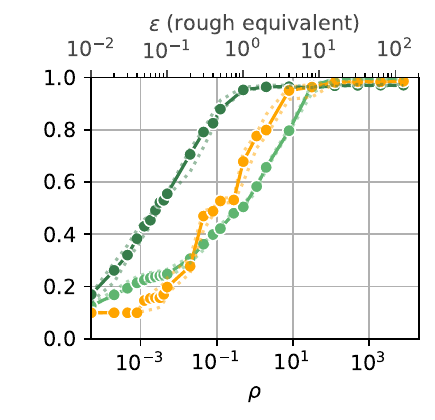}
         \end{subfigure}
         \hfill
         \begin{subfigure}[t]{0.22\columnwidth}
             \centering
             \includegraphics[width=\textwidth]{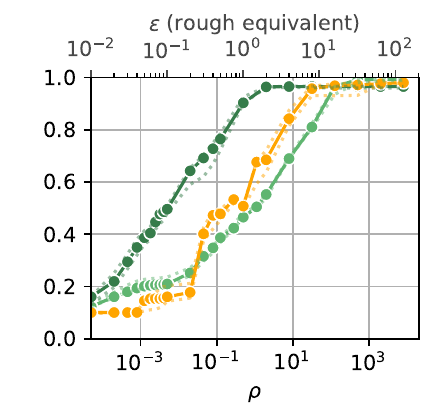}
         \end{subfigure}
    
        \makebox[20pt]{\raisebox{20pt}{\rotatebox[origin=c]{90}{ResNet-50}}}%
        \begin{subfigure}[t]{0.22\columnwidth}
             \centering
             \includegraphics[width=\textwidth]{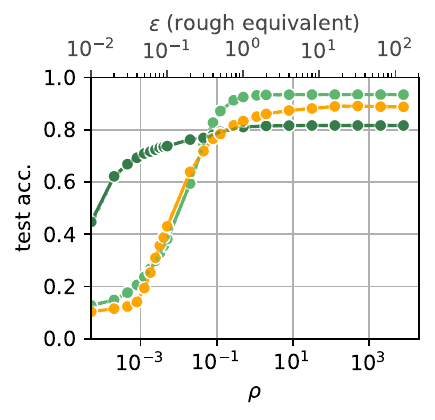}
             \caption{IR=1.}
         \end{subfigure}
         \hfill
         \begin{subfigure}[t]{0.22\columnwidth}
             \centering
             \includegraphics[width=\textwidth]{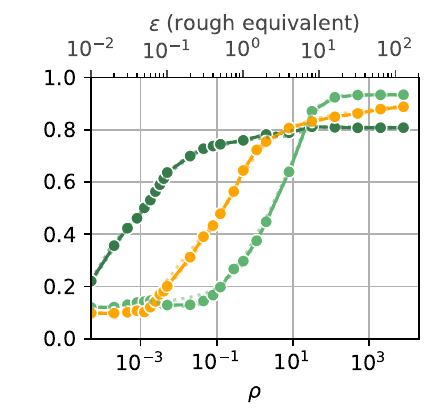}
             \caption{IR=10.}
         \end{subfigure}
         \hfill
         \begin{subfigure}[t]{0.22\columnwidth}
             \centering
             \includegraphics[width=\textwidth]{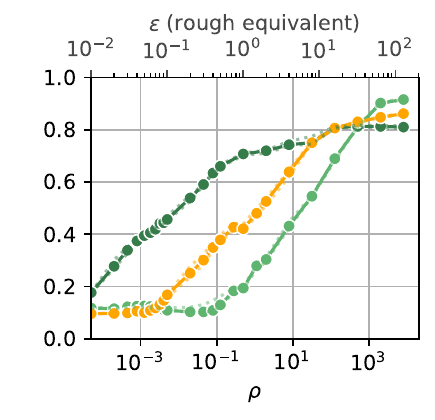}
             \caption{IR=50.}
         \end{subfigure}
         \hfill
         \begin{subfigure}[t]{0.22\columnwidth}
             \centering
             \includegraphics[width=\textwidth]{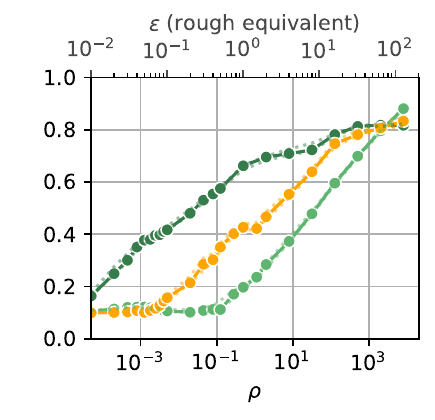}
             \caption{IR=100.}
         \end{subfigure}
    \caption{\textbf{\DPthreetext \, with Top-$K$.}
    We present the results including \DPthree Top-$K$ of STL10 on ViT-B-16, ViT-H-14, ViT-L-16 and ResNet-50, using ImageNet as public data for \DPthree, at different levels of imbalance rations (IR).
    }
    \label{fig:app-topk-stl10}
    \end{figure}

\subsubsection{Results}
\begin{figure}[t]
    \centering
    \includegraphics[width=0.6\columnwidth]{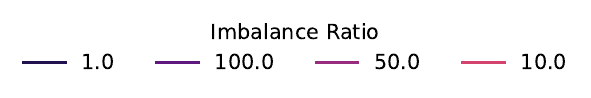}
    
    \includegraphics[width=0.5\columnwidth]{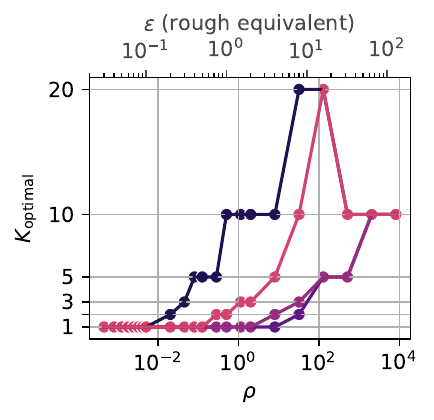}
    \caption{\textbf{Optimal $K$ for Top-$K$}}
    \label{fig:appendix-optimal-topk}
\end{figure}

We sweep over $K\in[1,2,3,5,10,20]$ and sort by balanced train accuracy to find the optimal $K$ per privacy value, which \Cref{fig:appendix-optimal-topk} shows. As we increase the privacy budget, the optimal accuracy is achieved at increasing $K$'s. For $\rho>100$ all $K_{\text{optimal}}$ converge to $K=10$. In this privacy regime, the Top-$K$ selection behaves essentially as it would non-privately. In this case, it seems to be detrimental to pick too many prototypes, although we note the accuracy of $K\in\{5,10,20\}$ is almost on par, being $76.7\%$, $77.1\%$ and $77.2\%$ respectively for $\rho=128$.
\Cref{fig:ablation-topk} shows the method requires a privacy budget somewhere between \DPtwo and \DPthree, while having a maximum accuracy also in between the maximum accuracy of those methods, closer to the higher accuracy of \DPtwo. We show the full set of results in 
\Cref{fig:app-topk-cifar10,fig:app-topk-cifar100,fig:app-topk-food101,fig:app-topk-stl10}.

\subsection{Sampling Mechanisms}
\label{app:mechanism}
We compare two different sampling mechanisms for \DPthree and find that both commonly utilized mechanisms, the Laplace and Exponential mechanism, produce very similar results. We show accuracies for various datasets and privacy budgets in \Cref{tab:app_sample_mechanism}. In fact, if the Laplace mechanism is used with a Gumbel noise distribution, the output distribution of the Laplace mechanism and Exponential mechanism are identical. This is commonly referred to as the \textit{Gumbel max trick} \cite{privacyBetterPrivacyAnalysis}.

We utilize the Exponential mechanism for its additional flexibility, which allows us to build \DPfour as described in \Cref{app:topk-method}.
\begin{table}[t]
\centering

\label{tab:app_sample_mechanism}
\resizebox{\columnwidth}{!}{%
\begin{tabular}{l|l|llll}
         & Method \textbackslash $\rho$ & $0.005$ & $0.02$ & $0.125$ & $0.5$ \\ \hline
CIFAR10  & Exponential & 91.45   & 91.33   & 91.27 & 91.25 \\
         & Laplace     & 91.45   & 91.33   & 91.28 & 91.26 \\ \hline
CIFAR100 & Exponential & 36.36   & 70.47   & 72.22 & 72.75 \\
         & Laplace     & 35.89   & 70.48   & 72.26 & 72.72 \\ \hline
Food101  & Exponential & 19.09   & 53.71   & 59.19 & 66.18 \\
         & Laplace     & 18.82   & 53.77   & 59.21 & 66.20
\end{tabular}%
}
\caption{C\textbf{omparison between Noisy Max (Laplace Mechanism) and Exponential Mechanism for sampling prototypes.}}
\end{table}

\subsection{Projection}
\label{app:projection}
\subsubsection{Setup}
The projection consists of a single layer linear network $f : \mathbb{R}^{d_{\text{avg}}}\rightarrow\mathbb{R}^{d_p}$ with no activation function and an average pooling layer with kernel size $p_{\text{before}}\in[1,64]$ before the linear layer. As we leave the total privacy budget $\rho$ unchanged, we introduce a hyperparameter $s\in[0.1,0.9]$ which defines the privacy budget of the projection layer $\rho_l = s*\rho$ and of the prototype estimation $\rho_p=(1-s)*\rho$. In total, the hyperparameters are $s\in[0.1,0.9]$, $p_{\text{before}}\in[1,64]$, output dimension $d_{p}$, the number of augments per step $n$, batch-size, learning-rate, gradient clipping norm and number of training steps. 

We train the projection with the original training rule from \citet{snellPrototypicalNetworksFewshot2017}, with some adaptions for privacy. We perform Bernoulli sampling (often more generally referred to as Poisson sampling) to receive a batch $B=(\*X,\*y)$. We split $(\*X,\*y)$ evenly into support and query $(\*X_S,\*y_S),(\*X_Q,\*y_Q)$, s.t. each part has the same number of samples per class. If a class has only one corresponding sample in $\*X$, we drop it. The prototypes for each class $\*p_c$ are estimated as the mean of samples in the support set $\*X_S$ that have the corresponding class label in $\*y_S$.
\begin{equation}
    \*X_{S,c}=\{\*x_i\in\*{X}_S | y_i = c\}
\end{equation}
\begin{equation}
    \*p_c = \frac{1}{|\*X_{S,c}|}\sum_{\*x\in\*X_{S,c}}\*x
\end{equation}Then, prototypes and query samples are projected with the linear layer.
\begin{equation}
    \*X'_{Q} = \{f(\*x)\ |\; \*x \in \*X_{Q}\}
\end{equation}
\begin{equation}
    \*p'_{c} = f(\*p_c)
\end{equation}
Finally, the model aims to classify each sample in $\*X'_Q$ by assigning it the label of the closest prototype 
\begin{equation}
    \hat{\*y} = \{\underset{c}{\arg\min} \; d(\*p'_{c}, \*x') | \*x'\in \*X'_Q\}
\end{equation}
We implement the classification training using a log-softmax over the distances to the prototypes and the negative log likelihood loss. 
This entire process, beginning with the split of $B$, is repeated $n$ times, before aggregating the loss and conducting the private gradient descent on the projection layer weights.
\subsubsection{\DPtwotext}
\begin{figure}
    \centering
    \includegraphics[width=0.5\columnwidth]{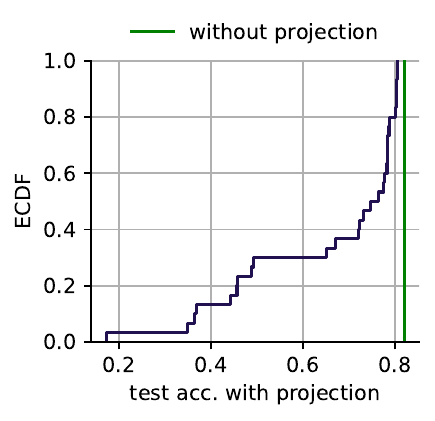}
    \caption{Results of the hyperparameter sweep of \DPtwo with projection.}
    \label{fig:app-projection-ecdf}
\end{figure}

For \DPtwo we expected the reduction in dimensionality to potentially improve to utility-privacy-tradeoff, but instead higher dimensions were strictly better, leading to the removal of $p_{\text{before}}$ and $d_{p}$ as hyperparameters. We show in \Cref{fig:app-projection-ecdf} the distribution of accuracies during the hyperparameter optimization and that no configuration reached the performance without projection. 
\subsubsection{\DPthreetext}
For \DPthree, we additionally need to project the public data embeddings, to find the prototypes in the projected latent space. Furthermore, we found that the utility is strictly lower than without projection. It seems that what the projection layers learns is fundamentally misaligned with the actual task. An obvious mismatch between training and application of the model is that we take the means as prototypes $\*p_c$ during training, but we later pick these prototypes from public data. Even after accounting for this, and using the actual prototypes during the projection training, utility didn't improve.

For both methods, the utility with projection is always worse \Cref{fig:ablation-projection} shows. The necessary accounting for the privacy costs of optimizing the additional hyperparameters~\cite{papernot2021hyperparameter} would further reduce the utility.

\section{Privacy Proofs and Privacy Conversion}
\label{app:privacyproofs}
\subsection{Full Proof for Privacy Guarantees of \DPthreetext}
\label{app:dp3proof}
We recall that our utility function is 
\begin{equation}
    u(\hat{\*x},c) = \sum_{i=0}^{|\*X_c|}1+\frac{\hat{M}(\*x_i)\cdot\hat{M}(\hat{\*x})}{\norm{\hat{M}(\*x_i)}  \norm{\hat{M}(\hat{\*x})}}
\end{equation}
where $\*X_c \in \*X$ are disjoint subsets of the private data $\*X$ we want to keep private.

We choose $u$ to be the sum and not, for example, the mean of the cosine similarity, to make $u$ monotonic w.r.t. $\*X$. It can be easily verified that the two different utility functions (mean and sum) lead to an identical mechanism, since the changes in $\Delta u$ and $u$ cancel each other out.
As we exhaust the full range $[0,2]$ of the cosine similarities, we clip each similarity to $[d_{\text{min}},d_{\text{max}}]$ and then substract $d_{\text{min}}$. This gives us the adapted utility function
\begin{equation}
    u(\hat{\*x},c) = \sum_{i=0}^{|\*X_c|}\text{clip}\left(1+\frac{\hat{M}(\*x_i)\cdot\hat{M}(\hat{\*x})}{\norm{\hat{M}(\*x_i)}  \norm{\hat{M}(\hat{\*x})}},d_{\text{min}},d_{\text{max}}\right) - d_{\text{min}}
\end{equation}
\begin{lemma}
\label{lem:monotonic-utility}
    $\Delta u = d_{\text{max}}-d_{\text{min}}$ and $u$ is positively monotonic w.r.t. to $\*X$.
\end{lemma}
\textit{Proof.} Since the cosine similarity's range is bound to $[0,d_{\text{max}}-d_{\text{min}}]$, each private sample contributes one non-negative summand in $[0,d_{\text{max}}-d_{\text{min}}]$. It immediately follows that $\Delta u=d_{\text{max}}-d_{\text{min}}$ and $u$ is positively monotonic w.r.t. $\*X$.

\begin{theorem}
    \DPthree is $\varepsilon$-DP.
\end{theorem}
\textit{Proof.} We sample the public prototypes independently for each class, using a utility function on disjoint sets $\*X_c \in \*X$ (each training data point only has one label), s.t. parallel composition applies. Each class prototype is sampled with the exponential mechanism, with probability $\Pr[\hat{\*x}] \propto \exp{\left(\epsilon u_{(\hat{\*x},c)}/\Delta u\right)}$ for outputting $\hat{\*x}$ as the class prototype, with $\Delta u$ denoting the sensitivity of $u_{(\hat{\*x},c)}$. Our utility function is monotonic (\Cref{lem:monotonic-utility}) and the described exponential mechanism is $\varepsilon$-DP for monotonic utility functions (\Cref{lem:monotonic-exp}). Since parallel composition applies and each parallel algorithm is $\varepsilon$-DP, the overall algorithm is $\varepsilon$-DP.

\subsection{Comparing Between Different Notions of DP}
\label{app:dp_conversions}
Note that to fairly compare our developed method that yields \textit{pure DP} guarantees against related work that yield zCDP, we convert our method using \Cref{lem:exp-zcdp}. To obtain a $\rho$-zCDP guarantee for the Linear Probing and DPSGD-Global-Adapt baselines, we perform full batch training and obtain the guarantee from \Cref{prop:gaussian-zcdp}.

\section{Additional Experimental Results}
\label{app:results}

\subsection{Imbalanced Experiments}
\label{sub:appendix-imbalanced}
\begin{figure}[t]
\centering 
    \includegraphics[width=\columnwidth]{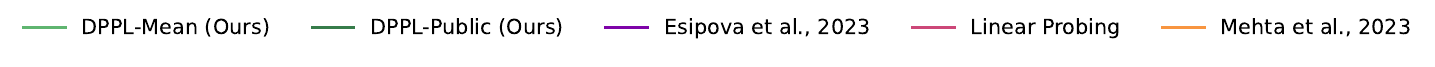}

    \makebox[20pt]{\raisebox{20pt}{\rotatebox[origin=c]{90}{ViT-B-16}}}%
    \begin{subfigure}[t]{0.22\columnwidth}
         \centering
         \includegraphics[width=\textwidth]{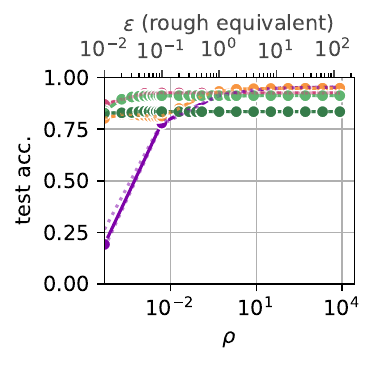}
     \end{subfigure}
     \hfill
    \begin{subfigure}[t]{0.22\columnwidth}
         \centering
         \includegraphics[width=\textwidth]{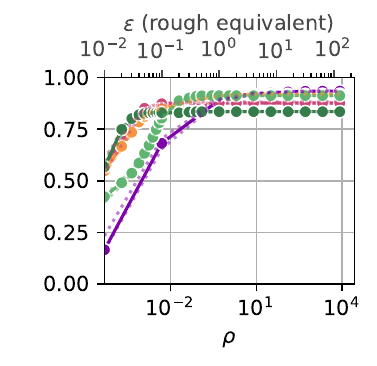}
     \end{subfigure}
     \hfill
     \begin{subfigure}[t]{0.22\columnwidth}
         \centering
         \includegraphics[width=\textwidth]{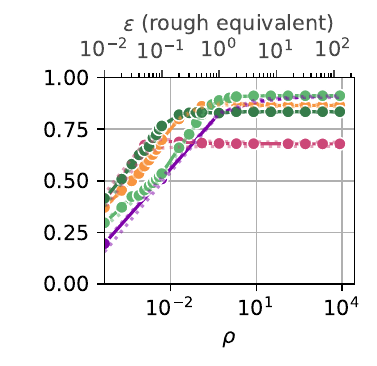}
     \end{subfigure}
     \hfill
     \begin{subfigure}[t]{0.22\columnwidth}
         \centering
         \includegraphics[width=\textwidth]{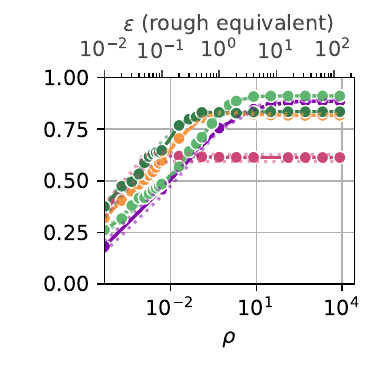}
     \end{subfigure}

    \makebox[20pt]{\raisebox{20pt}{\rotatebox[origin=c]{90}{ViT-L-16}}}%
    \begin{subfigure}[t]{0.22\columnwidth}
         \centering
         \includegraphics[width=\textwidth]{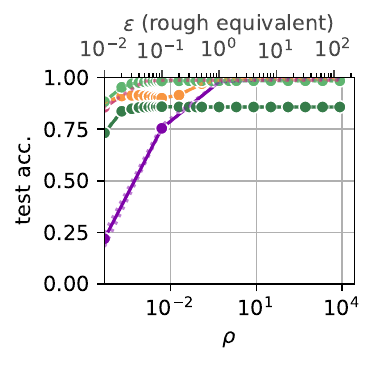}
     \end{subfigure}
     \hfill
     \begin{subfigure}[t]{0.22\columnwidth}
         \centering
         \includegraphics[width=\textwidth]{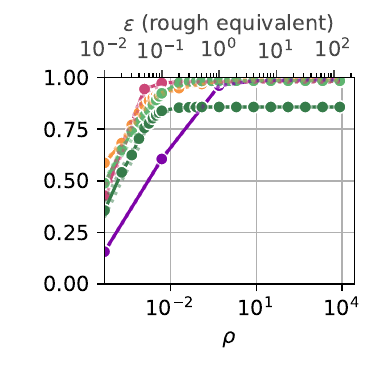}
     \end{subfigure}
     \hfill
     \begin{subfigure}[t]{0.22\columnwidth}
         \centering
         \includegraphics[width=\textwidth]{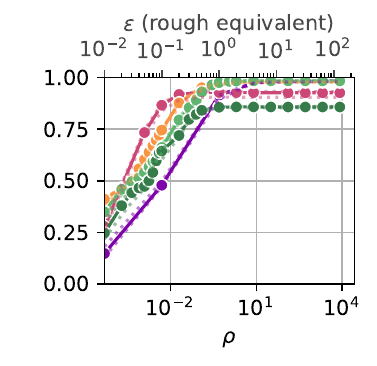}
     \end{subfigure}
     \hfill
     \begin{subfigure}[t]{0.22\columnwidth}
         \centering
         \includegraphics[width=\textwidth]{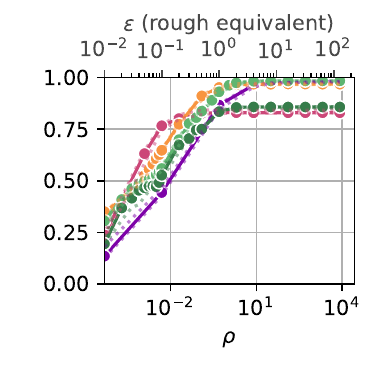}
     \end{subfigure}

    \makebox[20pt]{\raisebox{20pt}{\rotatebox[origin=c]{90}{ViT-H-14}}}%
     \begin{subfigure}[t]{0.22\columnwidth}
     \centering
     \includegraphics[width=\textwidth]{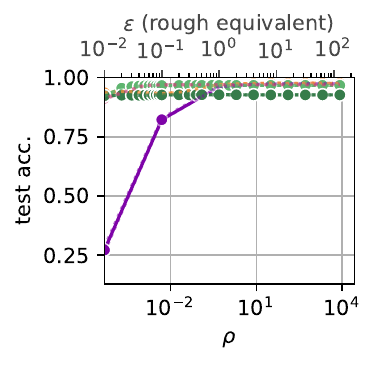}
     \end{subfigure}
     \hfill
    \begin{subfigure}[t]{0.22\columnwidth}
         \centering
         \includegraphics[width=\textwidth]{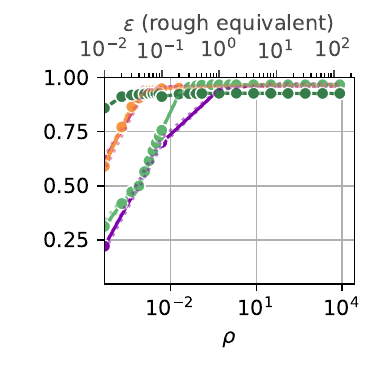}
     \end{subfigure}
     \hfill
     \begin{subfigure}[t]{0.22\columnwidth}
         \centering
         \includegraphics[width=\textwidth]{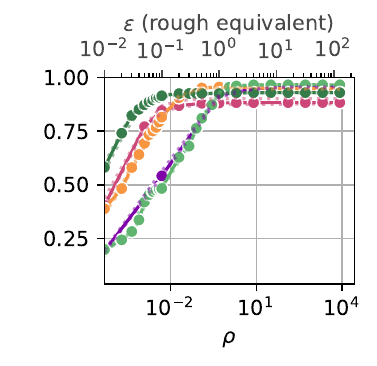}
     \end{subfigure}
     \hfill
     \begin{subfigure}[t]{0.22\columnwidth}
         \centering
         \includegraphics[width=\textwidth]{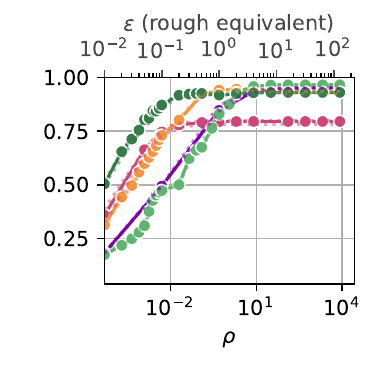}
     \end{subfigure}

    \makebox[20pt]{\raisebox{20pt}{\rotatebox[origin=c]{90}{ResNet-50}}}%
    \begin{subfigure}[t]{0.22\columnwidth}
         \centering
         \includegraphics[width=\textwidth]{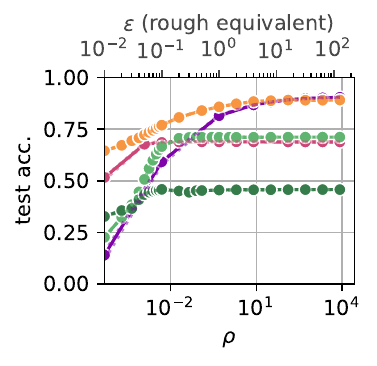}
         \caption{IR=1.}
     \end{subfigure}
     \hfill
     \begin{subfigure}[t]{0.22\columnwidth}
         \centering
         \includegraphics[width=\textwidth]{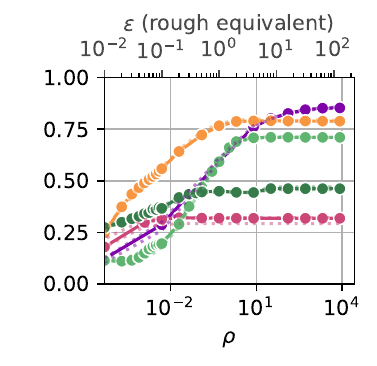}
         \caption{IR=10.}
     \end{subfigure}
     \hfill
     \begin{subfigure}[t]{0.22\columnwidth}
         \centering
         \includegraphics[width=\textwidth]{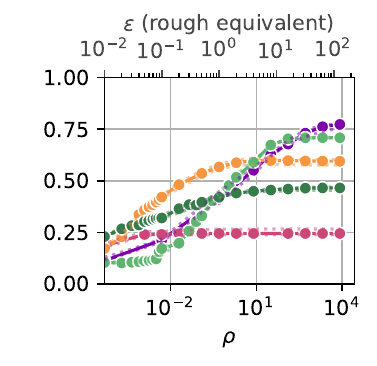}
         \caption{IR=50.}
     \end{subfigure}
     \hfill
     \begin{subfigure}[t]{0.22\columnwidth}
         \centering
         \includegraphics[width=\textwidth]{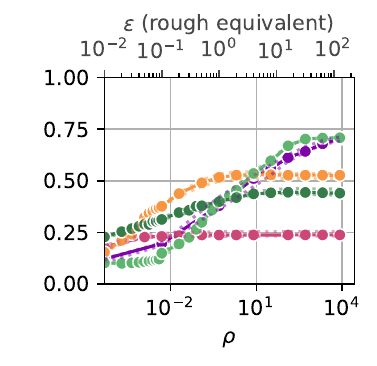}
         \caption{IR=100.}
     \end{subfigure}
\caption{\textbf{DP Prototypes on CIFAR10.}
We present the results for CIFAR10 on ViT-B-16, ViT-H-14, ViT-L-16 and ResNet-50, using ImageNet as public data for \DPthree, at different levels of imbalance rations (IR).
We compare to DP-LS by \citet{mehtaDifferentiallyPrivateImage2023} and DPSGD-Global-Adapt by \citet{esipova2023disparate}. Plotted is the median over multiple runs and dotted lines represent the upper and lower quantiles for all methods.
}
\label{fig:app-cifar10}
\end{figure}

\begin{figure}[t]
\centering 
    \includegraphics[width=\columnwidth]{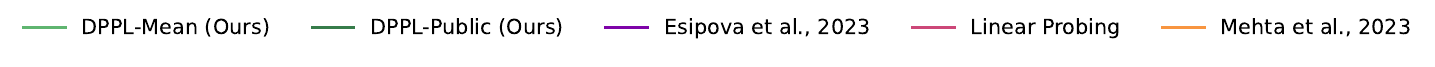}

    \makebox[20pt]{\raisebox{20pt}{\rotatebox[origin=c]{90}{ViT-B-16}}}%
    \begin{subfigure}[t]{0.22\columnwidth}
         \centering
         \includegraphics[width=\textwidth]{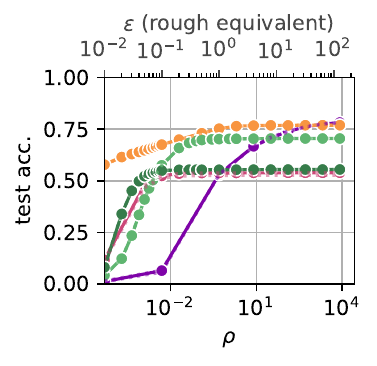}
     \end{subfigure}
     \hfill
    \begin{subfigure}[t]{0.22\columnwidth}
         \centering
         \includegraphics[width=\textwidth]{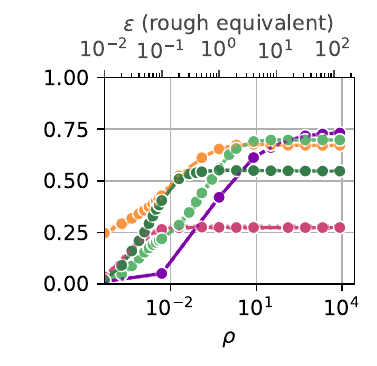}
     \end{subfigure}
     \hfill
     \begin{subfigure}[t]{0.22\columnwidth}
         \centering
         \includegraphics[width=\textwidth]{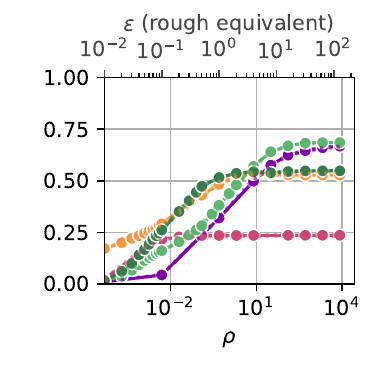}
     \end{subfigure}
     \hfill
     \begin{subfigure}[t]{0.22\columnwidth}
         \centering
         \includegraphics[width=\textwidth]{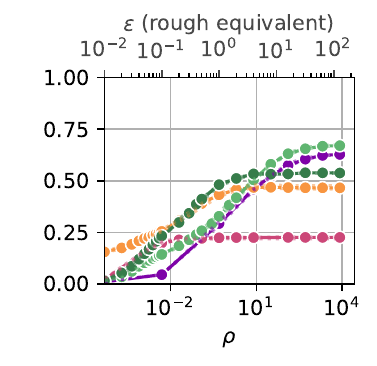}
     \end{subfigure}

    \makebox[20pt]{\raisebox{20pt}{\rotatebox[origin=c]{90}{ViT-L-16}}}%
    \begin{subfigure}[t]{0.22\columnwidth}
         \centering
         \includegraphics[width=\textwidth]{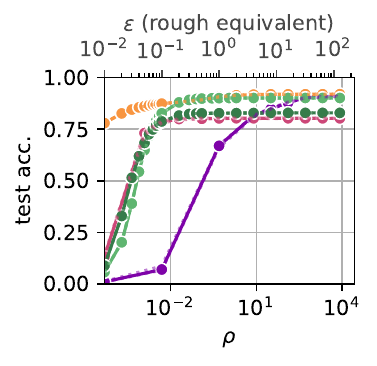}
     \end{subfigure}
     \hfill
     \begin{subfigure}[t]{0.22\columnwidth}
         \centering
         \includegraphics[width=\textwidth]{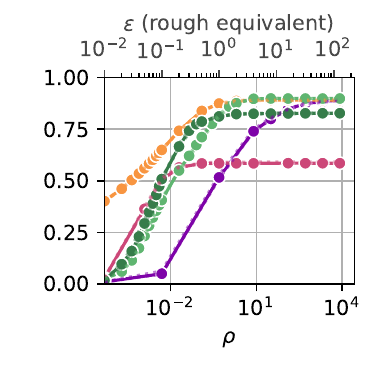}
     \end{subfigure}
     \hfill
     \begin{subfigure}[t]{0.22\columnwidth}
         \centering
         \includegraphics[width=\textwidth]{figures/04_results/single_plots/all_methods/cifar100/vit_large_patch14_dinov2.lvd142m/50_lim.pdf}
     \end{subfigure}
     \hfill
     \begin{subfigure}[t]{0.22\columnwidth}
         \centering
         \includegraphics[width=\textwidth]{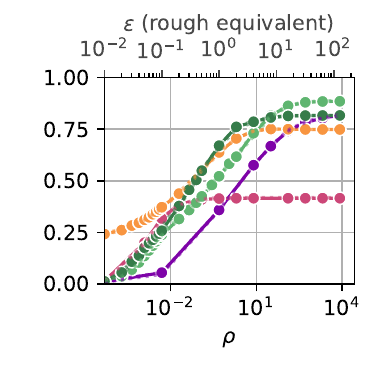}
     \end{subfigure}

    \makebox[20pt]{\raisebox{20pt}{\rotatebox[origin=c]{90}{ViT-H-14}}}%
     \begin{subfigure}[t]{0.22\columnwidth}
     \centering
     \includegraphics[width=\textwidth]{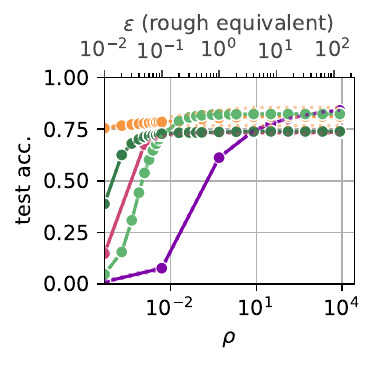}
     \end{subfigure}
     \hfill
    \begin{subfigure}[t]{0.22\columnwidth}
         \centering
         \includegraphics[width=\textwidth]{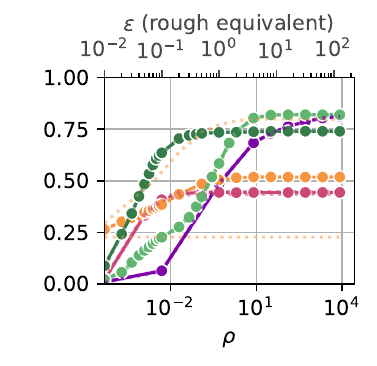}
     \end{subfigure}
     \hfill
     \begin{subfigure}[t]{0.22\columnwidth}
         \centering
         \includegraphics[width=\textwidth]{figures/04_results/single_plots/all_methods/cifar100/vit_h_14/50_lim.pdf}
     \end{subfigure}
     \hfill
     \begin{subfigure}[t]{0.22\columnwidth}
         \centering
         \includegraphics[width=\textwidth]{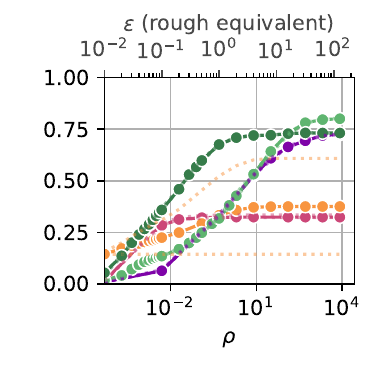}
     \end{subfigure}

    \makebox[20pt]{\raisebox{20pt}{\rotatebox[origin=c]{90}{ResNet-50}}}%
    \begin{subfigure}[t]{0.22\columnwidth}
         \centering
         \includegraphics[width=\textwidth]{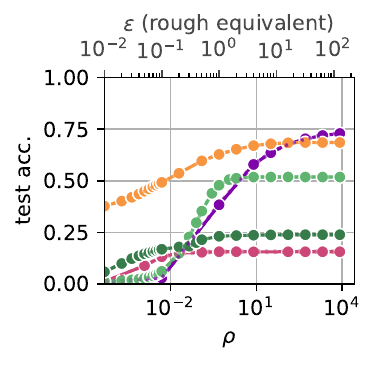}
         \caption{IR=1.}
     \end{subfigure}
     \hfill
     \begin{subfigure}[t]{0.22\columnwidth}
         \centering
         \includegraphics[width=\textwidth]{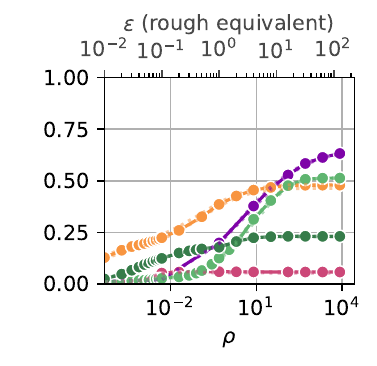}
         \caption{IR=10.}
     \end{subfigure}
     \hfill
     \begin{subfigure}[t]{0.22\columnwidth}
         \centering
         \includegraphics[width=\textwidth]{figures/04_results/single_plots/all_methods/cifar100/dino_resnet50/50_lim.pdf}
         \caption{IR=50.}
     \end{subfigure}
     \hfill
     \begin{subfigure}[t]{0.22\columnwidth}
         \centering
         \includegraphics[width=\textwidth]{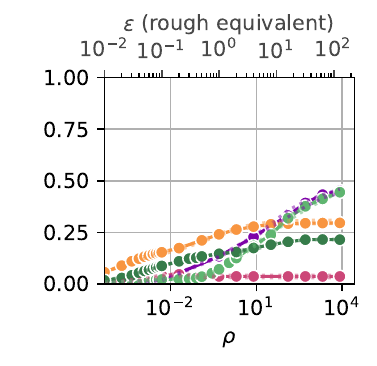}
         \caption{IR=100.}
     \end{subfigure}
\caption{\textbf{DP Prototypes on CIFAR100.}
We present the results for CIFAR100 on ViT-B-16, ViT-H-14, ViT-L-16 and ResNet-50, using ImageNet as public data for \DPthree, at different levels of imbalance rations (IR).
We compare to DP-LS by \citet{mehtaDifferentiallyPrivateImage2023} and DPSGD-Global-Adapt by \citet{esipova2023disparate}. Plotted is the median over multiple runs and dotted lines represent the upper and lower quantiles for all methods.
}
\label{fig:app-cifar100}
\end{figure}

\begin{figure}[t]
\centering 
    \includegraphics[width=\columnwidth]{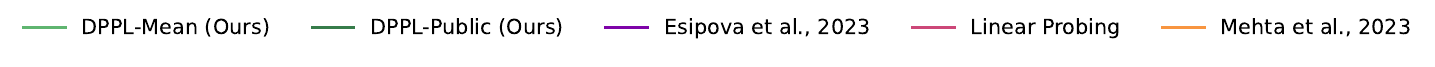}

    \makebox[20pt]{\raisebox{20pt}{\rotatebox[origin=c]{90}{ViT-B-16}}}%
    \begin{subfigure}[t]{0.22\columnwidth}
         \centering
         \includegraphics[width=\textwidth]{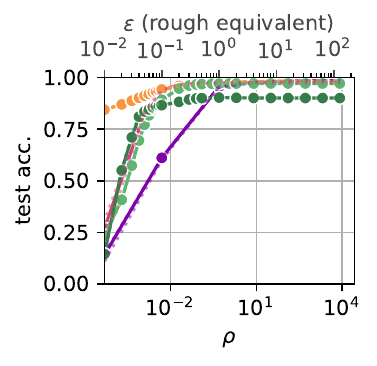}
     \end{subfigure}
     \hfill
    \begin{subfigure}[t]{0.22\columnwidth}
         \centering
         \includegraphics[width=\textwidth]{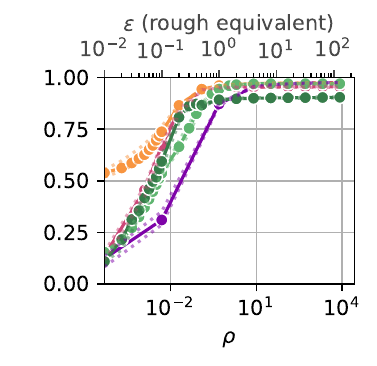}
     \end{subfigure}
     \hfill
     \begin{subfigure}[t]{0.22\columnwidth}
         \centering
         \includegraphics[width=\textwidth]{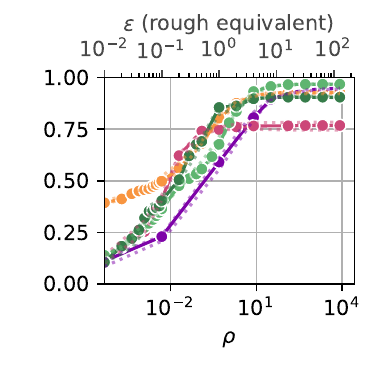}
     \end{subfigure}
     \hfill
     \begin{subfigure}[t]{0.22\columnwidth}
         \centering
         \includegraphics[width=\textwidth]{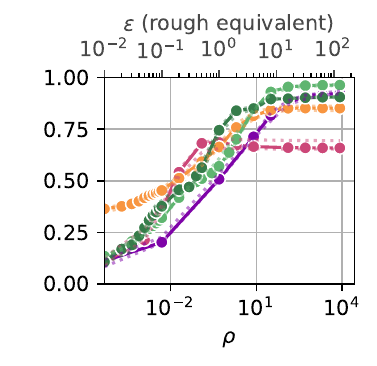}
     \end{subfigure}

    \makebox[20pt]{\raisebox{20pt}{\rotatebox[origin=c]{90}{ViT-L-16}}}%
    \begin{subfigure}[t]{0.22\columnwidth}
         \centering
         \includegraphics[width=\textwidth]{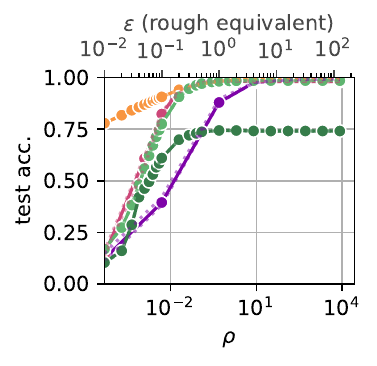}
     \end{subfigure}
     \hfill
     \begin{subfigure}[t]{0.22\columnwidth}
         \centering
         \includegraphics[width=\textwidth]{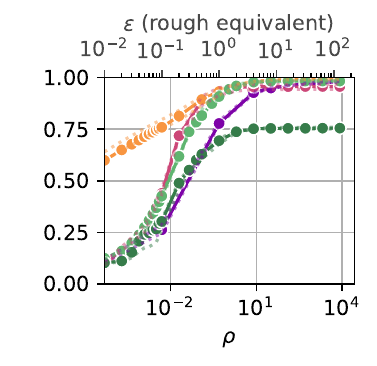}
     \end{subfigure}
     \hfill
     \begin{subfigure}[t]{0.22\columnwidth}
         \centering
         \includegraphics[width=\textwidth]{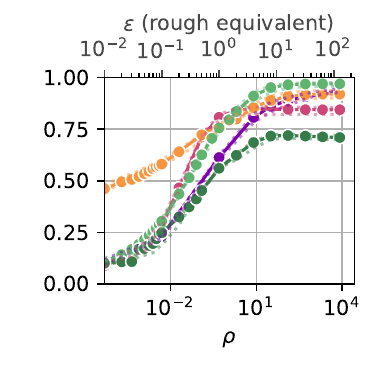}
     \end{subfigure}
     \hfill
     \begin{subfigure}[t]{0.22\columnwidth}
         \centering
         \includegraphics[width=\textwidth]{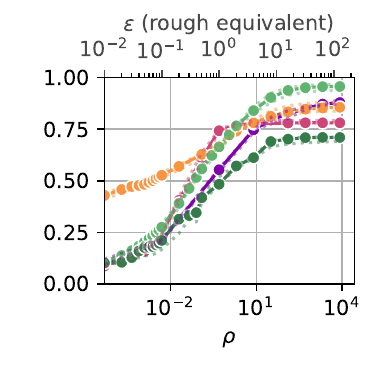}
     \end{subfigure}

    \makebox[20pt]{\raisebox{20pt}{\rotatebox[origin=c]{90}{ViT-H-14}}}%
     \begin{subfigure}[t]{0.22\columnwidth}
     \centering
     \includegraphics[width=\textwidth]{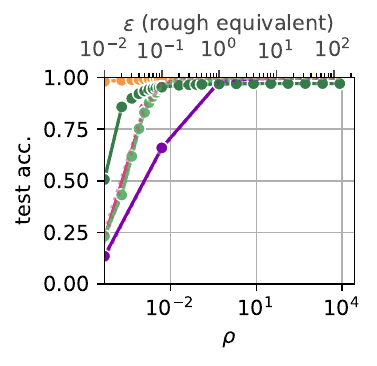}
     \end{subfigure}
     \hfill
    \begin{subfigure}[t]{0.22\columnwidth}
         \centering
         \includegraphics[width=\textwidth]{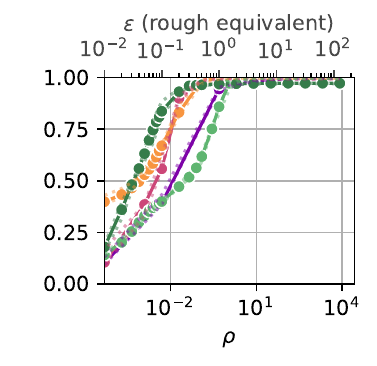}
     \end{subfigure}
     \hfill
     \begin{subfigure}[t]{0.22\columnwidth}
         \centering
         \includegraphics[width=\textwidth]{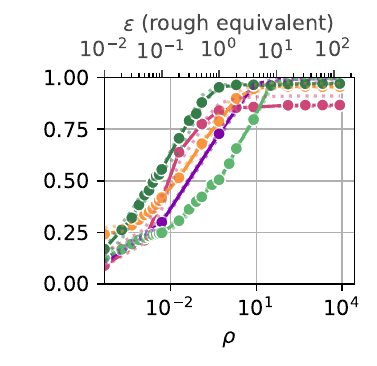}
     \end{subfigure}
     \hfill
     \begin{subfigure}[t]{0.22\columnwidth}
         \centering
         \includegraphics[width=\textwidth]{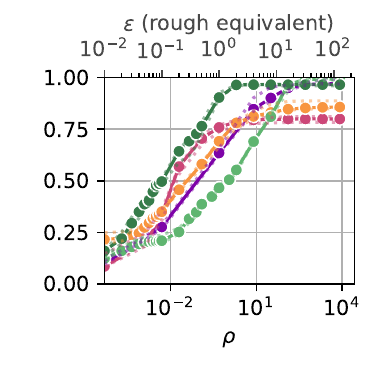}
     \end{subfigure}

    \makebox[20pt]{\raisebox{20pt}{\rotatebox[origin=c]{90}{ResNet-50}}}%
    \begin{subfigure}[t]{0.22\columnwidth}
         \centering
         \includegraphics[width=\textwidth]{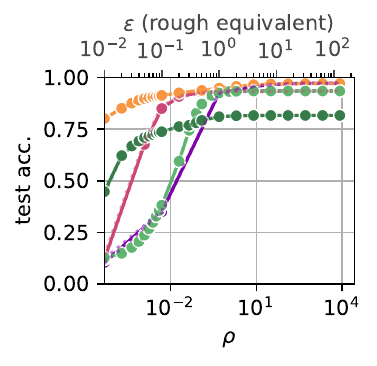}
         \caption{IR=1.}
     \end{subfigure}
     \hfill
     \begin{subfigure}[t]{0.22\columnwidth}
         \centering
         \includegraphics[width=\textwidth]{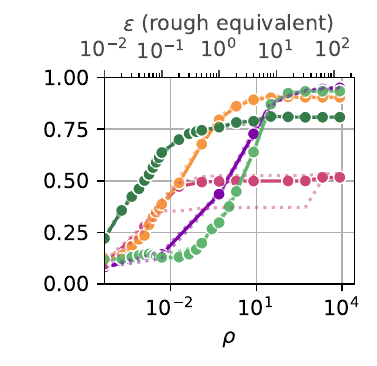}
         \caption{IR=10.}
     \end{subfigure}
     \hfill
     \begin{subfigure}[t]{0.22\columnwidth}
         \centering
         \includegraphics[width=\textwidth]{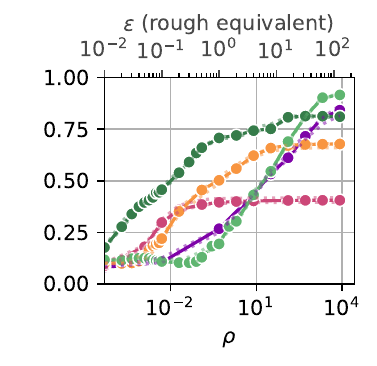}
         \caption{IR=50.}
     \end{subfigure}
     \hfill
     \begin{subfigure}[t]{0.22\columnwidth}
         \centering
         \includegraphics[width=\textwidth]{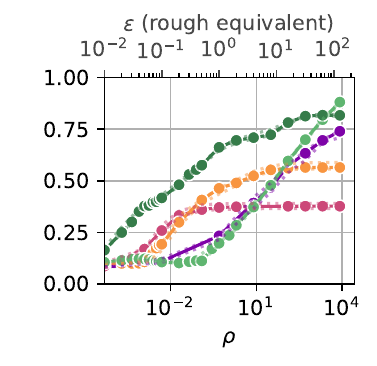}
         \caption{IR=100.}
     \end{subfigure}
\caption{\textbf{DP Prototypes on STL10.}
We present the results for STL10 on ViT-B-16, ViT-H-14, ViT-L-16 and ResNet-50, using ImageNet as public data for \DPthree, at different levels of imbalance rations (IR).
We compare to DP-LS by \citet{mehtaDifferentiallyPrivateImage2023} and DPSGD-Global-Adapt by \citet{esipova2023disparate}. Plotted is the median over multiple runs and dotted lines represent the upper and lower quantiles for all methods.
}
\label{fig:app-stl10}
\end{figure}

\begin{figure}[t]
\centering 
    \includegraphics[width=\columnwidth]{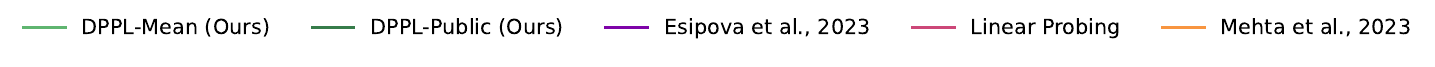}

    \makebox[20pt]{\raisebox{20pt}{\rotatebox[origin=c]{90}{ViT-B-16}}}%
    \begin{subfigure}[t]{0.22\columnwidth}
         \centering
         \includegraphics[width=\textwidth]{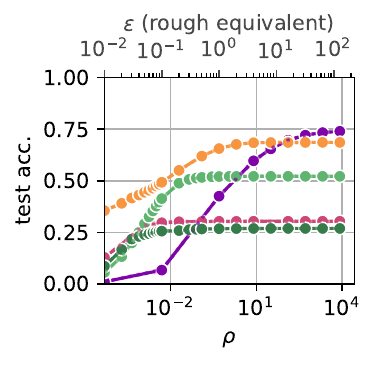}
     \end{subfigure}
     \hfill
    \begin{subfigure}[t]{0.22\columnwidth}
         \centering
         \includegraphics[width=\textwidth]{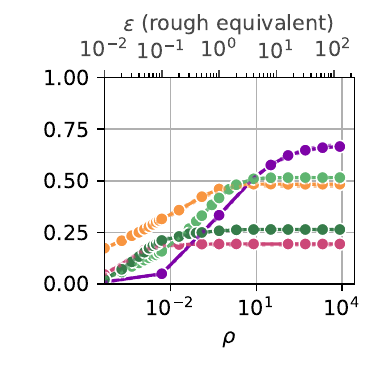}
     \end{subfigure}
     \hfill
     \begin{subfigure}[t]{0.22\columnwidth}
         \centering
         \includegraphics[width=\textwidth]{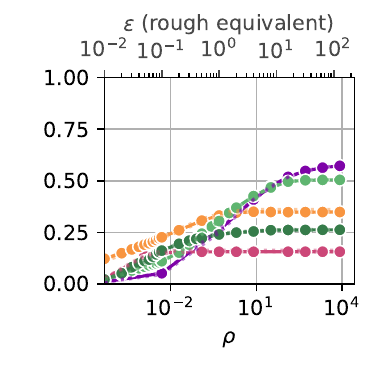}
     \end{subfigure}
     \hfill
     \begin{subfigure}[t]{0.22\columnwidth}
         \centering
         \includegraphics[width=\textwidth]{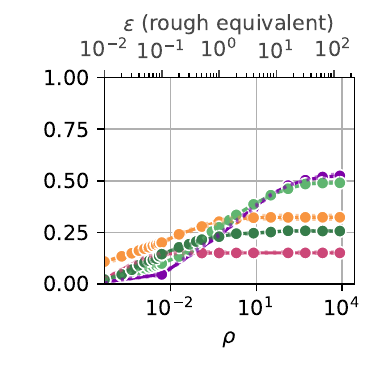}
     \end{subfigure}

    \makebox[20pt]{\raisebox{20pt}{\rotatebox[origin=c]{90}{ViT-L-16}}}%
    \begin{subfigure}[t]{0.22\columnwidth}
         \centering
         \includegraphics[width=\textwidth]{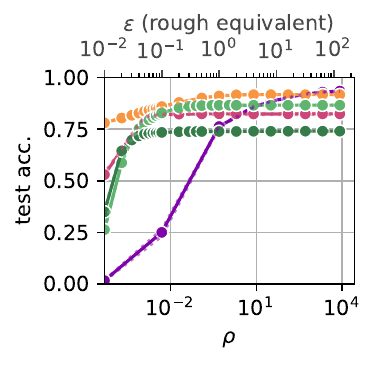}
     \end{subfigure}
     \hfill
     \begin{subfigure}[t]{0.22\columnwidth}
         \centering
         \includegraphics[width=\textwidth]{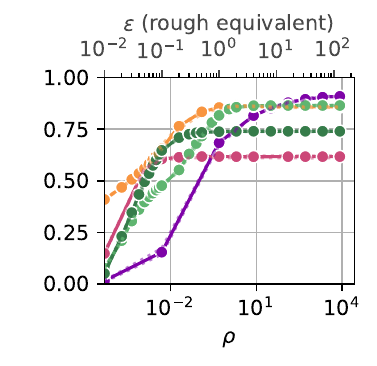}
     \end{subfigure}
     \hfill
     \begin{subfigure}[t]{0.22\columnwidth}
         \centering
         \includegraphics[width=\textwidth]{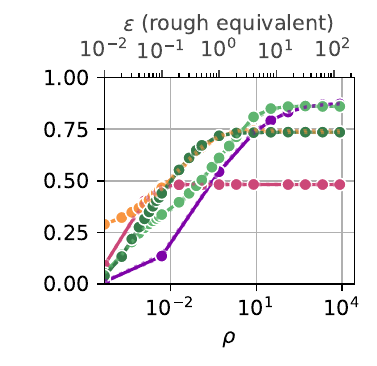}
     \end{subfigure}
     \hfill
     \begin{subfigure}[t]{0.22\columnwidth}
         \centering
         \includegraphics[width=\textwidth]{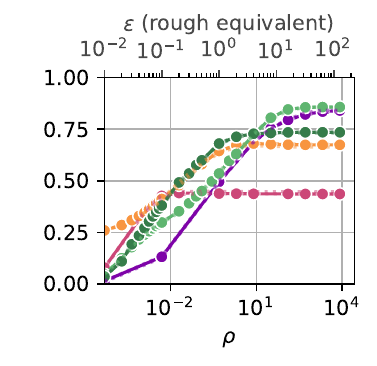}
     \end{subfigure}

    \makebox[20pt]{\raisebox{20pt}{\rotatebox[origin=c]{90}{ViT-H-14}}}%
     \begin{subfigure}[t]{0.22\columnwidth}
     \centering
     \includegraphics[width=\textwidth]{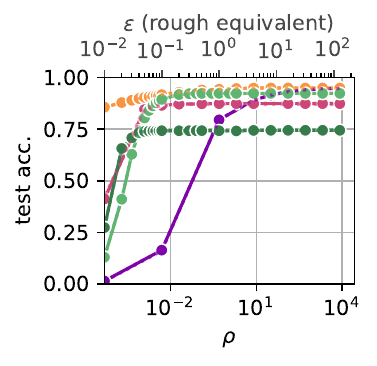}
     \end{subfigure}
     \hfill
    \begin{subfigure}[t]{0.22\columnwidth}
         \centering
         \includegraphics[width=\textwidth]{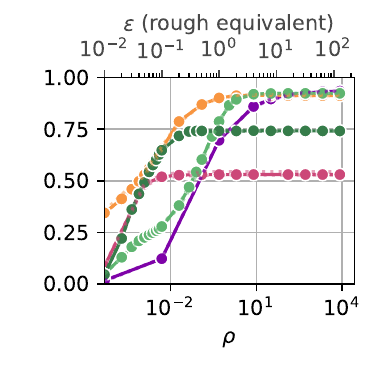}
     \end{subfigure}
     \hfill
     \begin{subfigure}[t]{0.22\columnwidth}
         \centering
         \includegraphics[width=\textwidth]{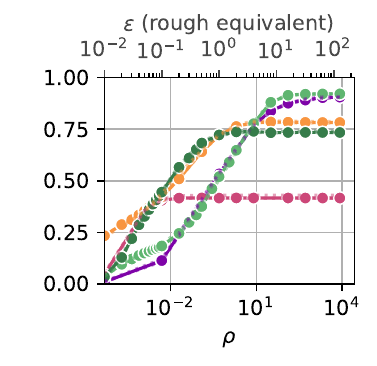}
     \end{subfigure}
     \hfill
     \begin{subfigure}[t]{0.22\columnwidth}
         \centering
         \includegraphics[width=\textwidth]{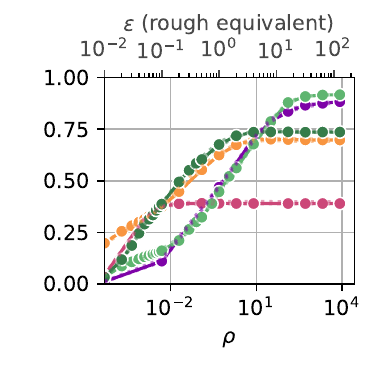}
     \end{subfigure}

    \makebox[20pt]{\raisebox{20pt}{\rotatebox[origin=c]{90}{ResNet-50}}}%
    \begin{subfigure}[t]{0.22\columnwidth}
         \centering
         \includegraphics[width=\textwidth]{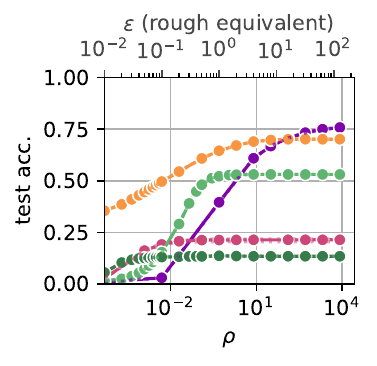}
         \caption{IR=1.}
     \end{subfigure}
     \hfill
     \begin{subfigure}[t]{0.22\columnwidth}
         \centering
         \includegraphics[width=\textwidth]{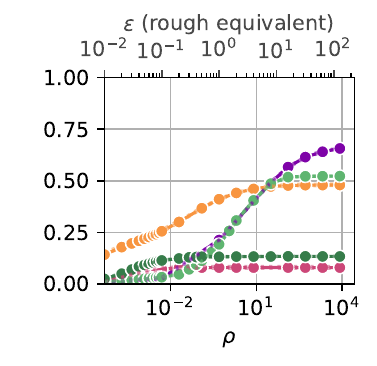}
         \caption{IR=10.}
     \end{subfigure}
     \hfill
     \begin{subfigure}[t]{0.22\columnwidth}
         \centering
         \includegraphics[width=\textwidth]{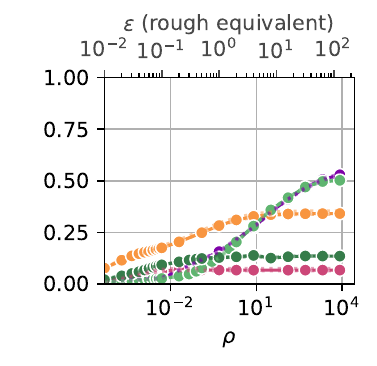}
         \caption{IR=50.}
     \end{subfigure}
     \hfill
     \begin{subfigure}[t]{0.22\columnwidth}
         \centering
         \includegraphics[width=\textwidth]{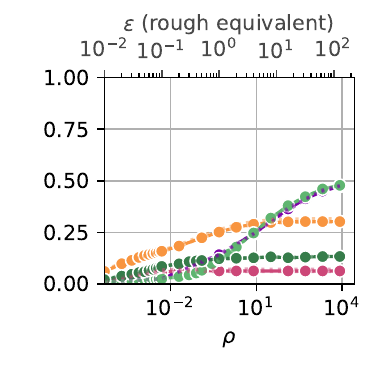}
         \caption{IR=100.}
     \end{subfigure}
\caption{\textbf{DP Prototypes on FOOD101.}
We present the results for FOOD101 on ViT-B-16, ViT-H-14, ViT-L-16 and ResNet-50, using ImageNet as public data for \DPthree, at different levels of imbalance rations (IR).
We compare to DP-LS by \citet{mehtaDifferentiallyPrivateImage2023} and DPSGD-Global-Adapt by \citet{esipova2023disparate}. Plotted is the median over multiple runs and dotted lines represent the upper and lower quantiles for all methods.
}
\label{fig:app-food101}
\end{figure}

We compare the accuracies for all methods, all encoders and imbalance ratios in $[1,10,50,100]$ in \Cref{fig:app-cifar10,fig:app-stl10,fig:app-food101,fig:app-cifar100}.
\subsection{Minority Class Accuracies}
\label{app:minority-acc}
\begin{figure}[t]
\centering 
    \includegraphics[width=\columnwidth]{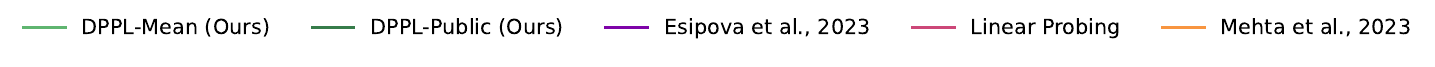}

    \makebox[20pt]{\raisebox{20pt}{\rotatebox[origin=c]{90}{ViT-B-16}}}%
    \begin{subfigure}[t]{0.22\columnwidth}
         \centering
         \includegraphics[width=\textwidth]{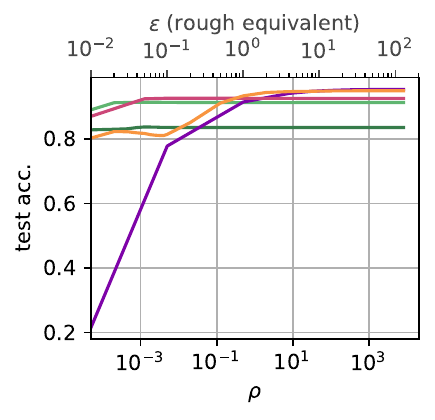}
     \end{subfigure}
     \hfill
    \begin{subfigure}[t]{0.22\columnwidth}
         \centering
         \includegraphics[width=\textwidth]{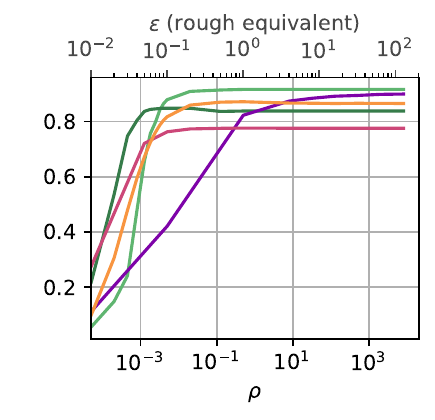}
     \end{subfigure}
     \hfill
     \begin{subfigure}[t]{0.22\columnwidth}
         \centering
         \includegraphics[width=\textwidth]{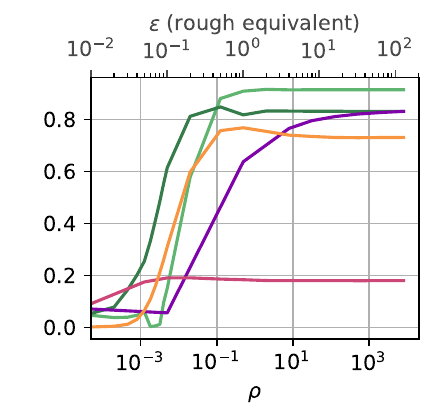}
     \end{subfigure}
     \hfill
     \begin{subfigure}[t]{0.22\columnwidth}
         \centering
         \includegraphics[width=\textwidth]{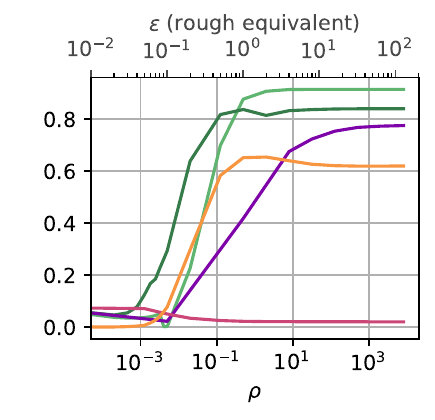}
     \end{subfigure}

    \makebox[20pt]{\raisebox{20pt}{\rotatebox[origin=c]{90}{ViT-L-16}}}%
    \begin{subfigure}[t]{0.22\columnwidth}
         \centering
         \includegraphics[width=\textwidth]{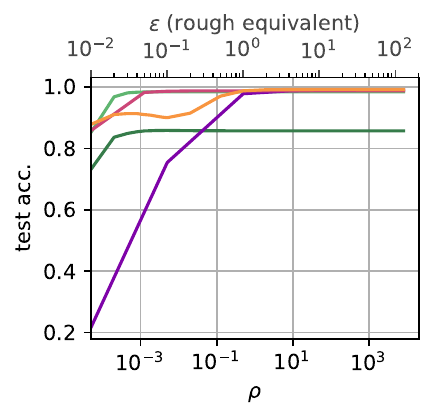}
     \end{subfigure}
     \hfill
     \begin{subfigure}[t]{0.22\columnwidth}
         \centering
         \includegraphics[width=\textwidth]{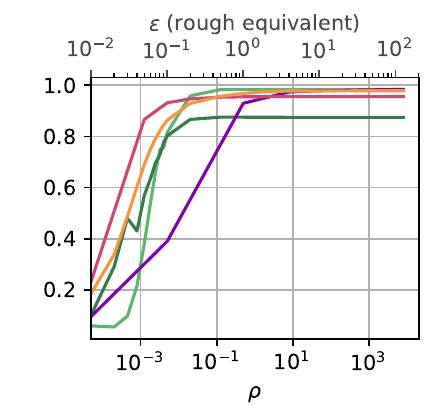}
     \end{subfigure}
     \hfill
     \begin{subfigure}[t]{0.22\columnwidth}
         \centering
         \includegraphics[width=\textwidth]{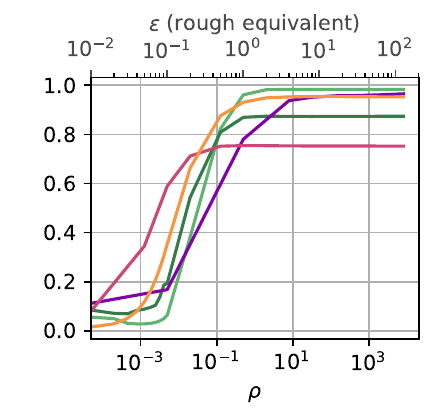}
     \end{subfigure}
     \hfill
     \begin{subfigure}[t]{0.22\columnwidth}
         \centering
         \includegraphics[width=\textwidth]{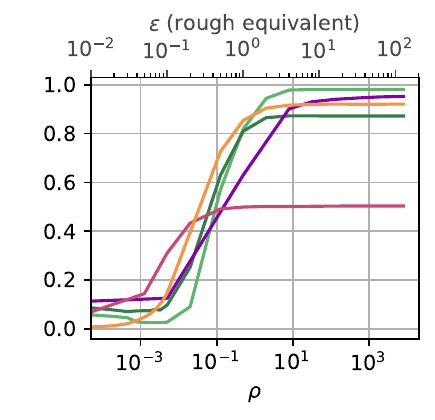}
     \end{subfigure}

    \makebox[20pt]{\raisebox{20pt}{\rotatebox[origin=c]{90}{ViT-H-14}}}%
     \begin{subfigure}[t]{0.22\columnwidth}
     \centering
     \includegraphics[width=\textwidth]{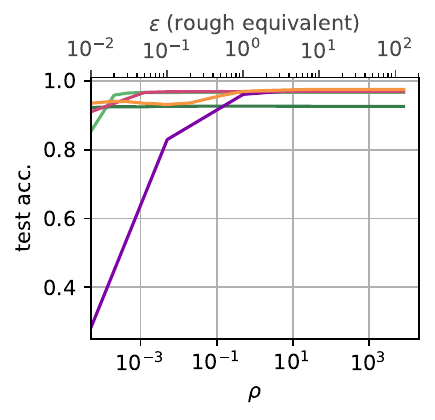}
     \end{subfigure}
     \hfill
    \begin{subfigure}[t]{0.22\columnwidth}
         \centering
         \includegraphics[width=\textwidth]{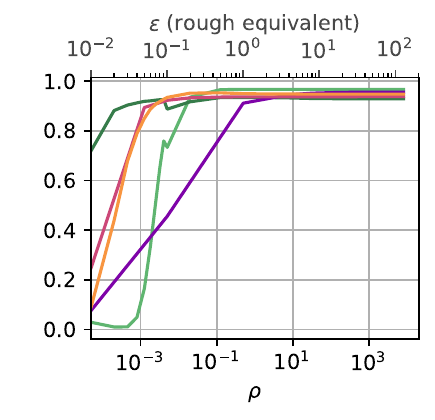}
     \end{subfigure}
     \hfill
     \begin{subfigure}[t]{0.22\columnwidth}
         \centering
         \includegraphics[width=\textwidth]{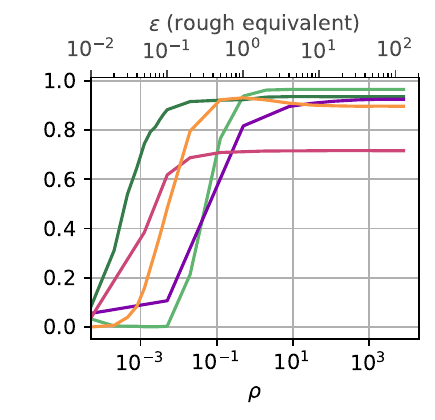}
     \end{subfigure}
     \hfill
     \begin{subfigure}[t]{0.22\columnwidth}
         \centering
         \includegraphics[width=\textwidth]{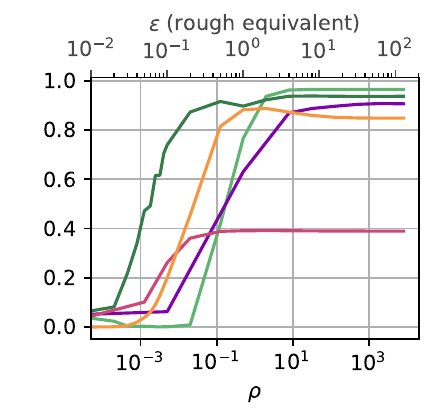}
     \end{subfigure}

    \makebox[20pt]{\raisebox{20pt}{\rotatebox[origin=c]{90}{ResNet-50}}}%
    \begin{subfigure}[t]{0.22\columnwidth}
         \centering
         \includegraphics[width=\textwidth]{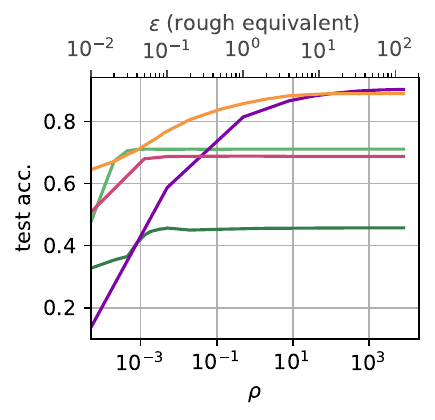}
         \caption{IR=1.}
     \end{subfigure}
     \hfill
     \begin{subfigure}[t]{0.22\columnwidth}
         \centering
         \includegraphics[width=\textwidth]{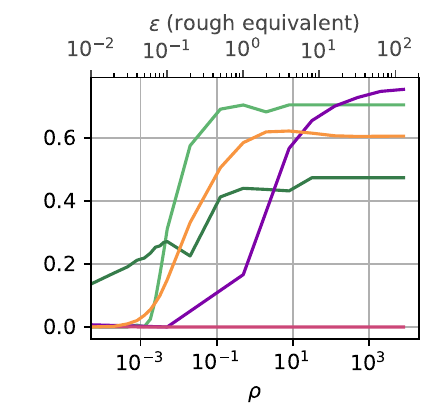}
         \caption{IR=10.}
     \end{subfigure}
     \hfill
     \begin{subfigure}[t]{0.22\columnwidth}
         \centering
         \includegraphics[width=\textwidth]{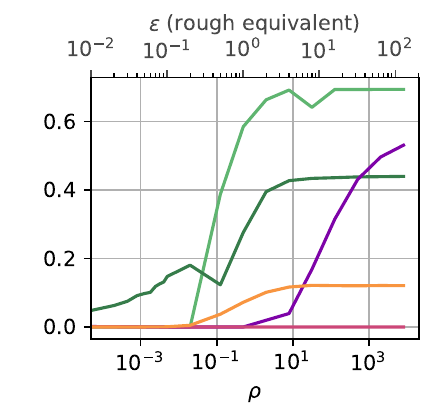}
         \caption{IR=50.}
     \end{subfigure}
     \hfill
     \begin{subfigure}[t]{0.22\columnwidth}
         \centering
         \includegraphics[width=\textwidth]{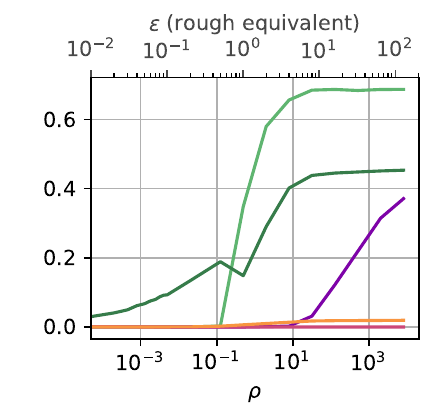}
         \caption{IR=100.}
     \end{subfigure}
\caption{\textbf{Minority class accuracies on CIFAR10.}
We present the results for the minority classes (lower 25\% quantile) of CIFAR10 on ViT-B-16, ViT-H-14, ViT-L-16 and ResNet-50, using ImageNet as public data for \DPthree, at different levels of imbalance rations (IR).
We compare to DP-LS by \citet{mehtaDifferentiallyPrivateImage2023} and DPSGD-Global-Adapt by \citet{esipova2023disparate}.
}
\label{fig:app-minority-cifar10}
\end{figure}

\begin{figure}[t]
\centering 
    \includegraphics[width=\columnwidth]{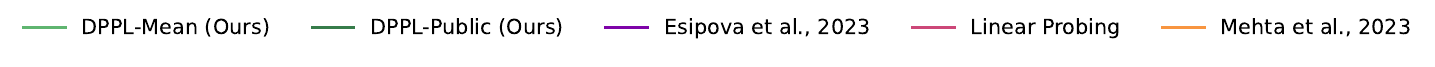}

    \makebox[20pt]{\raisebox{20pt}{\rotatebox[origin=c]{90}{ViT-B-16}}}%
    \begin{subfigure}[t]{0.22\columnwidth}
         \centering
         \includegraphics[width=\textwidth]{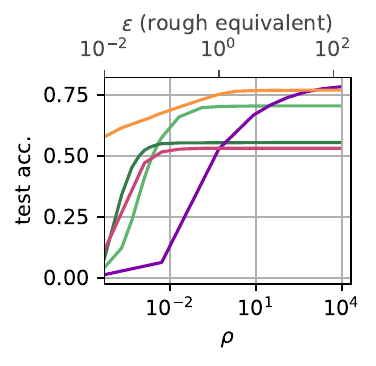}
     \end{subfigure}
     \hfill
    \begin{subfigure}[t]{0.22\columnwidth}
         \centering
         \includegraphics[width=\textwidth]{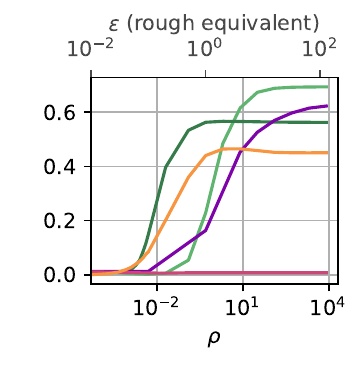}
     \end{subfigure}
     \hfill
     \begin{subfigure}[t]{0.22\columnwidth}
         \centering
         \includegraphics[width=\textwidth]{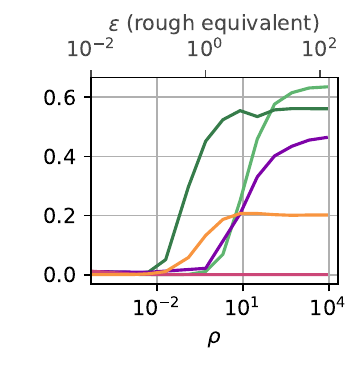}
     \end{subfigure}
     \hfill
     \begin{subfigure}[t]{0.22\columnwidth}
         \centering
         \includegraphics[width=\textwidth]{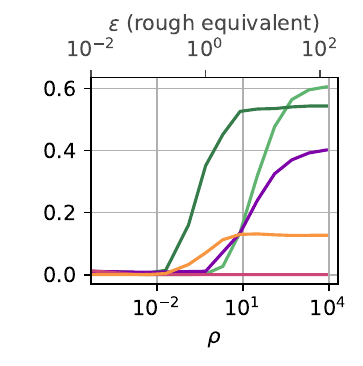}
     \end{subfigure}

    \makebox[20pt]{\raisebox{20pt}{\rotatebox[origin=c]{90}{ViT-L-16}}}%
    \begin{subfigure}[t]{0.22\columnwidth}
         \centering
         \includegraphics[width=\textwidth]{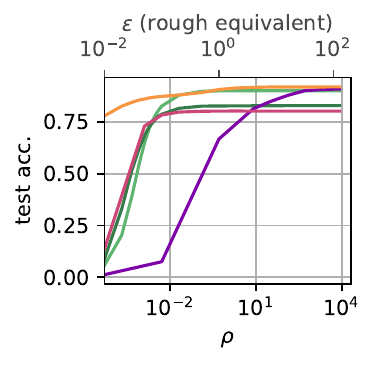}
     \end{subfigure}
     \hfill
     \begin{subfigure}[t]{0.22\columnwidth}
         \centering
         \includegraphics[width=\textwidth]{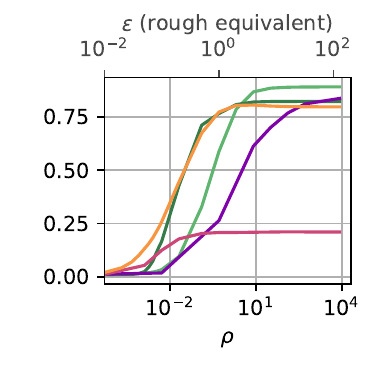}
     \end{subfigure}
     \hfill
     \begin{subfigure}[t]{0.22\columnwidth}
         \centering
         \includegraphics[width=\textwidth]{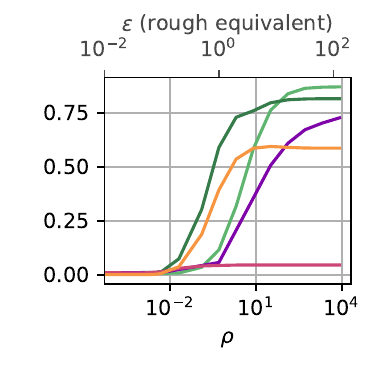}
     \end{subfigure}
     \hfill
     \begin{subfigure}[t]{0.22\columnwidth}
         \centering
         \includegraphics[width=\textwidth]{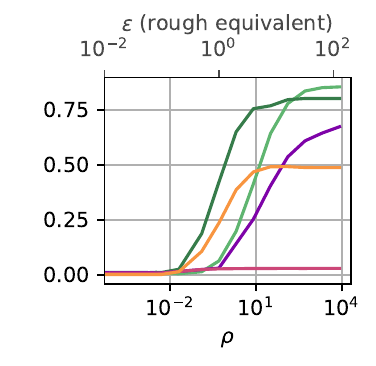}
     \end{subfigure}

    \makebox[20pt]{\raisebox{20pt}{\rotatebox[origin=c]{90}{ViT-H-14}}}%
     \begin{subfigure}[t]{0.22\columnwidth}
     \centering
     \includegraphics[width=\textwidth]{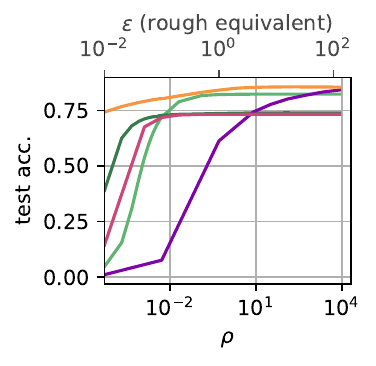}
     \end{subfigure}
     \hfill
    \begin{subfigure}[t]{0.22\columnwidth}
         \centering
         \includegraphics[width=\textwidth]{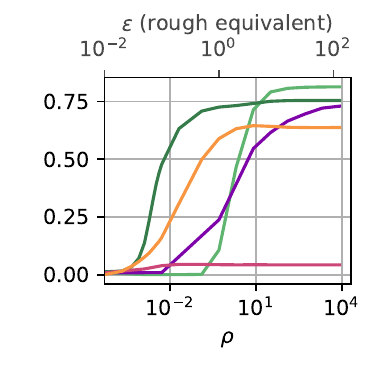}
     \end{subfigure}
     \hfill
     \begin{subfigure}[t]{0.22\columnwidth}
         \centering
         \includegraphics[width=\textwidth]{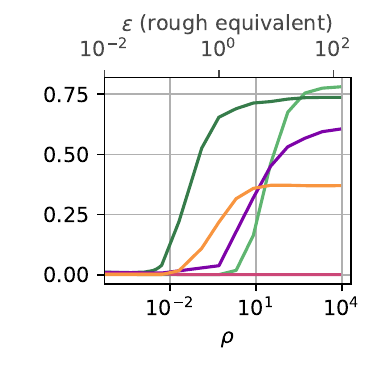}
     \end{subfigure}
     \hfill
     \begin{subfigure}[t]{0.22\columnwidth}
         \centering
         \includegraphics[width=\textwidth]{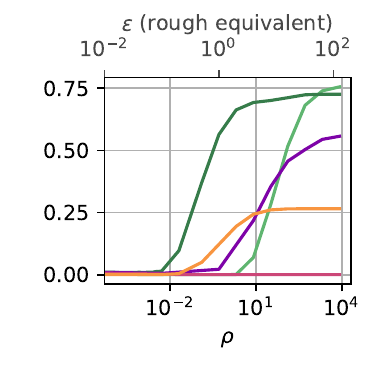}
     \end{subfigure}

    \makebox[20pt]{\raisebox{20pt}{\rotatebox[origin=c]{90}{ResNet-50}}}%
    \begin{subfigure}[t]{0.22\columnwidth}
         \centering
         \includegraphics[width=\textwidth]{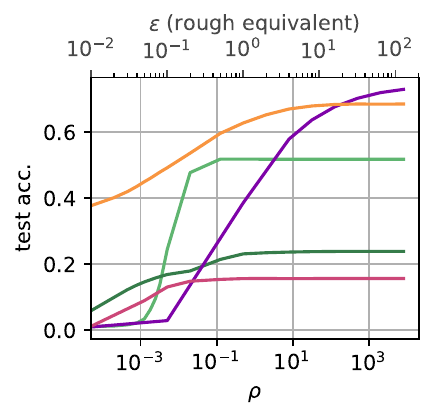}
         \caption{IR=1.}
     \end{subfigure}
     \hfill
     \begin{subfigure}[t]{0.22\columnwidth}
         \centering
         \includegraphics[width=\textwidth]{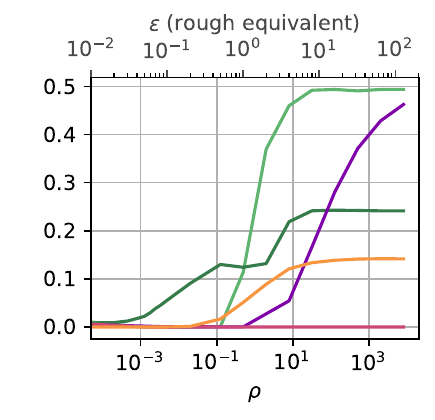}
         \caption{IR=10.}
     \end{subfigure}
     \hfill
     \begin{subfigure}[t]{0.22\columnwidth}
         \centering
         \includegraphics[width=\textwidth]{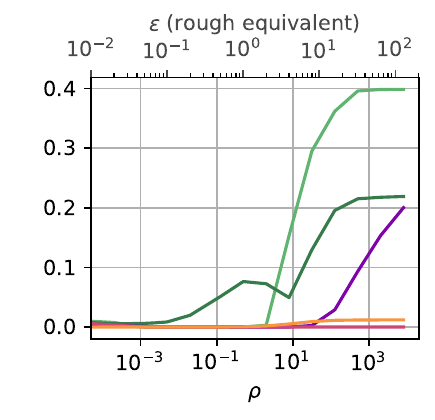}
         \caption{IR=50.}
     \end{subfigure}
     \hfill
     \begin{subfigure}[t]{0.22\columnwidth}
         \centering
         \includegraphics[width=\textwidth]{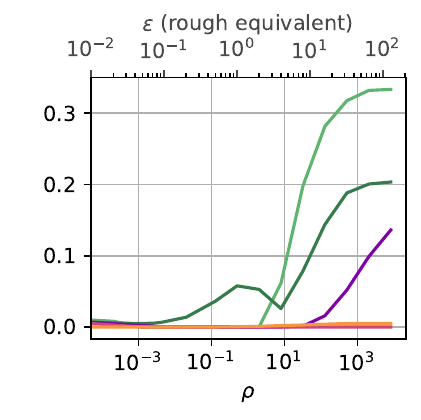}
         \caption{IR=100.}
     \end{subfigure}
\caption{\textbf{Minority class accuracies on CIFAR100.}
We present the results for the minority classes (lower 25\% quantile) of CIFAR100 on ViT-B-16, ViT-H-14, ViT-L-16 and ResNet-50, using ImageNet as public data for \DPthree, at different levels of imbalance rations (IR).
We compare to DP-LS by \citet{mehtaDifferentiallyPrivateImage2023} and DPSGD-Global-Adapt by \citet{esipova2023disparate}.
}
\label{fig:app-minority-cifar100}
\end{figure}

\begin{figure}[t]
\centering 
    \includegraphics[width=\columnwidth]{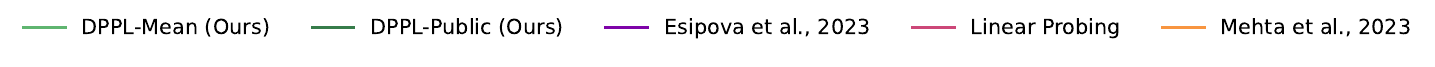}

    \makebox[20pt]{\raisebox{20pt}{\rotatebox[origin=c]{90}{ViT-B-16}}}%
    \begin{subfigure}[t]{0.22\columnwidth}
         \centering
         \includegraphics[width=\textwidth]{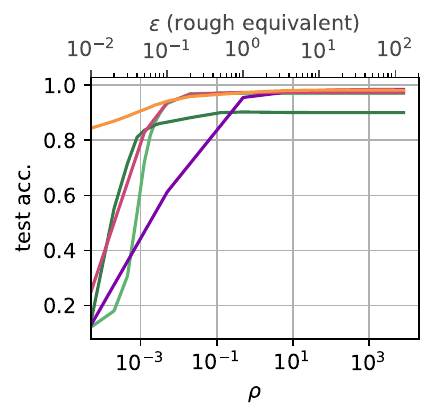}
     \end{subfigure}
     \hfill
    \begin{subfigure}[t]{0.22\columnwidth}
         \centering
         \includegraphics[width=\textwidth]{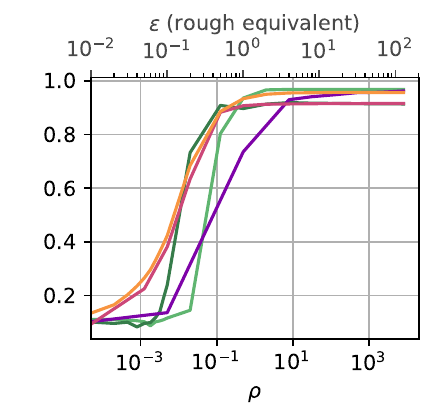}
     \end{subfigure}
     \hfill
     \begin{subfigure}[t]{0.22\columnwidth}
         \centering
         \includegraphics[width=\textwidth]{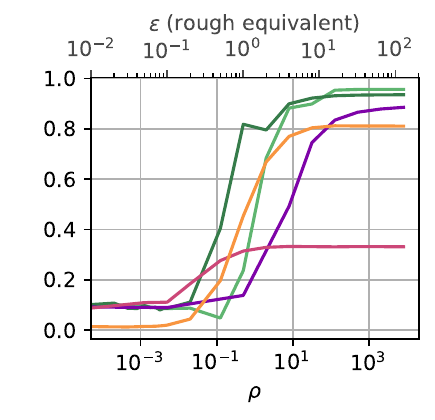}
     \end{subfigure}
     \hfill
     \begin{subfigure}[t]{0.22\columnwidth}
         \centering
         \includegraphics[width=\textwidth]{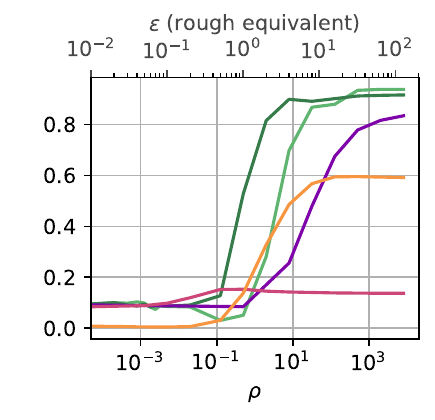}
     \end{subfigure}

    \makebox[20pt]{\raisebox{20pt}{\rotatebox[origin=c]{90}{ViT-L-16}}}%
    \begin{subfigure}[t]{0.22\columnwidth}
         \centering
         \includegraphics[width=\textwidth]{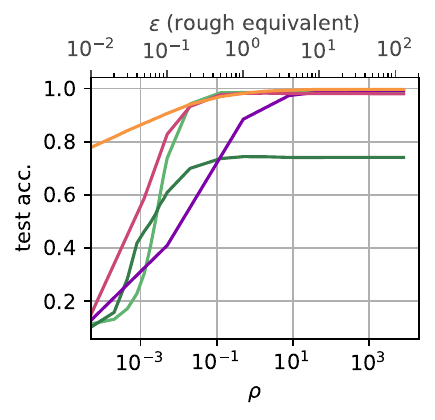}
     \end{subfigure}
     \hfill
     \begin{subfigure}[t]{0.22\columnwidth}
         \centering
         \includegraphics[width=\textwidth]{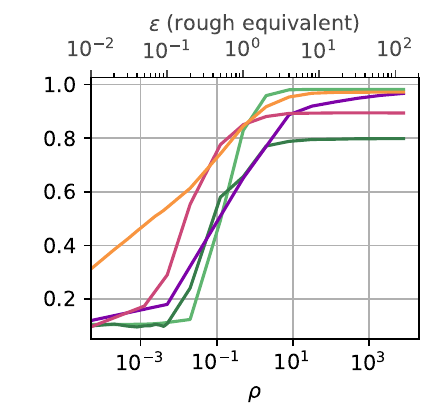}
     \end{subfigure}
     \hfill
     \begin{subfigure}[t]{0.22\columnwidth}
         \centering
         \includegraphics[width=\textwidth]{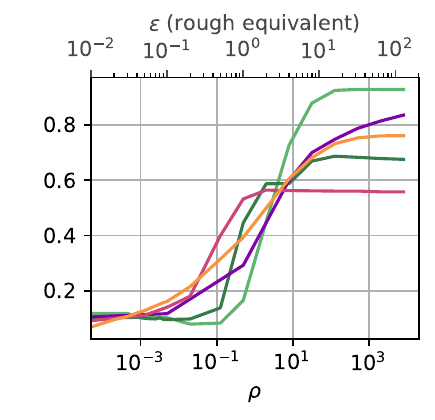}
     \end{subfigure}
     \hfill
     \begin{subfigure}[t]{0.22\columnwidth}
         \centering
         \includegraphics[width=\textwidth]{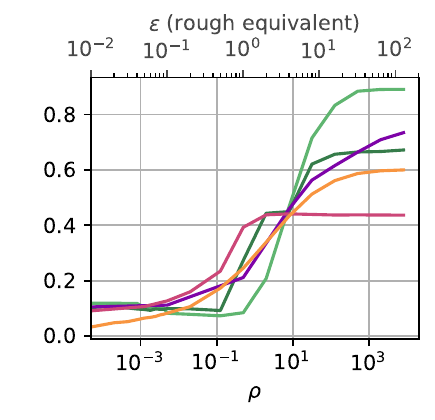}
     \end{subfigure}

    \makebox[20pt]{\raisebox{20pt}{\rotatebox[origin=c]{90}{ViT-H-14}}}%
     \begin{subfigure}[t]{0.22\columnwidth}
     \centering
     \includegraphics[width=\textwidth]{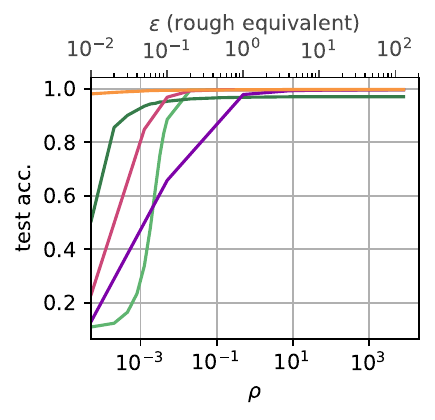}
     \end{subfigure}
     \hfill
    \begin{subfigure}[t]{0.22\columnwidth}
         \centering
         \includegraphics[width=\textwidth]{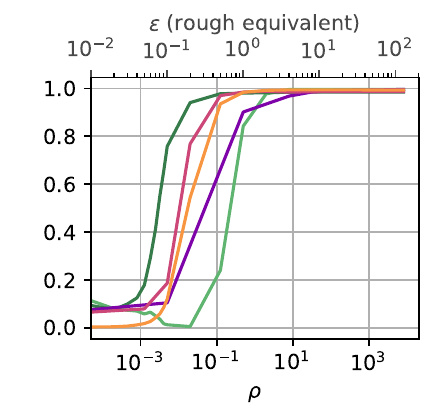}
     \end{subfigure}
     \hfill
     \begin{subfigure}[t]{0.22\columnwidth}
         \centering
         \includegraphics[width=\textwidth]{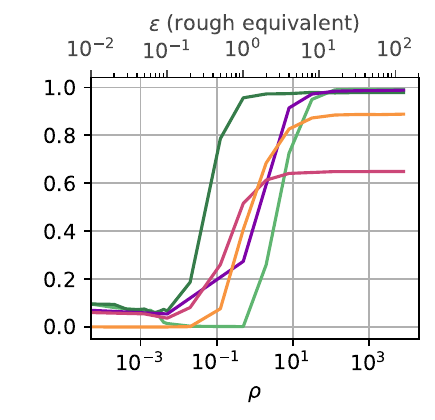}
     \end{subfigure}
     \hfill
     \begin{subfigure}[t]{0.22\columnwidth}
         \centering
         \includegraphics[width=\textwidth]{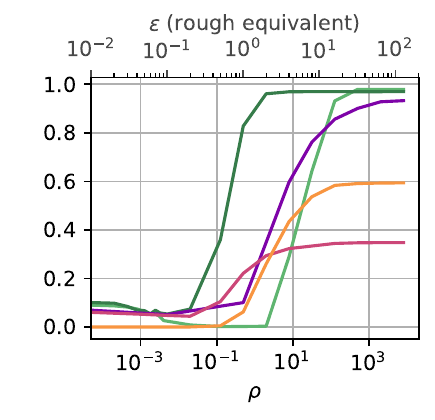}
     \end{subfigure}

    \makebox[20pt]{\raisebox{20pt}{\rotatebox[origin=c]{90}{ResNet-50}}}%
    \begin{subfigure}[t]{0.22\columnwidth}
         \centering
         \includegraphics[width=\textwidth]{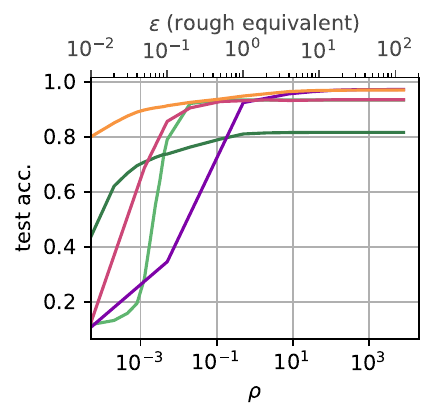}
         \caption{IR=1.}
     \end{subfigure}
     \hfill
     \begin{subfigure}[t]{0.22\columnwidth}
         \centering
         \includegraphics[width=\textwidth]{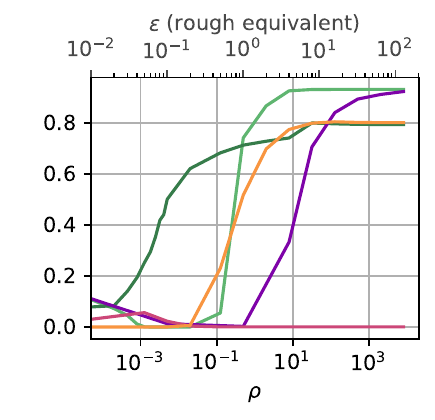}
         \caption{IR=10.}
     \end{subfigure}
     \hfill
     \begin{subfigure}[t]{0.22\columnwidth}
         \centering
         \includegraphics[width=\textwidth]{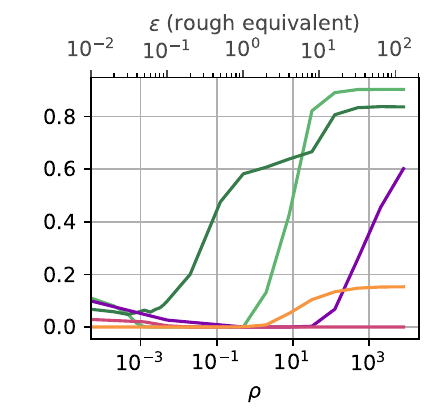}
         \caption{IR=50.}
     \end{subfigure}
     \hfill
     \begin{subfigure}[t]{0.22\columnwidth}
         \centering
         \includegraphics[width=\textwidth]{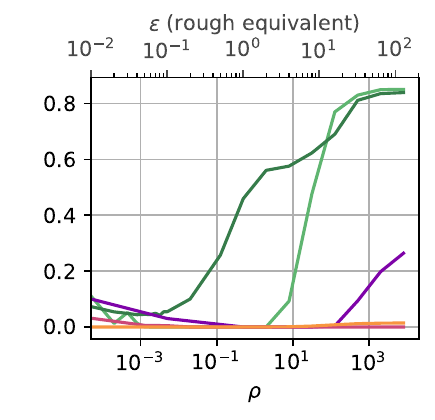}
         \caption{IR=100.}
     \end{subfigure}
\caption{\textbf{Minority class accuracies on STL10.}
We present the results for the minority classes (lower 25\% quantile) of STL10 on ViT-B-16, ViT-H-14, ViT-L-16 and ResNet-50, using ImageNet as public data for \DPthree, at different levels of imbalance rations (IR).
We compare to DP-LS by \citet{mehtaDifferentiallyPrivateImage2023} and DPSGD-Global-Adapt by \citet{esipova2023disparate}.
}
\label{fig:app-minority-stl10}
\end{figure}

\begin{figure}[t]
\centering 
    \includegraphics[width=\columnwidth]{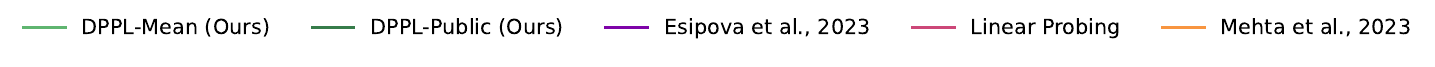}

    \makebox[20pt]{\raisebox{20pt}{\rotatebox[origin=c]{90}{ViT-B-16}}}%
    \begin{subfigure}[t]{0.22\columnwidth}
         \centering
         \includegraphics[width=\textwidth]{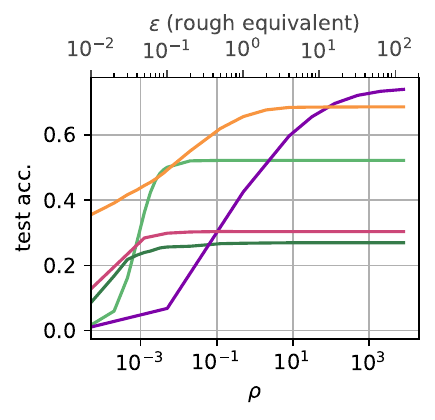}
     \end{subfigure}
     \hfill
    \begin{subfigure}[t]{0.22\columnwidth}
         \centering
         \includegraphics[width=\textwidth]{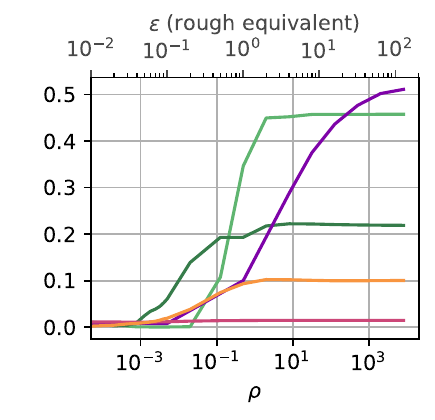}
     \end{subfigure}
     \hfill
     \begin{subfigure}[t]{0.22\columnwidth}
         \centering
         \includegraphics[width=\textwidth]{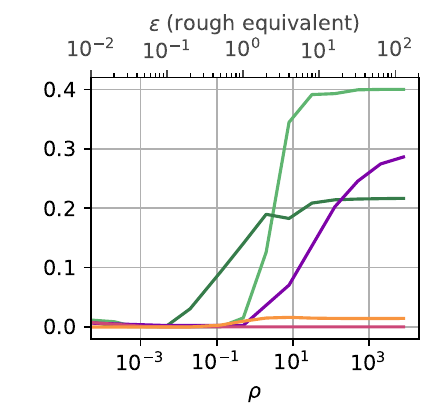}
     \end{subfigure}
     \hfill
     \begin{subfigure}[t]{0.22\columnwidth}
         \centering
         \includegraphics[width=\textwidth]{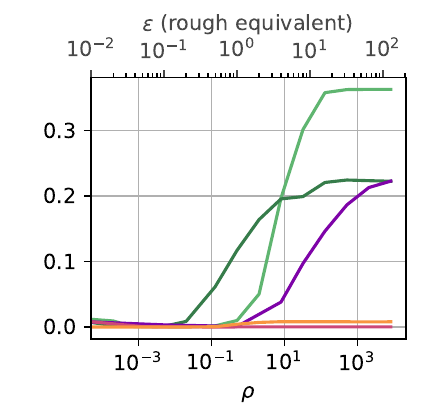}
     \end{subfigure}

    \makebox[20pt]{\raisebox{20pt}{\rotatebox[origin=c]{90}{ViT-L-16}}}%
    \begin{subfigure}[t]{0.22\columnwidth}
         \centering
         \includegraphics[width=\textwidth]{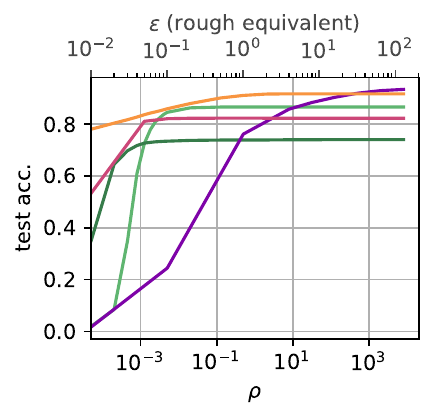}
     \end{subfigure}
     \hfill
     \begin{subfigure}[t]{0.22\columnwidth}
         \centering
         \includegraphics[width=\textwidth]{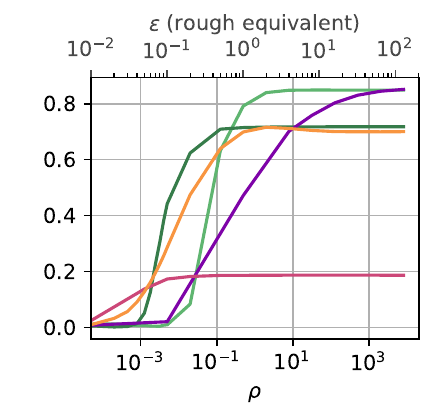}
     \end{subfigure}
     \hfill
     \begin{subfigure}[t]{0.22\columnwidth}
         \centering
         \includegraphics[width=\textwidth]{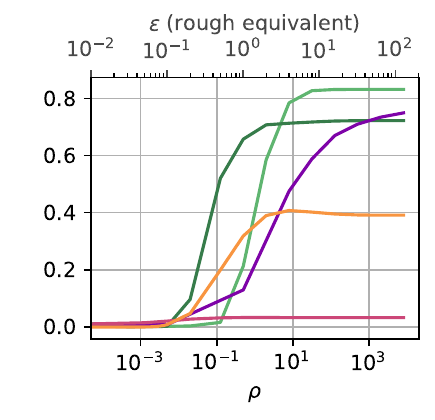}
     \end{subfigure}
     \hfill
     \begin{subfigure}[t]{0.22\columnwidth}
         \centering
         \includegraphics[width=\textwidth]{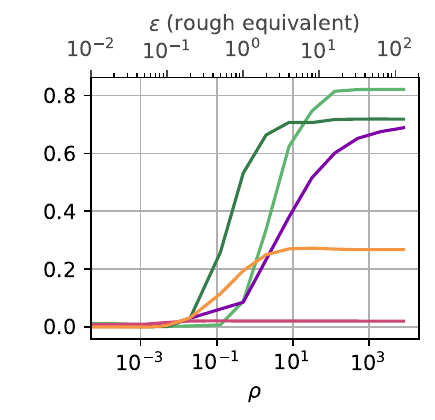}
     \end{subfigure}

    \makebox[20pt]{\raisebox{20pt}{\rotatebox[origin=c]{90}{ViT-H-14}}}%
     \begin{subfigure}[t]{0.22\columnwidth}
     \centering
     \includegraphics[width=\textwidth]{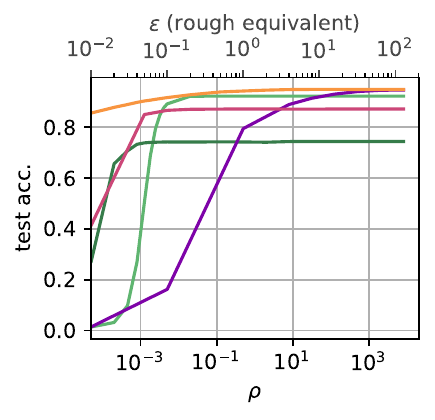}
     \end{subfigure}
     \hfill
    \begin{subfigure}[t]{0.22\columnwidth}
         \centering
         \includegraphics[width=\textwidth]{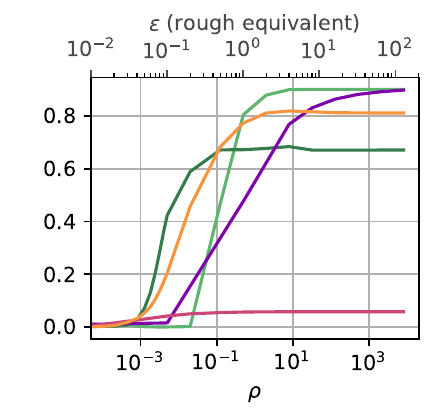}
     \end{subfigure}
     \hfill
     \begin{subfigure}[t]{0.22\columnwidth}
         \centering
         \includegraphics[width=\textwidth]{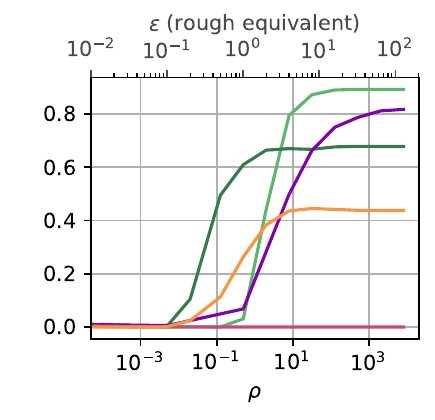}
     \end{subfigure}
     \hfill
     \begin{subfigure}[t]{0.22\columnwidth}
         \centering
         \includegraphics[width=\textwidth]{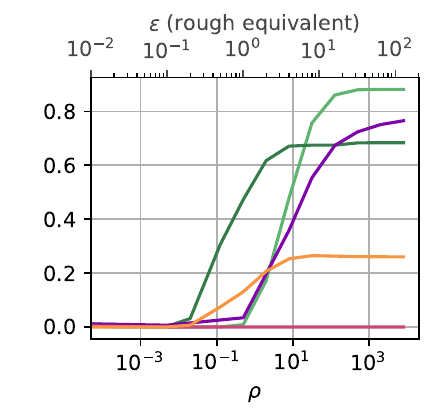}
     \end{subfigure}

    \makebox[20pt]{\raisebox{20pt}{\rotatebox[origin=c]{90}{ResNet-50}}}%
    \begin{subfigure}[t]{0.22\columnwidth}
         \centering
         \includegraphics[width=\textwidth]{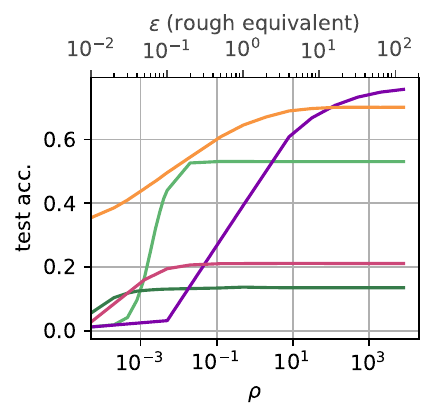}
         \caption{IR=1.}
     \end{subfigure}
     \hfill
     \begin{subfigure}[t]{0.22\columnwidth}
         \centering
         \includegraphics[width=\textwidth]{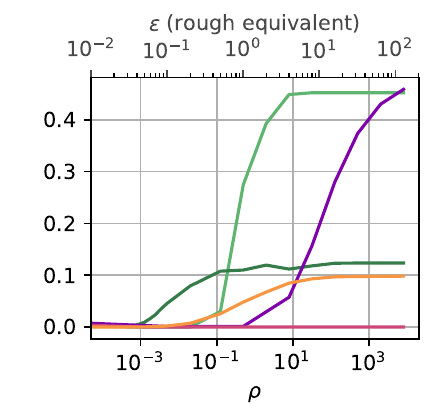}
         \caption{IR=10.}
     \end{subfigure}
     \hfill
     \begin{subfigure}[t]{0.22\columnwidth}
         \centering
         \includegraphics[width=\textwidth]{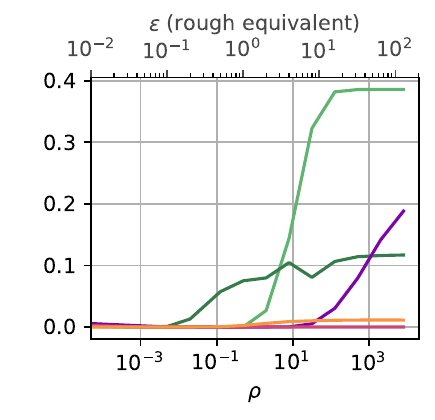}
         \caption{IR=50.}
     \end{subfigure}
     \hfill
     \begin{subfigure}[t]{0.22\columnwidth}
         \centering
         \includegraphics[width=\textwidth]{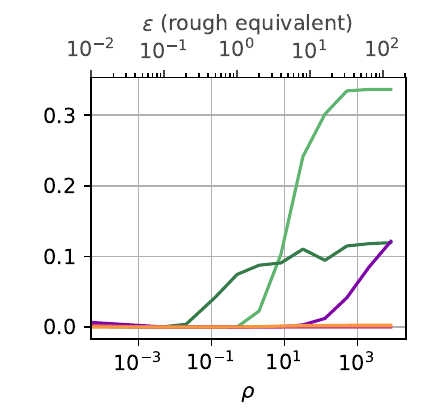}
         \caption{IR=100.}
     \end{subfigure}
\caption{\textbf{Minority class accuracies on FOOD101.}
We present the results for the minority classes (lower 25\% quantile) of FOOD101 on ViT-B-16, ViT-H-14, ViT-L-16 and ResNet-50, using ImageNet as public data for \DPthree, at different levels of imbalance rations (IR).
We compare to DP-LS by \citet{mehtaDifferentiallyPrivateImage2023} and DPSGD-Global-Adapt by \citet{esipova2023disparate}.
}
\label{fig:app-minority-food101}
\end{figure}

We compare the accuracies of the minority classes for all methods, all encoders and imbalance ratios in $[1,10,50,100]$ in \Cref{fig:app-minority-cifar10,fig:app-minority-stl10,fig:app-minority-food101,fig:app-minority-cifar100}.
\subsection{Computational Runtimes}
\label{app:runtimes}
\begin{table}[t]
\centering

\label{tab:runtime}
\resizebox{\columnwidth}{!}{%
\begin{tabular}{l|l|ll}
\multirow{2}{*}{}                         & \multirow{2}{*}{Step} & \multicolumn{2}{c}{Runtime [s]} \\
                                          &                       & CIFAR10        & CIFAR100       \\ \hline
\DPtwo (Ours)                             & Mean Estimation       & $0.079$        & $0.168$        \\ \hline
\DPthree (Ours)                           & Utility Calculation   & $5.0$          & $34.3$         \\
                                          & Private Sampling      & $0.0003$       & $0.14$         \\  \hline
Linear Probing (DPSGD)                   & Iterative Training    & $5.49$         & $6.3$          \\ \hline
Esipova et al., 2023 (DPSGD-Global-Adapt) & Iterative Training    & $174$          & $242$          \\ \hline
Mehta et al., 2023 (DP-LS)                & Setup                 & $0.49$         & $2.2$          \\
                                          & Solving               & $0.28$         & $2.5$           
\end{tabular}
}%
\caption{\textbf{Computational wall-time measurements} of a single training on a single machine, limiting each method to a single GPU. Where applicable, iterative training was limited to 15 epochs. \DPthree's score computation was conducted for 1,281,167 public samples.}
\end{table}

We compare the runtime for a single training in \Cref{tab:runtime}. We chose to compare CIFAR10 and CIFAR100 because the runtime of all methods scale with the number of classes on otherwise equally large training datasets.
\subsection{Potential Baselines}
\label{app:potentialbaselines}
We further considered DP-FC introduced by \citet{mehtaDifferentiallyPrivateImage2023} and DP-FiLM introduced by \citet{tobabenEfficacyDifferentiallyPrivate2023} as baseline methods. 
\subsubsection{DP-FiLM}
While DP-FiLM exhibits strong learning potential from few samples, it is an iterative algorithm and comes with the same drawbacks for unbalanced tasks as DP-SGD. We conducted initial experiments for which we show the results in \Cref{tab:filmvspublic}. We compare DP-FiLM on ViT-H-14 on CIFAR10 and CIFAR100. Both methods were trained at the same value of $\epsilon$. We set $\delta$ for DP-FiLM to ${1}/{2n}$ where $n$ is the number of training samples. Our method provides $\delta=0$ pure DP. We note the significantly lower utility of DP-FiLM in imbalanced cases and considering the high computational costs --- other methods require training in the range $10^{-1} \text{ to } 10^{2}$ GPU-seconds, whereas DP-FiLM requires $10^{5} \text{ to } 10^{6}$ GPU-seconds--- decided against a comprehensive comparison.
\subsubsection{DP-FC}
\begin{figure}[t]
\centering
\begin{minipage}{0.2\textwidth}
    \centering
    \includegraphics[width=\textwidth]{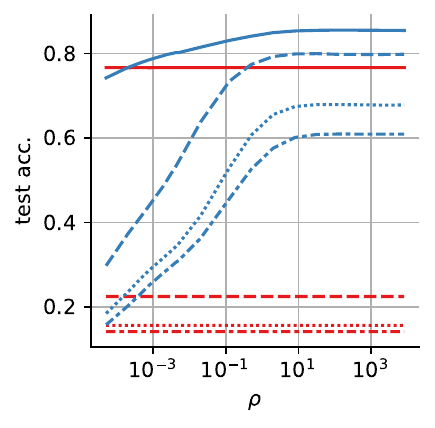}
\end{minipage}
\begin{minipage}{0.15\textwidth}
    \centering
    \includegraphics[width=\textwidth]{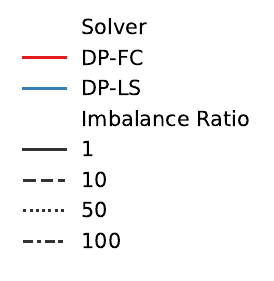}
\end{minipage}
\caption{\textbf{DP-LS vs. DP-FC}.}
\label{fig:ablation-lsfc}
\end{figure}
DP-FC is an iterative optimization algorithm that integrates second order information by utilizing the covariance of the features. 
\Cref{fig:ablation-lsfc} shows results on imbalanced datasets. 
While DP-FC slightly outperforms DP-LS for very strict privacy budgets in balanced cases, it exhibits the same disparate effects on minority classes as DP-SGD, resulting in reduced utility for imbalanced cases. Given that it has only a small advantage in balanced cases and otherwise large disadvantage in unbalanced cases, we focused on DP-LS as a non-iterative baseline instead.
\begin{table}[t]
\centering
\resizebox{\columnwidth}{!}{%
\begin{tabular}{l|l|llll}
Dataset                   & Method \ $\epsilon$ & 0.1  & 0.5  & 1.0  & 2.0  \\ \hline
\multirow{2}{*}{CIFAR10}  & \DPthree          & 66.1 & 92.9 & 92.5 & 92.9 \\
                          & DP-FiLM          & 34.1 & 63.3 & 67.7 & 84.4 \\ \hline
\multirow{2}{*}{CIFAR100} & \DPthree          & 27.4 & 51.2 & 59.9 & 70.0 \\
                          & DP-FiLM          & 4.8  & 20.1 & 34.1 & 45.2
\end{tabular}%
}
\caption{\textbf{DP-FiLM vs. DPPL-Public} using ImageNet-1K as public data on CIFAR10 and CIFAR100 with imbalance ratio $100$. We compare at the same $\epsilon$ value, although DP-FiLM provides approximate DP and our method provides more strict pure DP.}
\label{tab:filmvspublic}
\end{table}
\section{Discussion}
\subsection{Broader Impacts}
\label{app:broaderimpacts}
We expect the prevalence of machine learning and it's impact on society to ever increase. Our methods are especially useful at preserving the privacy of the training data, data that often consists of sensitive data from users. We consider contributing to an increase in the privacy of the training data and therefore protecting the users that contribute data to machine learning models to be a positive societal impact. Furthermore, our methods especially address the use case of imbalanced datasets. Real-world data is often long-tailed and models trained on unbalanced data can lead to unfair decisions w.r.t. to gender, ethnicity, disabilities, religion or social status, especially for minorities. We consider contributing to an improvement of the utility for minority classes as outlined in \Cref{fig:small_minority} and \Cref{app:minority-acc} to be a positive societal impact.
\subsection{Limitations}
\label{app:limitations}
We introduce \ours as a novel approach to private transfer learning. Like all transfer learning methods, our method depends on a suitable base model. We've seen that especially ResNet-50 poses significant challenges, while the vision transformers worked well. The largest vision transformer ViT-H-14 yielded the best results of the compared models. We note that the combination of less suitable base models in addition to further out-of-distribution tasks, relative to the pre-training data, has a larger negative effect on the performance of our method compared to other methods. When evaluating the most out-of-distribution dataset, FOOD101, in combination with using embeddings from ResNet-50 or ViT-B-16, our methods are outperformed (see \Cref{fig:app-food101}. We can still claim the highest accuracy for minority classes in that case (see \Cref{fig:app-minority-food101}), although the significance of that given the low utility is questionable. As we didn't include the projection layer for our methods, the ability to adapt to these distribution shifts is limited and possibilities to include it need to be investigated further.

\end{document}